\documentclass[twoside,11pt]{article}

\usepackage{xcolor} \pagecolor[rgb]{1,1,1} \color[rgb]{0,0,0}

\usepackage{amssymb,amsmath,amscd}
\usepackage{mathabx}
\usepackage{mathtools}
\usepackage{tabularx}
\usepackage{threeparttable}
\usepackage{booktabs}
\usepackage{subcaption}
\usepackage{multirow}
\usepackage{float}
\usepackage [english]{babel}
\usepackage [autostyle, english = american]{csquotes}
\usepackage{enumitem}
\MakeOuterQuote{"}

\usepackage[ruled,vlined]{algorithm2e}
\usepackage[preprint]{jmlr2e}
\newcommand{\dd}{\mathrm{d}}

\ShortHeadings{Geometry-Aware Hamiltonian VAE}{Chadebec et al. }
\firstpageno{1}

\begin{document}

\title{Geometry-Aware Hamiltonian Variational Auto-Encoder}

\author{\name  Clément Chadebec \email clement.chadebec@sorbonne-universite.fr \\
       \addr Centre de Recherche des Cordeliers UMRS 1138\\
       Université de Paris, INSERM, Sorbonne Université\\
       15 rue de l'école de médecine\\
       75006 Paris, France
       \\
       
       \AND
       \name Clément Mantoux \email clement.mantoux@inria.fr \\
       \addr Centre de Mathématiques Appliquées, Inria \\
       École Polytechnique, Institut Polytechnique de Paris\\
       91128 Palaiseau, France\\
       \\
       
       \AND
       \name Stéphanie Allassonnière \email stephanie.allassonniere@parisdescartes.fr \\
       \addr Centre de Recherche des Cordeliers UMRS 1138\\
       Université de Paris, INSERM, Sorbonne Université\\
       15 rue de l'école de médecine\\
       75006 Paris, France
       }

\editor{TBC}

\maketitle

\begin{abstract}
    Variational auto-encoders (VAEs) have proven to be a well suited tool for performing dimensionality reduction by extracting latent variables lying in a potentially much smaller dimensional space than the data. Their ability to capture meaningful information from the data can be easily apprehended when considering their capability to generate new realistic samples or perform potentially meaningful interpolations in a much smaller space. However, such generative models may perform poorly when trained on \textit{small} data sets which are abundant in many real-life fields such as medicine. This may, among others, come from the lack of structure of the latent space, the geometry of which is often under-considered. We thus propose in this paper to see the latent space as a Riemannian manifold endowed with a parametrized metric learned at the same time as the encoder and decoder networks. This metric is then used in what we called the Riemannian Hamiltonian VAE which extends the Hamiltonian VAE introduced by \citet{caterini2018hamiltonian} to better exploit the underlying geometry of the latent space. We argue that such latent space modelling provides useful information about its underlying structure leading to far more meaningful interpolations, more realistic data-generation and more reliable clustering.
\end{abstract}

\begin{keywords}
  Variational auto-encoders, Metric learning, Normalizing flows, Latent space modelling.
\end{keywords}

\section{Introduction}

    Driven by the apparent availability of always bigger data sets, deep generative models have become more and more data greedy. Most of the time they need thousands of training data samples to be able to generate faithfully new data-looking samples. Nonetheless, in many fields of application the number of data remains a key issue.  For example, in neuroscience, practitioners have to deal with high dimensional data combined with a very small number of samples which can make classic methods hard to rely on \citep{button2013power, turner2018small}. Recently, \citet{szucs2020sample} studied the sample size evolution (i.e. the number of participants) in neuroimaging studies of most cited papers published between 1990 and 2012 representing 1038 contributions. They compared them to 270 papers published in best-in-class neuroimaging journals between 2017 and 2018. One of the key outcome of such a study is that 96\% of most cited experimental functional Magnetic Resonance (brain) Imaging (fMRI) studies were based on a median sample size equals to 12, this number goes to 14.5 when one considers clinical studies and to 50 for clinical structural analysis. We refer the reader to Table~3 of their paper highlighting the number of participants in various studies. Their study concludes that the median sample size slightly increases at a rate of 0.74 participant/year.  These very small sample sizes make conventional machine learning methods unreliable because they do not provide statistically significant results and sufficient variability between subjects within a given study group. A way to address this "missing" data issue would consist in trying to create synthetic samples that could have been part of the "true" data set and use them in classic machine learning methods. Such an approach can also be used to create synthetic data sets to overcome the privacy issue of confidential data that cannot be used directly.
    
    One of the tools one may think of are variational auto-encoders. First introduced by \citet{kingma2013auto} and \citet{rezende2014stochastic}, they have proven to be well designed to perform dimensionality reduction and be able to represent potentially very high dimensional complex data within a much smaller space. Even more appealing is their ability to generate new \textit{realistic} data. These two aspects are of interest as ideally we could use the VAE framework to 1) reduce data dimension which may be useful for performing the analysis of high dimensional data such a fMRI; 2) be able to structure the latent space such that interpolations between images of subjects correspond to meaningful deformations and 3) create synthetic data-looking points having the desired properties and use them to train classic deep models.

    Unfortunately, when small data sets are considered the generated samples are most of the time very blurry and variational auto-encoders hardly perform well in terms of generation. To tackle such an issue \citet{loaiza2019continuous} proposed to use a continuous Bernoulli distribution instead of the discrete one which is usually used. However, they only changed the \textit{decoding} distribution in order to improve the Evidence Lower BOund (ELBO) and did not pay attention to geometrical aspects. Moreover, their model was only trained on large data sets (tens of thousands of samples) and it remains unclear if it could perform well on much smaller data sets (e.g. hundreds of samples). Improving the ELBO has been the subject of great interest in recent years and a central point in various research papers \citep[see][]{alemi2018fixing, burda2015importance, cremer2018inference, higgins2017beta, ruiz2019contrastive, zhang2018advances}. As mentioned above, one way to achieve a tighter lower bound would consist in changing the \textit{decoding} and/or the \textit{encoding} distribution (i.e. the approximate posterior distribution). While, the first point was explored in \citep{loaiza2019continuous}, many works focused on tweaking the approximate posterior distribution of the latent variables given the observations. For example, \citet{rezende2015variational} used \textit{Normalizing Flows} consisting in smooth invertible transformations applied to the latent variable and aiming at achieving richer approximate distributions. Similarly, \citet{salimans2015markov} proposed a method involving Markov Chain Monte Carlo sampling steps targeting the true posterior distribution and using \textit{deterministic} kernels based on Hamiltonian dynamics. This work was further extended by \citet{caterini2018hamiltonian} who introduced the Hamiltonian variational auto-encoder. Nonetheless, none of these methods uses the underlying geometry of the latent space which we believe may be of interest. Even though trying to improve the posterior distribution revealed to be a good idea, \citet{hoffman2016elbo} proposed a new writing of the ELBO objective highlighting that particular attention should be paid to the prior distribution as well. This led \citet{dilokthanakul2016deep} to use a Gaussian mixture as prior distribution for the latent variables. Going further, \citet{tomczak2017vae} proposed to use a "VAriational Mixture of Priors" (VAMP) resulting in better lower bound. By arguing that classic VAE fails to apprehend data with a specific geometry, \citet{davidson2018hyperspherical} used Von Mises-Fisher distributions for the prior and posterior distributions paving the way to further investigate geometrical aspects. \citet{rey2019diffusion} proposed the diffusion variational auto-encoder along with various latent space modellings. However, to the best of our knowledge such an approach still requires a prior knowledge of the latent space structure which is not necessary with the method we propose. While trying to bring some structuring to the latent space, \citet{arvanitidis2017latent} proposed to see it as a Riemannian manifold and so proposed to endow this space with a Riemannian metric. This metric is given by the Jacobian of the generator of the VAE. Their main objective was to use such a metric to perform clustering tasks using a $\mathcal{N}$-VAE. However, the Jacobian of the generator may be hard and time-consuming to compute and it remains unclear if such a metric is well suited to other models and different tasks such as generation and interpolations.
    
    Although tweaking the variational approximate posterior distribution using either normalizing flows or Markov Chain Monte Carlo sampling appears to be one of the most promising ways to improve the model, exploiting the underlying geometry of the latent space may provide useful information as well. We will introduce in this paper the Riemannian Hamiltonian variational auto-encoder aiming at combining both approaches. This model can be seen as a \textit{geometry-aware} Hamiltonian VAE based on Riemannian Hamiltonian dynamics as discussed in \citep{girolami2009riemannian} and using a metric we propose to learn directly form the data.  We will see that such a model is able to provide an interesting latent space structuring which reveals well suited for performing geodesic interpolation, generation and clustering especially in the context of \textit{small} size data sets.

    \section{Model Setting}
    
    Given a set of data $x \in \mathcal{X}$ and a parametric model $\{\mathbb{P}_{\theta}; \theta \in \Theta \}$, variational auto-encoders aim at finding the parameter $\theta$ maximising the marginal likelihood of the data $p_{\theta}(x)$. Assuming that the data generation process involves a continuous latent variable $z \in \mathcal{Z}$ living in a smaller space, the marginal likelihood can be written as follows:
    \begin{equation}\label{Eq: Model Setting}
        p_{\theta}(x) = \int p_{\theta}(x |z) q_{\mathnormal{prior}}(z) dz\,,
    \end{equation}
    where $q_{\mathnormal{prior}}(z)$ is a prior distribution over the latent variables generally chosen as a standard normal distribution. One way to compute $p_{\theta}(x)$ would consist in using both the joint distribution $p_{\theta}(x, z)$ and the posterior distribution $p_{\theta}(z| x)$. However, the latter is most of the time intractable. Hence, a variational approximation $q_{\phi}(z|x)$ of the true posterior distribution is introduced and is often referred to as the \textit{encoder} \citep{kingma2013auto}. An unbiased estimate of the marginal likelihood then writes
    \[
        \hat{p}_{\theta}(x) = \frac{p_{\theta}(x, z)}{q_{\phi}(z|x)}\,,
    \]
    where $z \sim q_{\phi}(z|x)$. Applying Jensen's inequality to the above expression, we obtain the Evidence Lower BOund (ELBO) on the log-likelihood of the marginal distribution:
    \begin{equation}\label{ELBO}
        \log p_{\theta}(x)  \geq \mathbb{E}_{z \sim q_{\phi}(z|x)} [\log p_{\theta}(x, z) - \log q_{\phi}(z|x)] = ELBO\,.
    \end{equation}

    Using the reparametrization trick \citep{kingma2013auto} makes an estimate of the ELBO differentiable with respect to $\phi$ and so gives access to an unbiased estimate of the gradient of the ELBO . Recent works have been trying to tweak the variational posterior approximation $q_{\phi}(z|x)$ to achieve a better estimate of the true posterior $p_{\theta}(z|x)$ which would ideally make the inequality in Eq.~\eqref{ELBO} an equality. An approach was proposed by \citet{salimans2015markov} and consists in adding a fixed number of MCMC steps to the variational posterior approximation targeting the true posterior distribution $p_{\theta}(z|x)$ as follows:
    \[
        \hat{p}_{\theta}(x) = \frac{p_{\theta}(x, z_T) \prod_{t=1}^{T} r(z_{t-1} | z_{t}, x)}{q_{\phi}(z_0|x) \prod_{t=1}^T r(z_t |z_{t-1}, x)} \,,
    \] where $z_0 \sim q_{\phi}(z|x)$, $r(z_t |z_{t-1}, x)$ is the transition kernel from which $z_t$ is sampled and $r(z_{t-1} | z_{t}, x)$ is the \textit{reverse} kernel. This method requires forward and reverse transition kernels that may have to be parametrized and learned as well. An other approach is to consider smooth invertible parametrized mappings $f$ called \textit{Normalizing flows} \citep{rezende2015variational}. $K$ transformations are then applied to a latent variable $z_0$ drawn from an initial distribution $q$ (here $q = q_{\phi}$) leading to a final random variable $z_K = f^K_{x} \circ \cdots \circ f^1_x(z_0)$ whose density writes
    \begin{equation}\label{Eq: Normailizing flows}
        q_{\phi}(z_K|x) = q_{\phi}(z_0|x) \prod_{k=1}^K |\det \mathbf{J}_{f^k_x}|^{-1} \,,
    \end{equation}
    where $\mathbf{J}_{f^k_x} = \frac{\partial f^k_x}{\partial z}$. These mappings are essentially parametrized and learned during the learning process. \citet{caterini2018hamiltonian}  proposed to use a method based on Hamiltonian Monte Carlo dynamics and combining both approaches to produce an unbiased estimate of
    ${p}_{\theta}(z|x)$.

    \subsection{Hamiltonian Markov Chain Monte Carlo}

    The method proposed in \citep{caterini2018hamiltonian} is inspired by the
Hamiltonian Monte Carlo sampler (HMC) which has been studied in several papers  \citep[see][]{neal2011mcmc,
livingstone2019geometric,durmus2017convergence,betancourt2017geometric}. In the
HMC framework, a random variable $z$ is assumed to live in an Euclidean space
and to follow a target density $\pi$ deriving from a potential $U$ such that
the distribution writes
\begin{equation}\label{Eq: Potential}
    \pi(z) = \frac{e^{-{U(z)}}}{\int e^{-{U(\widebar{z})}} d\widebar{z}}\,,
\end{equation}
where $U(z) = - \log \pi(z)$. Since it is most of the time impossible to sample directly from $\pi$, an independent auxiliary random variable $\rho \in \mathbb{R}^d$ is introduced and used to "sample" $z$. This variable is often referred to as the \textit{momentum}
and is such that $\rho \sim \mathcal{N}(0, \mathbf{M})$ where $\mathbf{M}$ is
called the \textit{mass matrix}. The idea behind the HMC is to work with the
extended target probability distribution $\pi(z, \rho) = p(z|\rho) p(\rho) = \pi(z) p(\rho) $ whose density writes
\[
    \pi(z, \rho) = \frac{e^{-H(z, \rho)}}{\int_{\mathbb{R}^{2d}} e^{- H(z, \rho)} dz d\rho}\,,
\]
where $H(z, \rho)$ is called the  \textit{Hamiltonian}
\citep{duane1987hybrid,leimkuhler2004simulating} and corresponds to the negative
log-density of the extended target distribution
\begin{equation}
\begin{aligned}\label{eq: Euclidean Hamiltonian}
    H(z, \rho) = - \log \pi(z, \rho) &= - \log \pi(z) + \frac{1}{2} \log ((2 \pi)^{d} |\mathbf{M}|) + \rho^{\top} \mathbf{M}^{-1} \rho\\
                        &=U(z) + \kappa(\rho)\,.
\end{aligned}
\end{equation}
In physics, the equation gives the total energy of a physical system having a \textit{position} $z$ and a \textit{momentum} $ \rho$. $U$ is referred to as the potential energy and $\kappa$ is called the kinetic energy. The evolution in time $(z(t), \rho(t))$ of such a system is given by Hamilton's equations as follows:

\begin{equation}
\left\{\begin{aligned}\label{Eq: PDE}
    \frac{\partial z}{\partial t} =  \frac{\partial H}{\partial \rho} &= \mathbf{M}^{-1} \rho\,, \\ \frac{\partial \rho}{\partial t} = - \frac{\partial H}{\partial z} &= \nabla_z \log \pi(z)\,.
    \end{aligned}
\right.
\end{equation}

The solution flow $\phi_t$ of the above PDE system has to:
\begin{enumerate}[label=(\roman*)]
    \item preserve the Hamiltonian i.e. $H(\phi_t(z_0, \rho_0)) = H(z_0, \rho_0)$.
    \item be volume preserving $|\mathbf{J}_{\phi_t}| = 1$.
    \item be time-reversible.
\end{enumerate}
Unfortunately, this system of PDE is most of the time intractable and a discretization scheme
is then needed to approximate the solution and is referred to as the \textit{Stormer-Verlet} integrator. 
\begin{equation}\label{Eq: Stormer-Verlet}
    \begin{aligned}
        \rho(t + \varepsilon/2) &= \rho(t) - \frac{\varepsilon}{2} \cdot \nabla_z H(z(t), \rho(t))\,,\\
        z(t + \varepsilon) &= z(t) + \varepsilon \cdot \nabla_{\rho}(H(z(t), \rho(t + \varepsilon/2)))\,,\\
        \rho(t + \varepsilon) &= \rho(t + \varepsilon/2) - \frac{\varepsilon}{2} \cdot \nabla_z H(z(t + \varepsilon), \rho(t + \varepsilon/2))\,,
    \end{aligned}
\end{equation}
where $\varepsilon$ is the leapfrog step size.  Such a scheme is run $n_{\mathnormal{lf}}$ times to sample a proposal $(\widebar{z}, \widebar{\rho})$ which is accepted with probability $\min \Bigl(1, \exp \big(-H(\widebar{z}, \widebar{\rho}), H(z, \rho) \big) \Bigr)$. It is easy to see that negating $\varepsilon$ on each step makes the integrator  reversible (iii). In addition, the volume preserving property (ii) is ensured since the Jacobian matrix of each transformation has a unit determinant. The acceptation-rejection steps allows for the approximate preservation of the energy (i) of the integrator. Finally, this procedure is applied several times and creates an ergodic, time-reversible Markov Chain having $\pi$ as stationary distribution \citep{duane1987hybrid, liu2008monte, neal2011mcmc}.

\subsection{HMC within the VAE}\label{Sec: HMC within VAE}

The idea first introduced in \citep{salimans2015markov} and further applied to the VAE framework by \citet{caterini2018hamiltonian} is to exploit the fact that the flow created by the integrator is informed by the gradient of the target density through Eq.~\eqref{Eq: PDE}. In the VAE framework the target density $\pi$ is the true posterior distribution of the latent variables given an input data point $x$ (i.e. $\pi_x \coloneqq p_{\theta}(z|x)$). Ideally, we would like to be able to sample directly from this distribution. Unfortunately, $p_{\theta}(z|x)$ is most of the time intractable and so direct sampling is made impossible. Thinking of the HMC sampler, we would then need to be able to compute the gradient of the true posterior distribution so we can use it to sample from $p_{\theta}(z|x)$. One way to access to it is to consider the Bayesian framework. As is common one may remark that for a datapoint $x \in \mathcal{X}$, $p_{\theta}(z|x) = \frac{p_{\theta}(x, z)}{p_{\theta}(x)}$ $\propto$ $p_{\theta}(x, z)$ since we only consider the random variable $z$. Then, targeting $p_{\theta}(z|x)$ is strictly equivalent to targeting the joint distribution $p_{\theta}(x,z)$. Recall from Eq.~\eqref{Eq: Model Setting} that the model's joint distribution is such that $p_{\theta}(x,z) = p_{\theta}(x|z) q_{\mathnormal{prior}}(z)$ where $p_{\theta}(x|z)$ is the \textit{decoding} distribution and $q_{\mathnormal{prior}}(z)$ the prior distribution. Therefore, we can set the potential of Eq.~\eqref{Eq: Potential} such that $U_x(z) = - \log p_{\theta}(x, z)$ is defined for each $x \in \mathcal{X}$. Now that we have access to the gradient of the true posterior distribution trough the joint distribution which is tractable, we can use the HMC framework. Including the independent auxiliary random variable $\rho$ and writing the negative logarithm of the extended joint distribution leads to the \textit{Hamiltonian}
\[
    H_x(z, \rho) = - \log p_{\theta}(x, z, \rho) = U_x(z) +\frac{1}{2} \log ((2 \pi)^{d} |\mathbf{M}|) + \rho^{\top} \mathbf{M}^{-1} \rho \,.
\]
 Finally, $K$ iterations of the integrator as described in Eq.~\eqref{Eq: Stormer-Verlet} can then be used to sample $(z_K, \rho_K)$. We define the iterates $\{\Phi_{\varepsilon, x}^{\circ (l)}: \mathbb{R}^d \times \mathbb{R}^d \to \mathbb{R}^d \times \mathbb{R}^d, l \in \mathbb{N}^{*}\}$ where $\varepsilon$ is the leapfrog step size by induction as follows:

\[
    \Phi_{\varepsilon, x}^{\circ (l+1)} = \Phi_{\varepsilon, x}^{\circ (l)} \circ \Phi_{\varepsilon, x}^{\circ(1)}, \hspace{5mm} \Phi_{\varepsilon, x}^{\circ(0)} = I_d\,.
\]
\citet{caterini2018hamiltonian} used the Stormer-Verlet integrator combined with a tempering step as proposed in \citep{neal2005hamiltonian} to create transition kernels used to sample $(z_K, \rho_K)$. The tempering steps consist in starting from an initial temperature $\beta_0$ (which can be learned) and decreasing the \textit{momentum} $\rho$ by a factor $\alpha_k = \sqrt{\beta_{k-1} / \beta_k}$ after each leapfrog step $k$. The temperature is then updated as follows:
    \[
        \sqrt{\beta_k} = \Biggl(\Biggl(1 - \frac{1}{\sqrt{\beta_0}} \Bigg) \frac{k^2}{K^2} + \frac{1}{\sqrt{\beta_0}} \Bigg)^{-1} \,.
    \]
     The idea is to produce an effect similar to that of the Annealed Importance Sampling \citep{neal2001annealed}. The acceptance/rejection step is avoided as it is not amenable to the reparametrization trick \citep{salimans2015markov}. This creates a smooth invertible transformation $\mathcal{H}_x = g^K \circ \Phi_{\varepsilon, x}^{\circ(1)} \circ \cdots \circ g^0 \circ \Phi_{\varepsilon, x}^{\circ(1)}$ mapping ($z_0, \rho_0) \in \mathbb{R}^d \times \mathbb{R}^d$ to $(z_K, \rho_K) \in \mathbb{R}^d \times \mathbb{R}^d$ with $g^k$ being a tempering step. This transformation can be interpreted as a \textit{target-informed} normalizing flow since each integrator step $\Phi_{\varepsilon, x}^{\circ(1)}$ is guided by the gradient of true posterior distribution $p_{\theta}(z|x)$. Since each transformation is smooth and differentiable, the whole scheme is also amenable to the reparametrization trick so that we have access to an unbiased estimate of the gradient of the $ELBO$. Using the volume preservation property (i.e. $|\det \mathbf{J}_{\Phi_{\varepsilon, x}^{\circ(1)}} | = 1$)
    and Eq.~\eqref{Eq: Normailizing flows}, we have

    \begin{equation}\label{eq: HVAE equation}
        \begin{aligned}
            q_{\phi}(z_K, \rho_K|x) = q_{\phi}(z_0|x) q(\rho_0) \prod_{t=1}^K |\det \mathbf{J}_{g^k}{}|
             &= q_{\phi}(z_0|x) q(\rho_0) \underbrace{\prod_{k=1}^K \Big(\frac{\beta_{k-1}}{\beta_k}\Big)^{d/2}}_{\beta_0^{d/2}} \,.
        \end{aligned}
    \end{equation}
    In their work, the latent space had an Euclidean structure and they considered a fixed \textit{mass matrix} equals to $I_d$. This choice was motivated by the fact that optimizing the leapfrog step sizes is equivalent to optimizing the \textit{mass matrix} itself \citep{neal2011mcmc} provided that this matrix is diagonal. However, as a variant approach using a space-dependant \textit{mass matrix} and exploiting the manifold structure of probability densities could lead to far better and faster samplings \citep{girolami2009riemannian}, we do not see any apparent reason to restrict $\mathbf{M}$ to be constant and the latent space to be euclidean. This is what led us to introduce the Riemannian Hamiltonian VAE.

    \section{Proposed Method: Geometry-Aware Hamiltonian VAE}
In this section we introduce the Riemannian Hamiltonian VAE and describe and motivate the choice in the metric we use.
    \subsection{Riemannian Hamiltonian Markov Chain Monte Carlo}

    We will now assume that the latent variables $z$ live in a Riemannian manifold $\mathcal{Z}$ endowed with a Riemannian metric $\mathbf{G}$. It has been shown that an extension to Riemannian manifolds of the Hamiltonian Monte Carlo sampler is also possible \citep{girolami2009riemannian} and \citep{girolami2011riemann}. In such a context, the \textit{momentum} is such that $\rho \sim \mathcal{N}(0, \mathbf{G}(z))$ and so is no longer independent from $z$. Keeping the same notation as before and writing the negative logarithm of the extended joint distribution $\log p_{\theta}(x, z, \rho)$, the (Riemannian) \textit{Hamiltonian} follows
    
    \begin{equation}
        \label{eq: riemannian hamiltonian}
        H_x^{\mathnormal{Riem}}(z, \rho) = U_x(z) + \frac{1}{2} \log((2 \pi)^D \det \mathbf{G}(z)) + \frac{1}{2} \rho^{\top} \mathbf{G}(z)^{-1} \rho\,;
    \end{equation}
    
    such that the target distribution remains

    \[\begin{aligned} \pi_x(z) = \int \pi_x(z, \rho) d\rho = \frac{\int
        e^{-H_x(z, \rho)} d\rho}{\int
        e^{-H_x(z, \rho)}
        d\rho dz} =\frac{\frac{e^{-U_x(z)}}{(2 \pi)^{D/2} \sqrt{|\mathbf{G(z)}|}} \int
        e^{-\frac{1}{2} \rho^{\top} \mathbf{G}(z) \rho} d\rho}{\int
        \frac{e^{-U_x(z)}}{(2 \pi)^{D/2}\sqrt{ |\mathbf{G(z)|}}} \int e^{-\frac{1}{2}
        \rho^{\top} \mathbf{G}(z) \rho} d\rho dz}
        &= \frac{e^{-U_x(z)}}{\int
        e^{-U_x(z)} dz}\\
        &= \frac{p_{\theta}(x, z)}{\int p_{\theta}(x, z) dz} \\
        &= p_{\theta}(z|x)\,.
    \end{aligned}
        \]
    Considering now that we have a position-specific metric tensor $\mathbf{G}(z)$ defined on the manifold, the kinetic energy of Eq.~\eqref{eq: Euclidean Hamiltonian} writes
    \[
        \kappa(z, \rho) = \frac{1}{2} \log((2 \pi)^D \det \mathbf{G}(z)) + \frac{1}{2} \rho^{\top} \mathbf{G}(z)^{-1} \rho\,.
    \]
    Again differentiating Eq.~\eqref{eq: riemannian hamiltonian} with respect to $z$ and $\rho$ leads to a system of PDE \citep{girolami2011riemann} known as Hamilton's equations

    \begin{equation}\label{eq: PDE Riemann}\left\{\begin{aligned}
        \frac{\dd z_i}{\dd t} =\frac{\partial H_x^{\mathnormal{Riem}}}{\partial \rho_i} &= \big( \mathbf{G}^{-1}(z) \rho \big)_i \hspace{5mm} \,, \\
        \frac{\dd \rho_i}{\dd t} = - \frac{\partial H_x^{\mathnormal{Riem}}}{\partial z_i} &= \frac{\partial \log \pi_x(z)}{\partial z_i} - \frac{1}{2} \mathnormal{tr} \Biggl(\mathbf{G}^{-1} \frac{\partial \mathbf{G}(z)}{\partial z_i} \Biggr) + \frac{1}{2} \rho^{\top} \mathbf{G}^{-1}(z) \frac{\partial \mathbf{G}(z)}{\partial z_i} \mathbf{G}^{-1}(z) \rho \,.
    \end{aligned}   \right. 
    \end{equation}
    Unfortunately, the integrator proposed in Eq.~\eqref{Eq: Stormer-Verlet} is no longer volume preserving since the variable $\rho$ is no longer independent from $z$. Hence, a new integration scheme with the volume preserving and reversibility properties has been proposed and writes
    \begin{equation}\label{Eq: Riemann Stormer Verlet}
        \begin{aligned}
            \rho(t + \varepsilon/2) &= \rho(t) - \frac{\varepsilon}{2} \nabla_z H_x^{\mathnormal{Riem}}\Bigl(z(t), \rho(t+\varepsilon/2)\Bigr)\,,\\
            z(t + \varepsilon)      &= z(t) + \frac{\varepsilon}{2} \Bigl[\nabla_{\rho} H_x^{\mathnormal{Riem}}\Bigl(z(t), \rho(t + \varepsilon/2)\Bigr) + \nabla_{\rho} H_x^{\mathnormal{Riem}}\Bigl(z(t+\varepsilon),  \rho(t + \varepsilon/2)\Bigr)\Bigr]\,,\\
            \rho(t + \varepsilon)   &= \rho(t + \varepsilon/2) - \frac{\varepsilon}{2} \nabla_z H_x^{\mathnormal{Riem}} \Bigl(z(t+\varepsilon), \rho(t+ \varepsilon/2) \Bigr)\,.
        \end{aligned}
    \end{equation}
    This integrator is referred to as the \textit{generalized leapfrog integrator} and ensures that the target distribution is preserved by Hamiltonian dynamics. It has been shown by \citet{hairer2006geometric} and \citet{leimkuhler2004simulating} that this integrator is also volume preserving and time reversible. Again, if the acceptation/rejection ratio is added, the Riemannian Hamiltonian Monte Carlo sampler (RHMC) produces an ergodic, time-reversible Markov Chain having $\pi_x$ as stationary distribution \citep{girolami2011riemann,duane1987hybrid,neal2011mcmc, liu2008monte, neal2012bayesian}.

    We propose an approach similar to the one discussed in \citep{caterini2018hamiltonian} but taking into account the non-Euclidean structure of the latent space. In our method, $\mathcal{Z}$ is assumed to be a Riemannian space whose metric is given by $\mathbf{G}(z)$. This makes us use the \textit{generalized leapfrog integrator} along with a tempering step to create a smooth mapping $\mathcal{H}^{\mathnormal{Riemann}}_x$ that takes $(\rho_0, z_0) \in \mathbb{R}^d \times \mathbb{R}^d$ and returns $(\rho_K, z_K)$. Again this transformation $\mathcal{H}^{\mathnormal{Riemann}}_x$ can be seen as a specific kind of normalizing flow informed by the target through Eq.~\eqref{eq: PDE Riemann} and by the latent space geometry thanks to the metric $\mathbf{G}$. Our intuition is that using the underlying geometry of the manifold in which the latent variables live would better guide the approximate posterior distribution leading to better Log-Likelihood (LL) estimate and will also structure this space. One may remark that the \textit{generalized leapfrog integrator} is no longer explicit and so requires the use of fixed point iterations to be solved. Fortunately, only few iterations are needed to stabilize the scheme (we use 3 iterations). While these fixed point iterations add some computation time to the training process when compared to the Hamiltonian VAE, this is counter-balanced by the more efficient sampling achieved with the RHMC \citep{girolami2009riemannian} which basically requires a fewer number of leapfrog iterations (3 vs. 10/15) to sample "accurately". We refer the reader to Section~\ref{Sec: Auto-Encoding comparision} for quantitative metrics comparison. Finally, using Eq.~\eqref{Eq: Normailizing flows} and the volume preservation leads to the same kind of equation as Eq.~\eqref{eq: HVAE equation} that is
    
\[
    q_{\phi}(z_K, \rho_K|x) = q_{\phi}(z_0|x) q(\rho_0|z_0) \prod_{t=1}^K |\det \mathbf{J}_{g^k}{}|
    = q_{\phi}(z_0|x) q(\rho_0|z_0) \prod_{k=1}^K \Big(\frac{\beta_{k-1}}{\beta_k}\Big)^{d/2} \,.
\]
    The major difference with the Hamiltonian variational auto-encoder is that we propose to sample $\rho$ using a position-specific distribution. Again, omitting the acceptation/rejection step makes the flow $\mathcal{H}^{\mathnormal{Riemann}}_x$ differentiable with respect to $\phi$ and so the reparametrization trick can be used and gives access to an unbiased estimate of the gradient of the $ELBO$.

\subsection{The Metric}
Since the choice of the metric appears to be quite crucial, we first propose to discuss some Riemannian metrics that have been exposed in the literature before introducing the one we propose. 
\subsubsection{Metric Proposed in the Literature}

A quite "natural" way to introduce a Riemannian structure in the latent space of deep generative models is to consider a metric deriving from Taylor's theorem. The idea is to consider $z \in \mathcal{Z}$, $\Delta z$ a small variation around $z$ and $f :z \in \mathcal{Z} \to f(z) \in \mathcal{X}$ the generator function. Taking the square norm between two decoded samples gives

\[
    \lVert f(z + \Delta z) - f(z) \rVert^{2} \approx  \Delta z^{\top} \mathbf{J}_z^{\top} \mathbf{J}_z \Delta z\,, 
\]
where $\mathbf{J}_z = \frac{\partial f}{\partial z}$. $\mathbf{J}_z^{\top} \mathbf{J}_z$ can now be seen as a Riemannian metric in the latent space. This modelling has been a common point in many papers trying to bring geometry to the latent space of deep generative models. While \citet{chen2018metrics} and \citet{shao2018riemannian} directly used the metric $\mathbf{M}_z = \mathbf{J}_z^{\top} \mathbf{J}_z$,  \citet{arvanitidis2017latent, yang2018geodesic} and \citet{hauberg2018only}  went a bit further and considered a \textit{stochastic} metric. In their papers, the authors considered the $\mathcal{N}$-VAE meaning that $p_{\theta}(x|z)$ is modeled by a Gaussian distributions
$\mathcal{N}(\mu_{\theta}(z), \Sigma_{\theta}(z))$ where $\Sigma_{\theta} = \sigma_{\theta}(z) I_D$ is diagonal. In such a context a data point $x_g$ can be generated using the reparametrization trick as follows:
\[
    x_g = \mu_{\theta} + \sigma_{\theta} \odot \varepsilon, \hspace{5mm}  \varepsilon \sim \mathcal{N}(0, I_D)\,,
\]
where $\odot$ is the element-wise product. With that being said, the generator function is now stochastic and so the authors demonstrated that if the mean function $\mu_{\theta}$ and the variance function $\sigma_{\theta}$ are twice differentiable, the expected value of the metric $\mathbf{M}_z = \mathbf{J}_z^{\top} \mathbf{J}_z^{\top}$ writes

\begin{equation}\label{Eq: Arvan metric}
    \mathbb{E}_{\varepsilon}[\mathbf{M}_z] = \Bigl(\mathbf{J}^{(\mu)}_z \Bigr)^{\top}\Bigl(\mathbf{J}^{(\mu)}_z \Bigr) + \Bigl(\mathbf{J}^{(\sigma)}_z \Bigr)^{\top} \Bigl(\mathbf{J}^{(\sigma)}_z \Bigr)\,.
\end{equation}
They used this equation as an approximation of the "true" underlying metric $\mathbf{M}_z$ on the ground that Var $(\mathbf{M}_z) \xrightarrow[D \to \infty]{}0$. An interesting aspect of Eq.~\eqref{Eq: Arvan metric} is that it involves directly the variance function. Intuitively, we would expect the metric to have high values in locations with high uncertainty that is where no data is available. Hence, the geodesics would stay close to the data. Unfortunately, arguing that neural networks interpolate badly in uncertain regions, they paid particular attention to the modelling of the variance function. Their idea consisted in proposing a variance function such that it achieves, as expected, higher values far from the data impeding geodesic paths to explore these regions. This led \citet{arvanitidis2017latent} to consider the following modelling:

\[\begin{aligned}
    \frac{1}{\sigma_{\psi}^{2}(z)} = W v(z) + \xi, \hspace{5mm} \text{with } v_k(z) = \exp(- \lambda_k \lVert z - c_k \rVert ^2)\,,
\end{aligned}
\] where $(c_k)_{1 \leq k \leq K}$ are $K$ centroids obtained using $k$-means algorithm on the encoded samples, $W$ is a matrix of weights and $\lambda_k$ writes

\begin{equation} \label{eq: arvanitidis centroids}
  \lambda_k = \frac{1}{2} \Biggl( a \frac{1}{|\mathcal{C}_k|} \sum \limits_{z_j \in \mathcal{C}_k} \lVert z_j - c_k \rVert \Biggr)^{-2}\,.  
\end{equation}
The VAE is then trained in two times: 1) The mean $\mu_{\theta}$ of the generator function along with the mean $\mu_{\phi}$ and variance $\Sigma_{\phi}$ functions of the inference networks are trained while keeping $\sigma_{\theta}$ fixed; 2) The variance function $\sigma_{\psi}$ is trained with all other parameters fixed.

One drawback of the metrics involving the Jacobian of the generator function is that they strongly constraint the model used which rigorously needs to be at least $\mathcal{C}^2$ since the Riemannian metric must be smooth enough. This is made impossible if non-smooth activation functions such a $ReLu$ are used. Moreover, most of the time there is no \textit{closed-form} expression of the Jacobian available and it needs to be approximated using finite differences \citep[see][]{shao2018riemannian} adding potentially large biases or with automatic differentiation which can reveal very costly for deep networks.

\subsubsection{Proposed Latent Space Modelling}
    
    As highlighted in Eq.~\eqref{eq: riemannian hamiltonian}, the choice of the metric tensor $\mathbf{G}$ is crucial since it defines the topology of the latent space. While the previous section illustrated some candidate metrics, we take a rather different approach as we decide to learn a parametrized metric directly from the data using a neural network. The metric model we propose is a generalization of the one exposed in \citep{louis2019computational}. Note that, in that paper, the author assumes that the data live in a Riemannian manifold and the latent space is Euclidean whereas we do not make any assumption on the data space and assume a Riemannian structure of the latent space. We parametrize the inverse of the metric tensor rather than the metric itself since Hamiltonian dynamics only require the inverse of the metric tensor $\mathbf{G}^{-1}(z)$ and its determinant $\det \mathbf{G}(z)$ to be computed (see Eq.~\eqref{eq: riemannian hamiltonian}). This implies that we do not have to inverse the metric tensor at each leapfrog step in Eq.~\eqref{Eq: Riemann Stormer Verlet}. Our parametrization writes

    \begin{equation}\label{eq: metric}
        \mathbf{G}^{-1}(z) = \sum_{i=1}^N L_{\psi_i} L_{\psi_i}^{\top} \exp \Big(-\frac{\lVert z - c_i \rVert_2^2}{T^2} \Big) + \lambda I_d \,,
    \end{equation}
    where $L_{\psi_i}$ are lower triangular matrices with positive diagonal coefficients. $T$ is a temperature to smooth the metric and $\lambda$ a regularization factor. $c_i$ are referred to as the \textit{centroids} and are such that $c_i = \mu(x_i)$ where $\mu(x_i)$ is the mean of the density distribution of the latent variable $z_i \sim \mathcal{N}(\mu(x_i), \Sigma(x_i)) = q_{\phi}(z_i|x_i)$ associated to the data point $x_i$. $L_{\psi_i}$ can intuitively be seen as the triangular matrix in the Cholesky decomposition of $\mathbf{G}^{-1}(c_i)$ up to a regularization factor. The $L_{\psi_i}$ are learned using a neural network $m_{\psi}$ mapping a data point $x_i$ from the training set to a lower triangular matrix $L_{\psi_i}$.  The \textit{hyper-parameters} $T$ and $\lambda$ can be learned or kept fixed. The influence of each of these parameters is discussed in Section~\ref{Sec: Sensitivities}. At the end, the centroids $c_i$, the matrices $L_{\psi_i}$ along with the temperature $T$ and regularizing factor $\lambda$ are fixed and stored.

    We found this metric very interesting as it demonstrates very powerful properties. First, the metric is smooth and even $\mathcal{C}^{\infty}$ which allows for an easier usage. Second, it is easy to evaluate its value at any given point $z$ of the latent space since it does not require the computation of a potentially time-consuming function such as the
    Jacobian. The proposed parametrization can be easily integrated in the learning process as described in Algorithm \ref{RHVAE}. Even though by design the proposed metric scales in memory with the number of training points and the dimension of the latent space one can easily reduce the number of centroids by electing $k$ clusters centers using $k$-means or $k$-medoids algorithm amongst the actual centroids $c_i$. These centers are then used as references points during training.

    \begin{algorithm}[p]
    \SetAlgoLined \textbf{Initialize} $\mathbf{G}$ \tcp*{We put $c_i = 0$ and
     $L_{\psi_i}=I_d$} \While{not converged}{$\mathcal{L} \leftarrow 0$ \;
     \For{$n = 1 \to N_B$}{Collect a batch of data $X_n = (x_1, \cdots,
     x_{\text{bs}})$\; $c_i \leftarrow \text{encode}(x_i)$\; $L_{\psi_i}
     \leftarrow m_{\psi}(x_i) $\; Update the metric $\mathbf{G}$ according to
     Eq.~\eqref{eq: metric}\; $z_0 \sim \mathcal{N}(\mu(x), \Sigma(x)),$ $\rho_0
     \sim \mathcal{N}(0, \mathbf{G}(z_0))$\; $\rho \leftarrow \rho_0 /
     \sqrt{\beta_0}$\; \For{$k = 1 \to K$}{$\bar{\rho} \leftarrow \rho_{k-1} -
     \frac{\varepsilon}{2} \nabla_{z} H(x, z_{k-1}, \bar{\rho})$ \tcp*{fixed
     point it.} $z_k \leftarrow z_{k-1} + \frac{\varepsilon}{2}
     \Big(\nabla_{\rho} H (x, z_{k-1}, \bar{\rho}) + \nabla_{\rho} H(x, z_k,
     \bar{\rho}) \Big)$ \tcp*{fixed point it.} $\rho' \leftarrow \bar{\rho}
     - \frac{\varepsilon}{2} \nabla_{z} H(x, z_k, \bar{\rho})$\; $\sqrt{\beta_k}
     \leftarrow \Big(\Big( 1 - \frac{1}{\sqrt{\beta_0}}\Big) \frac{k^2}{K^2}+
     \frac{1}{\sqrt{\beta_0}} \Big)^{-1}$ \; $\rho_k \leftarrow
     \frac{\sqrt{\beta_{k-1}}}{\sqrt{\beta_{k}}}\rho'$ \;} $p \leftarrow
     p_{\theta}(x, z_K, \rho_K)$ \; $q \leftarrow q_{\phi}(z_0, \rho_0|x)$\;
     $\mathcal{L}_{\mathrm{batch}} \leftarrow \log p - \log q$ \; $\mathcal{L} =
     \mathcal{L} + \mathcal{L}_{\mathrm{batch}} / N_B$ \;} Update $\theta$,
     $\phi$ and $\psi$ using gradient descent\;}
     \caption{RHVAE with metric learning}
     \label{RHVAE}
    \end{algorithm}

    \section{Experiments}
    In this section, we propose to empirically assess the proposed model's enhancements in terms of Log-Likelihood estimate, reconstruction error, samples interpolation, generation and clustering. 
    
    \subsection{Models Architectures}

    \begin{table}[p]
    \centering
    
    \begin{tabular}{c|l|l}
        \toprule
            Networks   & \multicolumn{2}{c}{Configurations}\\
            \hline                               
        $\mu_{\varphi}$ &  MLP - $(D, 400, ReLu)^{*}$\tnote{*}    & MLP - $(400, d, Linear)$ \\
        $\Sigma_{\varphi} $ & MLP - $(D, 400, ReLu)^{*}$ & MLP - $(400, d, Linear)$ \\                                                 
        $\pi_{\theta}$      & MLP - $(d, 400, ReLu)$  & MLP - $(400, D, Sigmoid)$   \\
        $L_{\psi} $ (diag)  & MLP - $(D, 150, ReLu)^{**}$& MLP - $(150, d, Linear)$ \\
        $L_{\psi} $ (lower) & MLP - $(D, 150, ReLu)^{**}$& MLP - $(150, \frac{d(d-1)}{2}, Linear)$\\
        \bottomrule
    \end{tabular}
    \begin{tablenotes}[*]\footnotesize
        \item[*] * Same layers, ** Same layers
        \end{tablenotes}
        \caption{Inference and generator neural networks used for the VAE, HVAE and RHVAE along with the neural network architecture used for metric learning.}
        \label{Table: Model architectures}
\end{table}
    For each experiment we consider a $\mathcal{B}$-VAE with the architectures as described in Table~\ref{Table: Model architectures} unless stated otherwise. The metric used within the RHVAE is given by Eq.~\eqref{eq: metric}. We recall that the $\mathcal{B}$-VAE framework is as follows:
    \[
    \left\{
        \begin{aligned}
        z &\sim  \mathcal{N}(0, I_d) \,,\\
        x| z &\sim p_{\theta}(x|z) = \prod_{i=1}^D \mathcal{B}(x_i | \pi_{\theta}(z)_i)\,, \\
        z | x &\sim q_{\phi}(z|x) = \mathcal{N}(\mu_{\phi}(x), \Sigma_{\phi}(x))\,.
        \end{aligned}
        \right.
    \]

    \subsection{Auto-Encoder}
First of all, we test the auto-encoding ability of the proposed model and compare it to other VAE architectures.
    \subsubsection{Comparison with Peers}\label{Sec: Auto-Encoding comparision}
    Although enhancing the Log-Likelihood estimate is not our primary objective when adding a Riemannian metric in the latent space, we nonetheless try to see if it does improve it on relatively \textit{small} size data sets extracted from two well-known databases. To do so, the Log-Likelihood values along with the reconstruction errors obtained with a RHVAE are compared to the ones of a vanilla VAE and several Hamiltonian VAEs trained with different sets of parameters as proposed in \citep{caterini2018hamiltonian}. We consider 2 data sets respectively extracted from the MNIST \citep*{lecun1998mnist} and the FashionMNIST \citep{xiao2017online} data sets. To stick to the \textit{small} data set framework, we decide to only select 50 random samples from each class of the group \{"0", "1", "2"\} (resp. \{"T-shirt", "Sandal", "Bag"\}) of the MNIST (resp. FashionMNIST) data set. Then, the created sets are split into a training set ($80\%$ of the data set) and a test set ($20 \%$) ensuring balanced classes. For each model the latent space dimension is set to 10 and we employ an {\it early-stopping} strategy consisting in stopping the training if the $ELBO$ does not improve on the validation set for 100 epochs. The Log-likelihood is evaluated using 200 importance samples from the approximate posterior distribution $q_{\phi}(z|x)$ and is estimated 5 times. We present the mean value across these 5 estimates along with the associated standard deviation between parenthesis in Table~\ref{Table: model comparison reconstruction}. Interestingly, the model we propose is able to outperform both the VAE and HVAE models on each data set. Although, a smaller number of leapfrog steps is considered when compared to the best HVAE, the proposed model still achieves a slightly better log-likelihood estimate than peers on the FashionMNIST data set (271.45 vs. 271.67). This is even more sticking on the MNIST data set where the proposed RHVAE outperforms competitors as well (110.60 vs. 112.28).
    
\begin{figure}[p]
    \centering
    \begin{minipage}[c]{0.48\linewidth}
        \centering
         \includegraphics[scale=0.20]{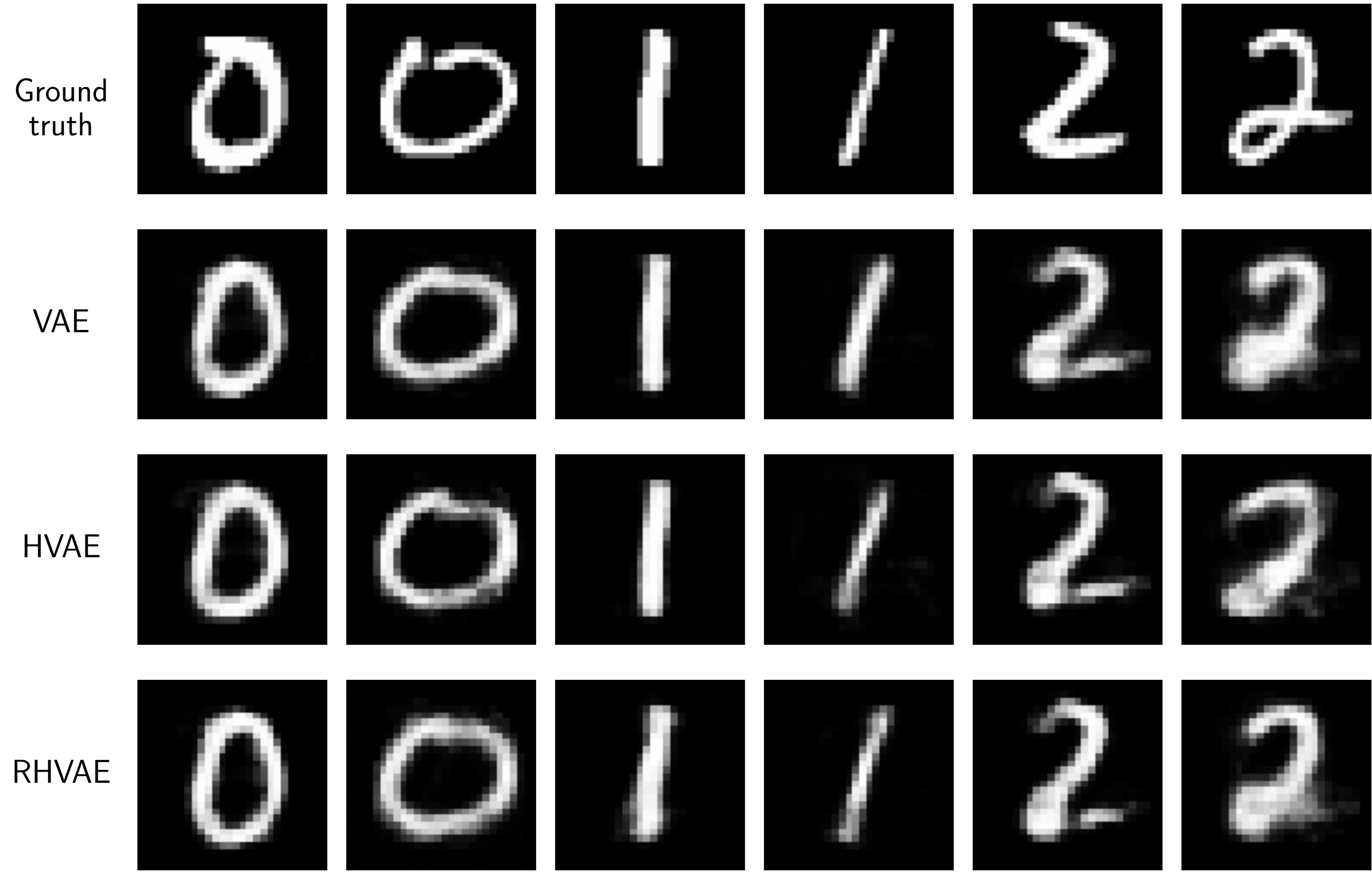}
    \end{minipage}
    \begin{minipage}[c]{0.48\linewidth}
        \centering
         \includegraphics[scale=0.20]{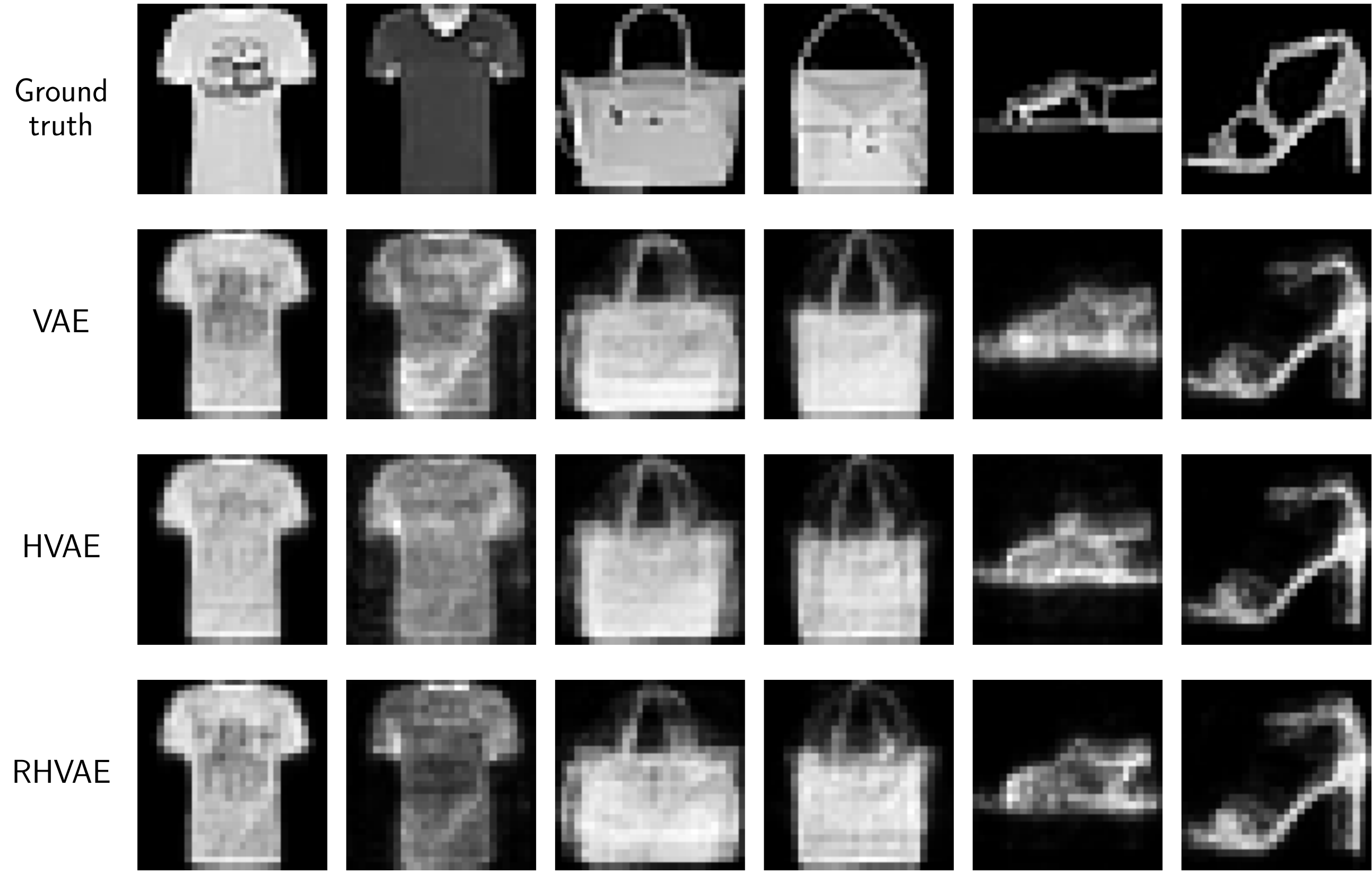}
    \end{minipage}
    \caption{Reconstruction of 2 samples per class extracted from the test set. To ensure a fair comparison only the models achieving the best test $ELBO$ are considered.}
    \label{fig: Model comparison reconstructed samples}
\end{figure}

    In addition to the Log-likelihood estimate, it is interesting to compute another metric assessing the reconstruction faithfulness. Even though we acknowledge that assessing the distance between images may reveal challenging we propose to use the $L$-$2$ norm to quantify the quality of the reconstructed samples for each model. As any of the pixels of the image is considered independent from the others, we believe that such a metric still provides a fairly good assessment of how "far" the reconstructed distribution is from the target. To ensure a fair comparison between models, we use the model achieving the best test $ELBO$ on the validation set. The models' ability to reconstruct samples faithfully is then assessed by computing the relative $L$-2 distance between the ground truth images of the test and train sets and the reconstructed samples. The results are made available in Table~\ref{Table: reconstruction error}. As expected since it achieves a strongly better Log-likelihood estimate on the MNIST data set, the RHVAE outperforms other models in terms of pure reconstruction on the test set. Interestingly it also performs the "worst" on the training set which is a good indicator that compared to other models it does not over-fit the training data. Although the Log-likelihood estimate it achieves on the FashionMNIST data set only slightly outperforms both the VAE and HVAEs, the proposed RHVAE strongly outperforms competitors in terms of reconstruction on both the training and testing set. In Figure~\ref{fig: Model comparison reconstructed samples} reconstructed samples extracted from the test set are also presented.
    
    As to parameters setting, we use a batch size set to $60$. The temperature $\beta_0$ is learned for HVAE and fixed to 0.3 for our RHVAE since we consider that the parameter $\beta_0$ can be "learned" directly within the metric which becomes now {\it position-specific}. For the MNIST database, we learn the metric temperature and the leapfrog integrator step size $\varepsilon_{\mathnormal{lf}}$, the regularization is set to  $10^{-3}$ while we use a regularization of $10^{-2}$ along with a fixed $\varepsilon_{\mathnormal{lf}}$ set to $10^{-2}$ for the FashionMNIST data set. Hyper-parameters influence is discussed in the following section.

\begin{table}[p]
    \centering
    \begin{tabular}{c|c|c||c|c}
        \toprule
        \multicolumn{3}{c||}{Models}                                 & MNIST & FashionMNIST \\
        Name &\multicolumn{2}{c||}{Parameters}                       &                                 &       \\ 
         &  $n_{\mathnormal{lf}}$ & $\varepsilon_{\mathnormal{lf}}$  &  $\log p(x)$                    & $\log p(x)$ \\
        \hline                           
        VAE                       & -  & -                           &  $-113.48~ (  0.21)$          & $-275.35~ (  0.63)$   \\
        \hline
        HVAE                      & 1  & \textit{learned}            &  $ -113.44~(  0.33)$          & $-273.96~ (  0.52)$ \\  
        HVAE                      & 3  & \textit{learned}            &  $ -115.05~(  0.34)$          & $-272.40~ (  0.30)$           \\
        HVAE                      & 5  & \textit{learned}            &  $ -113.78~(  0.47)$          & $-271.67~ (  0.35)$           \\
        HVAE                      & 10 & \textit{learned}            &  $ -112.97~(  0.47)$          & $-271.96~ (  0.28)$           \\
        HVAE                      & 15 & \textit{learned}            &  $ -112.07~(  0.33)$          & $-272.64~ (  0.20)$           \\
        HVAE                      & 10 & $10^{-4}$                   &  $ -113.40~(  0.21)$          & $-274.35~ (  0.19)$           \\
        HVAE                      & 10 & $10^{-3}$                   &  $ -112.43~(  0.13)$          & $-275.32~ (  0.44)$           \\
        HVAE                      & 10 & $10^{-2}$                   &  $ -112.28~(  0.33)$          & $-274.16~ (  0.26)$           \\
        \hline           
        RHVAE                     & 3  &         \textit{learned}$ / 10^{-2}$           &  $ \mathbf{-110.60~(  0.17)}$ & $\mathbf{-271.45~(  0.32)}$    \\
        \bottomrule
    \end{tabular} 
    \caption{Maximum Log-Likelihood estimate achieved by each model along with the main parameters values. The models are trained on 2 \textit{small} data sets extracted from MNINST and FashionMNIST. The training set is created by randomly selecting 80\% of a data set composed by 3 classes of 50 samples each and ensuring balanced classes. Training is stopped if the $ELBO$ does not improve on the validation set (20\% of the initial data set) for 100 epochs. }
    \label{Table: model comparison reconstruction}
\end{table}

\begin{table}[p]
    \centering
    \begin{tabular}{c||c|c|c|c}
        \toprule
        Model           & \multicolumn{2}{c|}{MNIST}                 & \multicolumn{2}{c}{FashionMNIST} \\
                        & Train            &       Test            &    Train               & Test \\       
        \hline
        VAE             & $\mathbf{9.35\%}$           &    $26.19\%$              &    $18.81\%$           &  $10.86\%$    \\
        HVAE            & $10.05\%$                    &    $25.41\%$              &    $18.22\%$           &  $10.14\%$    \\
        RHVAE           & $10.97\%$                    &    $ \mathbf{24.88\%}$    &    $\mathbf{16.96\%}$  &  $\mathbf{9.66\%}   $    \\
        \bottomrule
    \end{tabular}
    \caption{Relative $L$-$2$ reconstruction error on the test and training sets for each model ($\varepsilon_{\mathnormal{err}} = \frac{\sum_{i} \lVert x_i - x_i^{\mathnormal{reconstructed}} \rVert_2^{2}}{\sum_i \lVert x_i \rVert_2^2}$). To ensure a fair comparison only the models achieving the best test $ELBO$ are considered.}
    \label{Table: reconstruction error}
    
\end{table}


    \subsubsection{Sensitivites} \label{Sec: Sensitivities}

    We acknowledge that the model we introduce comprises several hyper-parameters the influence of which is discussed in this section. We run our algorithm and compute several sensitivities on the same data sets as those used in the previous section. In particular, the RHVAE models are trained with different metric temperatures $T$ either learned or set fixed and ranging from $0.4$ to $5$, a fixed regularization factor $\lambda$ ranging from $10^{-3}$ to $10$, a fixed number of leapfrog steps $n_{\mathnormal{lf}}$ ranging from $1$ to $10$ and different leapfrog step sizes either learned or fixed and ranging from $10^{-2}$ to $10^{-4}$. We decide not to change the value of $\sqrt{\beta_0}$ since we believe its value is already optimized within the learned {\it position-specific} metric. The metrics (Log-likelihood and minimum $ELBO$) are reported for all the models and are presented in Table~\ref{Table: Sensitivities metrics}. For the sake of readability, we also provide the training curves in Appendix~\ref{app: Training curves} presenting the moving average (10 epochs) of both metrics. The same {\it early-stopping} as in Section~\ref{Sec: Auto-Encoding comparision} is employed.  For all sensitivities, the latent space dimension is set to 2.
    
    \begin{table}[ht]
    \centering
    \begin{tabular}{c|c|c|c||c|c||c|c}
        \toprule
             \multicolumn{4}{c||}{Models}                                  & \multicolumn{2}{c||}{MNIST}          & \multicolumn{2}{c}{FashionMNIST} \\
                     \multicolumn{4}{c||}{Parameters}                      &                            &                 &               \\
             $n_{\mathnormal{lf}}$ & $\varepsilon_{\mathnormal{lf}}$ & $T$              & $\lambda$          & $\log p(x)$                & $ ELBO$                   & $\log p(x)$      & $ELBO$\\
        \hline
               1                   &  $10^{-2}$                                & 0.8              & $10^{-2}$          &     $-137.52~ (  0.26)$                        &  4277.19                  &   $-292.89~ (  0.33)$          &  9132.64          \\
               3                   &  $10^{-2}$                                & 0.8              & $10^{-2}$          &     $-136.76~ (  0.30)$                        &  4236.07                  &   $-289.65~ (  0.13)$          &  9115.90          \\ 
               10                  &  $10^{-2}$                                & 0.8              & $10^{-2}$          &     $-137.63~ (  0.07)$                        &  4263.47                  &   $-294.79~ (  0.12)$          &  9216.46          \\
        \hline    
               5                   &  $10^{-4}$                                & 0.8              & $10^{-2}$          &     $-137.44~ (  0.12)$                        &  4246.53                  &   $-288.96~ (  0.17)$          &  8945.68 \\ 
               5                   &  $10^{-3}$                                & 0.8              & $10^{-2}$          &     $-134.50~ (  0.17)$                        &  4115.16                  &   $-285.53~ (  0.13)$          &  8827.40        \\ 
               5                   &  $ \textit{l.}^{*}$                       & 0.8              & $10^{-2}$          &     $-137.95~ (  0.23)$                        &  4250.48                  &   $-291.84~ (  0.51)$          &  9078.35        \\ 
       \hline
               5                   &  $10^{-2}$                                         & 0.4              & $10^{-2}$          &  $-137.82~ (  0.15)$                   &         4284.54            & $-288.17~ (  0.51)$            &     8918.95       \\
               5                   &  $10^{-2}$                                         & 0.6              & $10^{-2}$          &  $-136.47~ (  0.09)$                   &         4239.84            & $-288.39~ (  0.23)$            &      8985.83      \\
               5                   &  $10^{-2}$                                         & 1                & $10^{-2}$          &  $-136.33~ (  0.24)$                   &         4326.44            & $-287.93~ (  0.25)$            &      8942.79      \\
               5                   &  $10^{-2}$                                         & 2                & $10^{-2}$          &  $-136.11~ (  0.24)$                   &         4246.98            & $-285.77~ (  0.18)$            &      8841.37      \\
               5                   &  $10^{-2}$                                         & 5                & $10^{-2}$          &  $-137.22~ (  0.14)$                   &         4230.40             & $-285.47~ (  0.11)$            &      8770.49      \\
               5                   &  $10^{-2}$                                         & $\textit{l.}^{*}$ & $10^{-2}$          &  $-135.53~ (  0.10)$                   &         4160.34            & $-283.84~ (  0.07)$            &      8732.34      \\
        \hline
               5                   &  $10^{-2}$                               & 0.8              & $10^{-3}$          &     $-136.87~ (  0.25)$                 &        4228.59             &  $-285.68~ (  0.29)$           &    8904.92        \\
               5                   &  $10^{-2}$                               & 0.8              & $10^{-1}$          &     $-137.11~ (  0.14)$                 &        4234.00             &  $-286.53~ (  0.30)$           &     8959.55       \\
               5                   &  $10^{-2}$                               & 0.8              & $1$                &     $-136.83~ (  0.13)$                 &        4252.50             &  $-285.90~ (  0.17)$           &      8946.34      \\
               5                   &  $10^{-2}$                               & 0.8              & $10$                &    $-136.12~ (  0.09)$                 &        4172.34             &  $-287.47~ (  0.10)$           &      8857.29      \\
        \hline
               \textbf{5}          & $\mathbf{10^{-2}} $ & \textbf{0.8}     & {$\mathbf{10^{-2}}$} & $ \mathbf{-135.88~(  0.07)}$  &   $\mathbf{4246.90}$            &  $ \mathbf{-287.49~(  0.34)}$  & $\mathbf{9015.70} $            \\

               \bottomrule
    \end{tabular}
    \begin{tablenotes}[*]\footnotesize
        \item[*] * learned
        \end{tablenotes}
    \caption{Hyper-parameters sensitivities. Maximum Log-Likelihood and minimum $ELBO$ achieved by RHVAEs trained on two \textit{small} data sets extracted from MNIST and FashionMNIST with different sets of parameters. The latent space dimension is set to 2.  Training is stopped if the $ELBO$ does not improve on the validation set (20\% of the initial data set) for 100 epochs. }
    \label{Table: Sensitivities metrics}
\end{table}
    
    The first outcome of such a study is that the parameters which seem to have the greatest impact on the overall model's performance are those directly linked to the generalized leapfrog integrator (i.e. $n_{\mathnormal{lf}}$ and $\varepsilon_{\mathnormal{lf}}$). The Log-Likelihood can indeed jump from 287.49 ($n_{\mathnormal{lf}} = 5$) to 292.89 ($n_{\mathnormal{lf}} = 1$) on the FashionMNIST data set for example. The leapfrog step size has a strong influence as well and interestingly trying to learn this parameter may not always improve the model (e.g. 137.95 vs. 134.50 for $\varepsilon_{\mathnormal{lf}} = 10^{-3}$ on the MNIST data set). This may be due to the fact that the metric is learned at the same time and so the learning of the leapfrog step size might be somehow integrated within the metric learning process. 
    Secondly, the model seems to remain quite robust to metric's hyper-parameters change which is good news since it will allow us to model the latent space as desired without completely degrading the model. Some findings are nevertheless interesting to discuss. First, setting a small temperature does not reveal to improve the model and could even induce instability in the training (for very small temperatures). Even though learning the temperature results in improved metrics, this choice remains dependant on the usage of the model. Actually, it may be of interest to fix the temperature to better apprehend the proposed modelling. The regularization factor $\lambda$ seems to have a weaker influence on the model since the Log-likelihood ranges from 135.88 ($\lambda = 10^{-2}$) to 137.11 ($\lambda = 10^{-1}$) on MNIST. Nonetheless, the value of this parameter may be of interest for interpolation or clustering since it strongly influences geodesic distances as discussed in Section~\ref{Sec: Metric Computation}.

    \subsection{On Geometrical Aspects}

    In this section, the geometrical aspects of the proposed metric are discussed and illustrated through various experiments.
    \subsubsection{Metric Computation} \label{Sec: Metric Computation}

\begin{figure}[ht]
    \centering
    \begin{minipage}[c]{0.32\linewidth}
        \centering
         \includegraphics[scale=0.38]{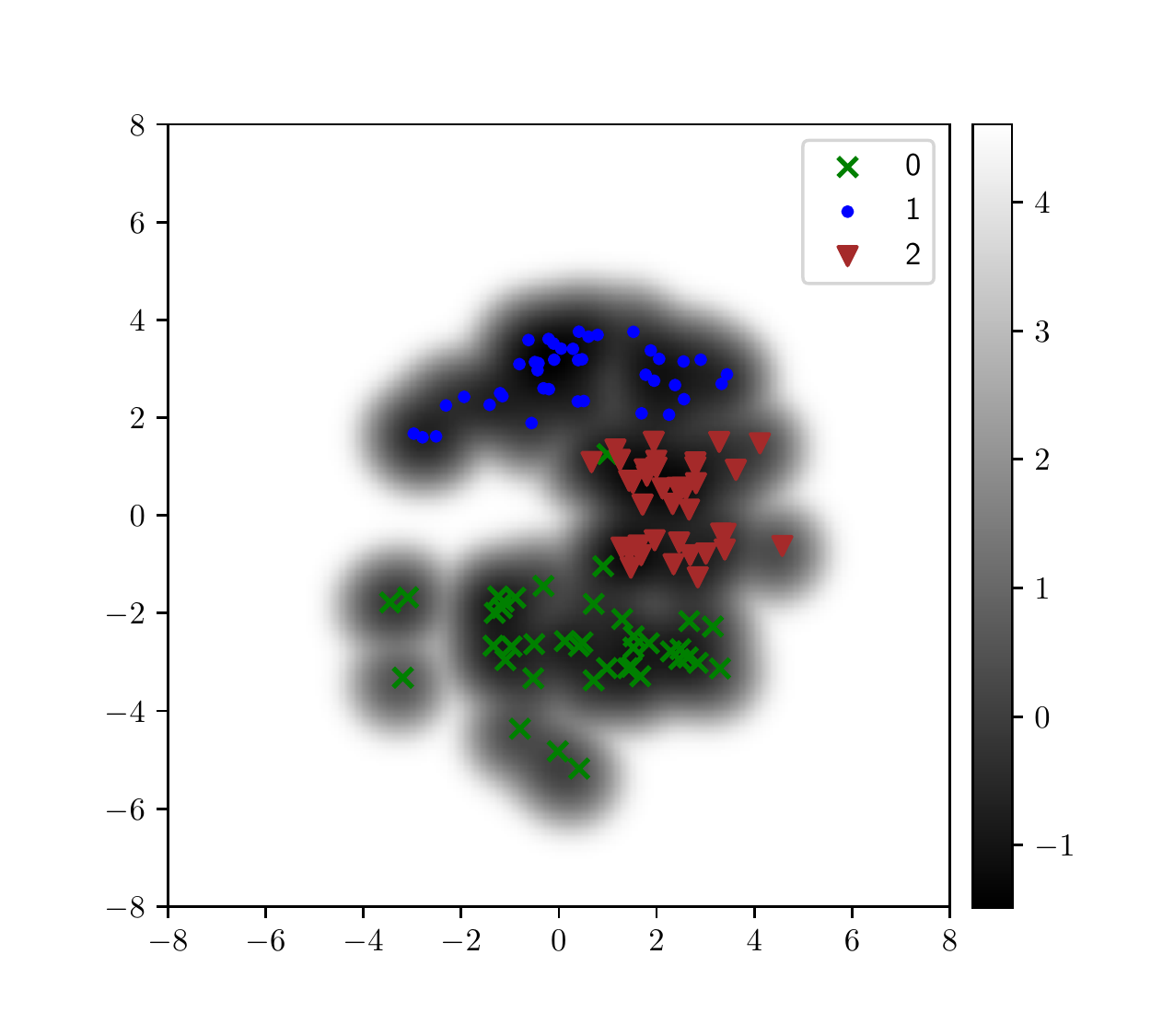}
    \end{minipage}
    \begin{minipage}[c]{0.32\linewidth}
        \centering
         \includegraphics[scale=0.38]{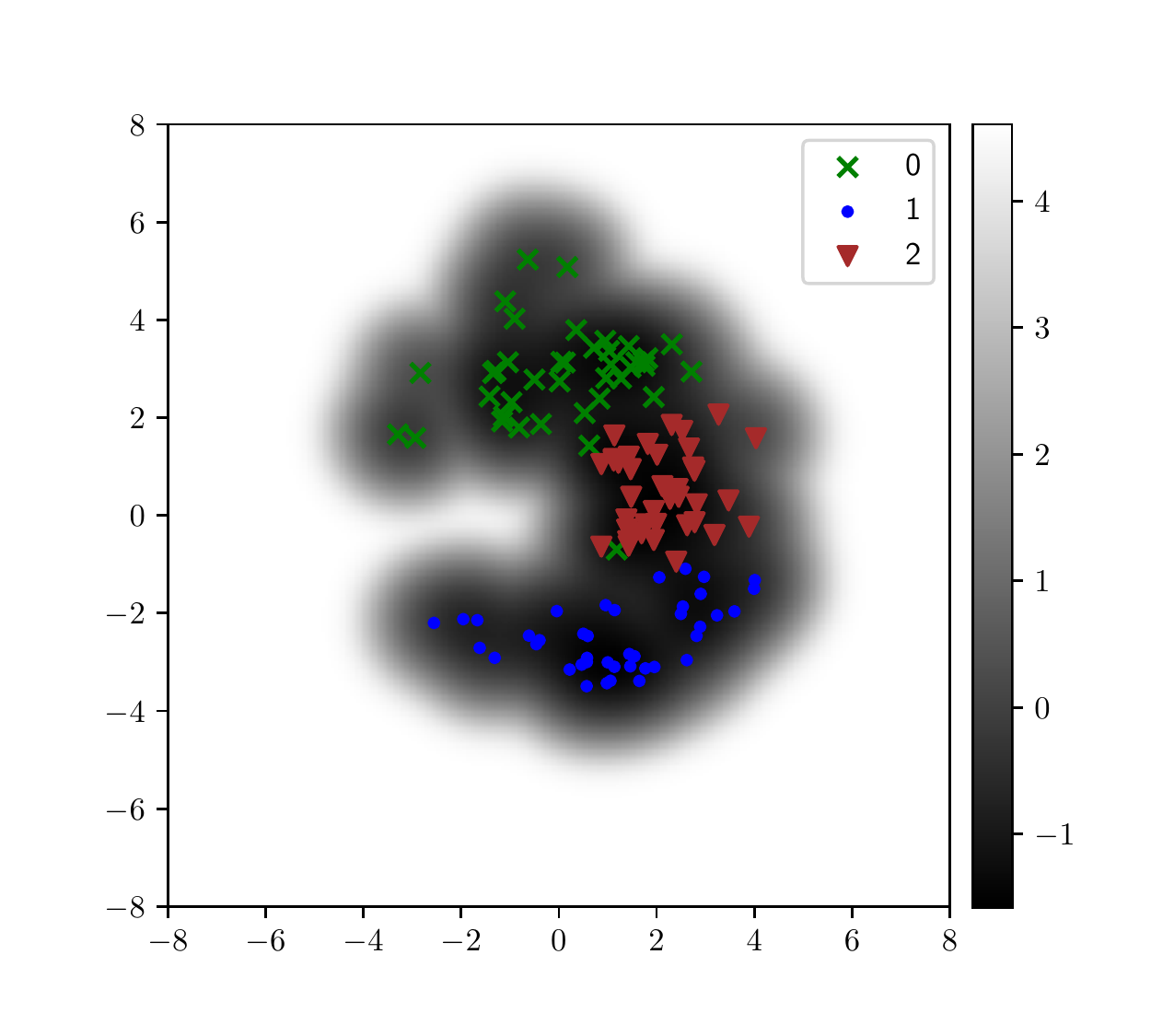}
    \end{minipage}
    \begin{minipage}[c]{0.32\linewidth}
         \centering
         \includegraphics[scale=0.38]{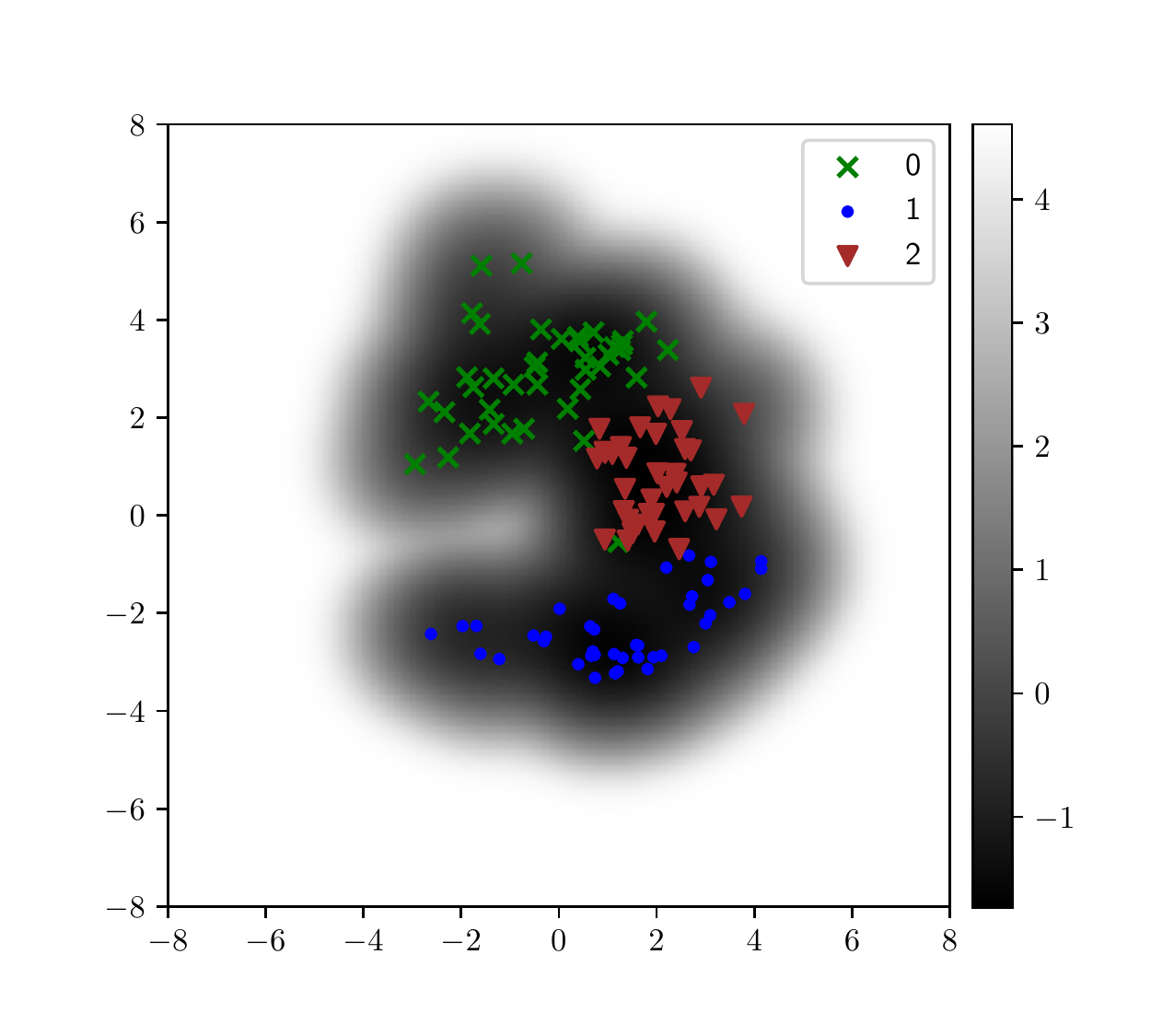}
    \end{minipage}
    \centering
    \vskip -1.em
    \begin{minipage}[c]{0.32\linewidth}
        \centering
         \includegraphics[scale=0.38]{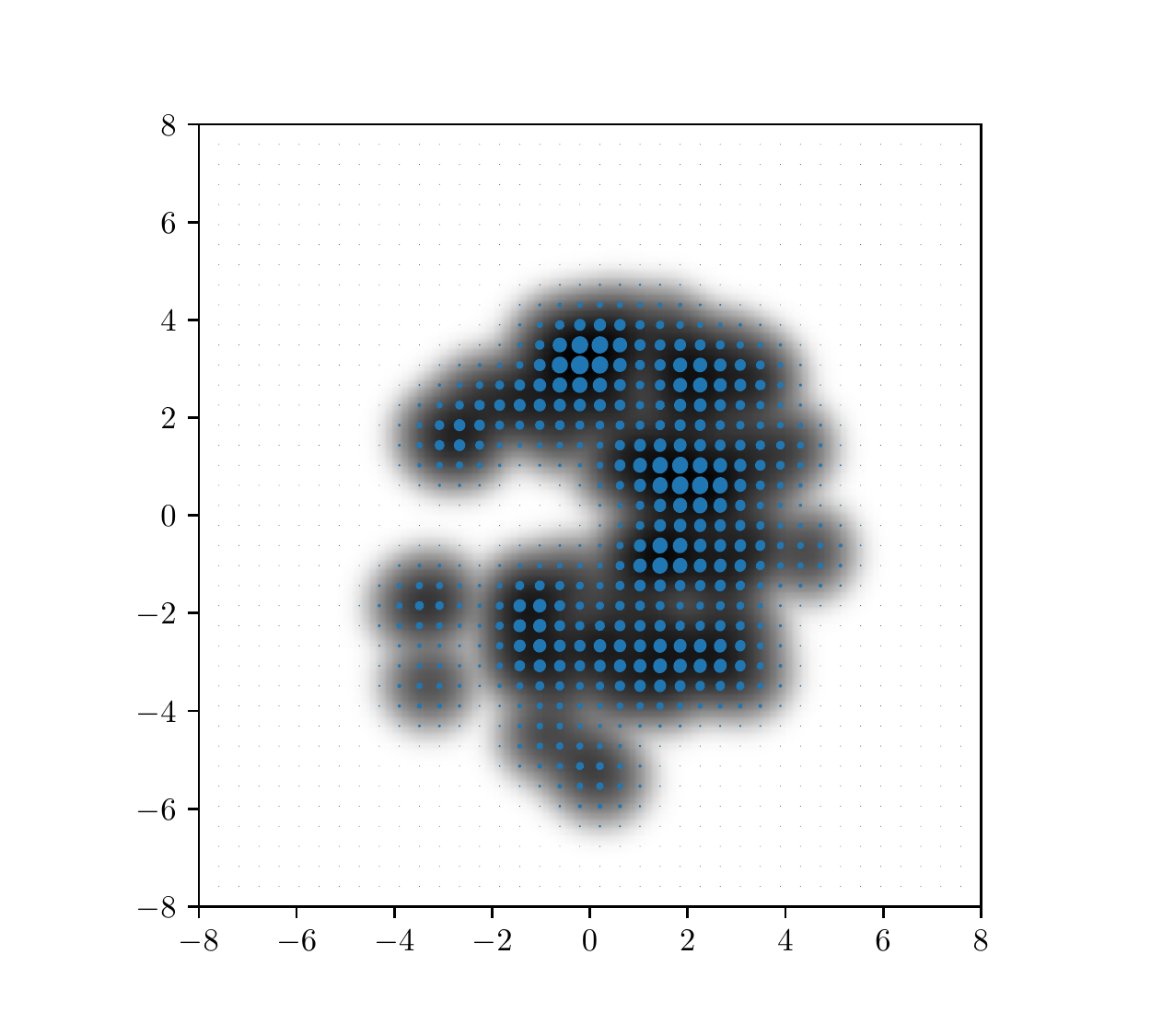}
    \end{minipage}
    \begin{minipage}[c]{0.32\linewidth}
        \centering
         \includegraphics[scale=0.38]{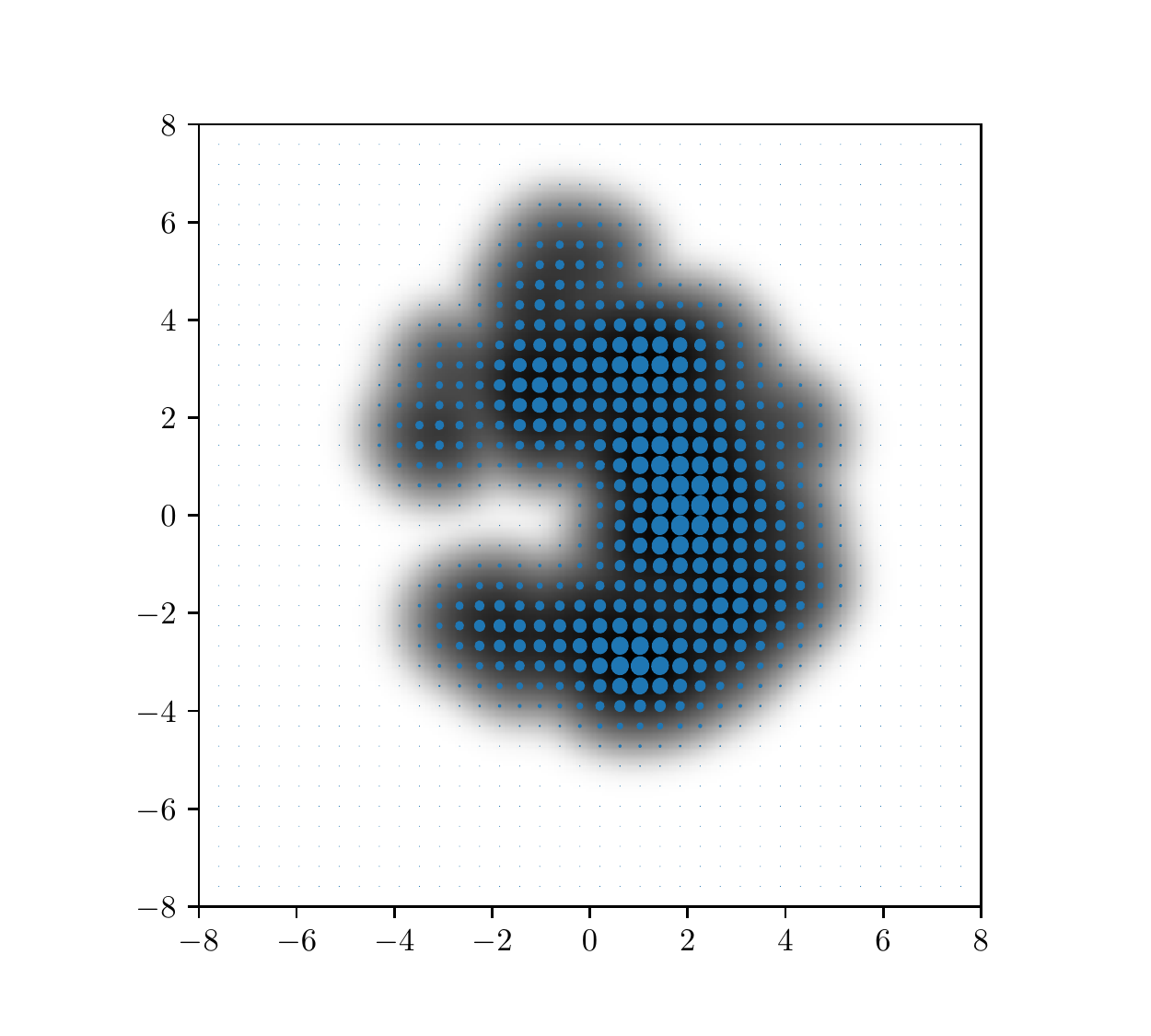}
    \end{minipage}
    \begin{minipage}[c]{0.32\linewidth}
         \centering
         \includegraphics[scale=0.38]{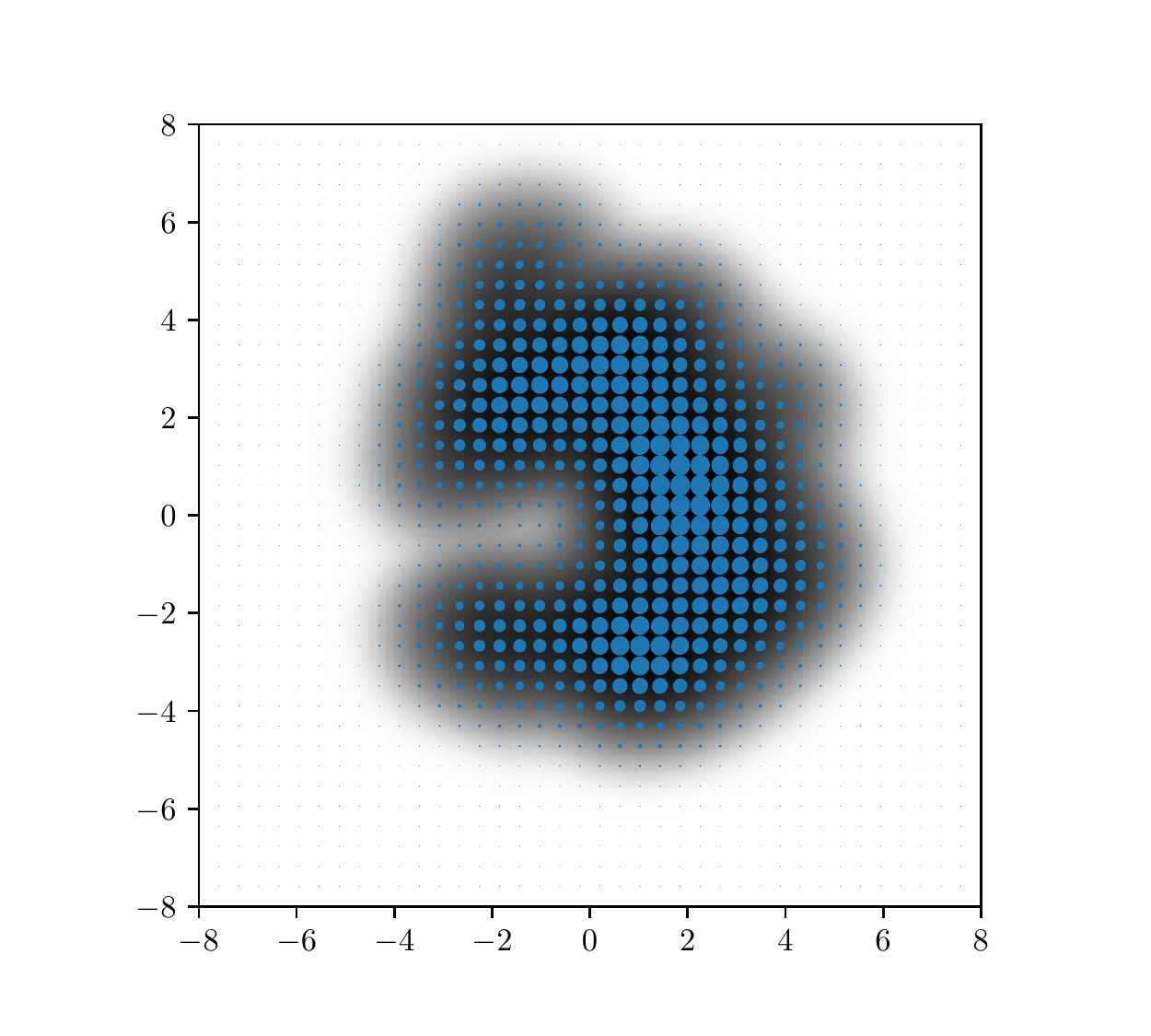}
    \end{minipage}
    \caption{Estimated metrics with 3 classes of 50 elements each. \textit{Top:}
    The logarithm of the volume element $\sqrt{\det G(z)}$ along with the
    means $\mu(x_i)$ of the distribution associated to the latent  variable $z_i \sim \mathcal{N}(\mu(x_i), \Sigma(x_i))$ for each class and for models trained with a fixed
    temperature of $T=0.6$ (left), $T=0.8$ (middle) and $T=1$ (right).
    \textit{Bottom:} The metric's eigenvalues and eigenvectors. Models are
    trained on 80\% of the data set randomly split and ensuring balanced classes with 300 epochs. We use
    $n_{\mathnormal{lf}}=5$ and a fixed regularization factor set to $\lambda =
    10^{-2}$.}
    \label{fig: metric}
\end{figure}

\begin{figure}[ht]
    \centering
    \begin{minipage}[c]{0.32\linewidth}
        \centering
         \includegraphics[scale=0.38]{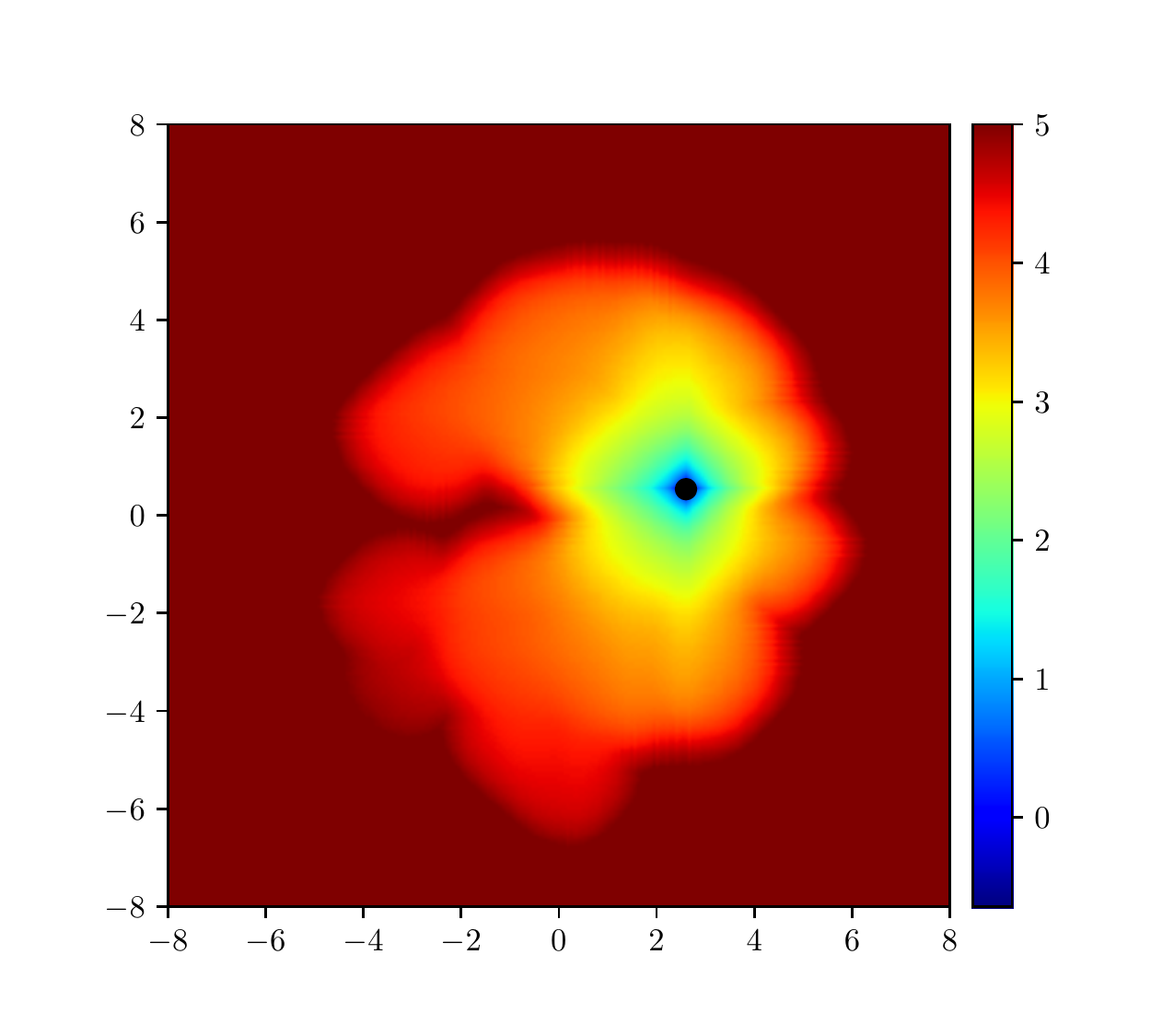}
    \end{minipage}
    \begin{minipage}[c]{0.32\linewidth}
        \centering
         \includegraphics[scale=0.38]{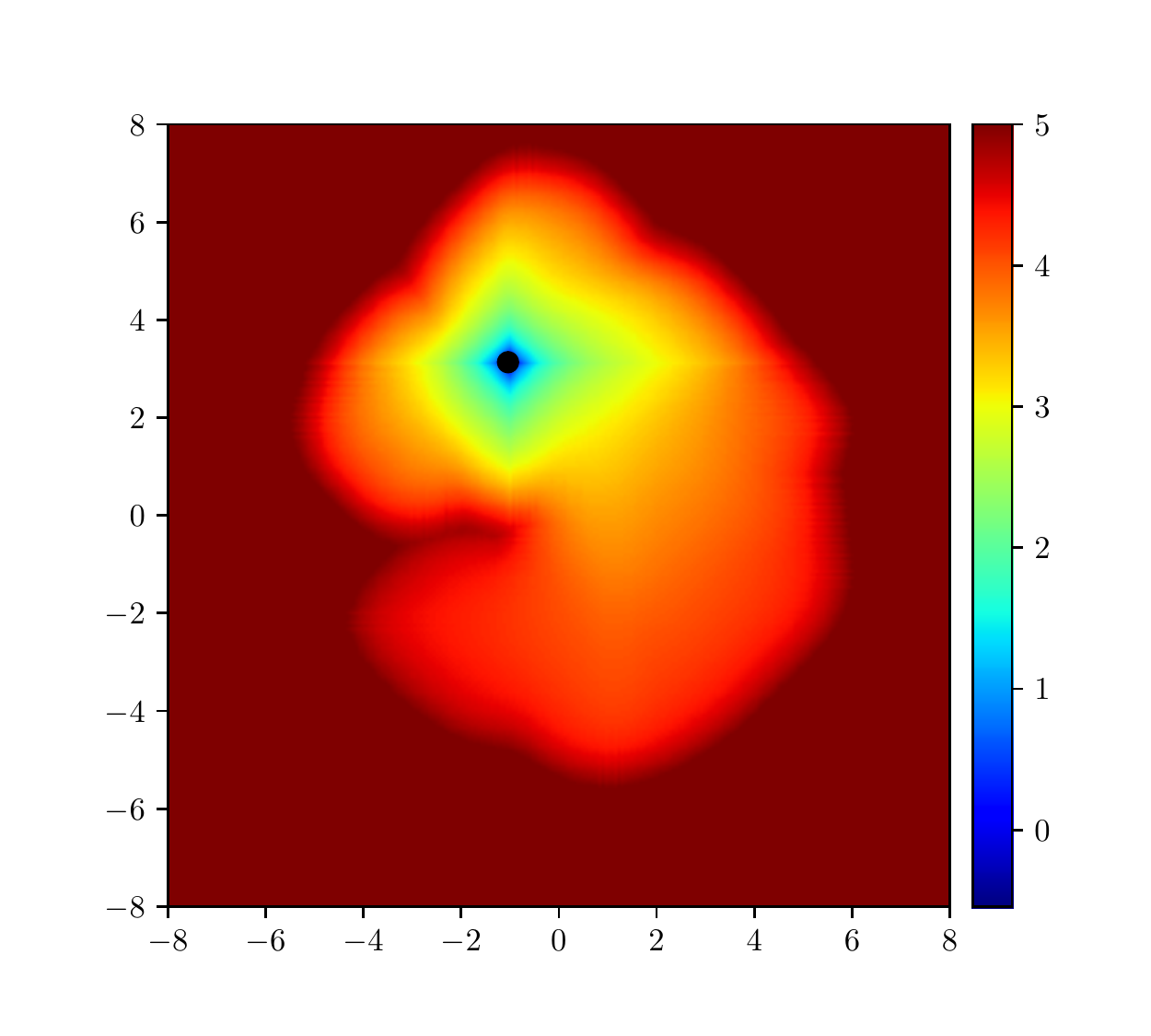}
    \end{minipage}
    \begin{minipage}[c]{0.32\linewidth}
         \centering
         \includegraphics[scale=0.38]{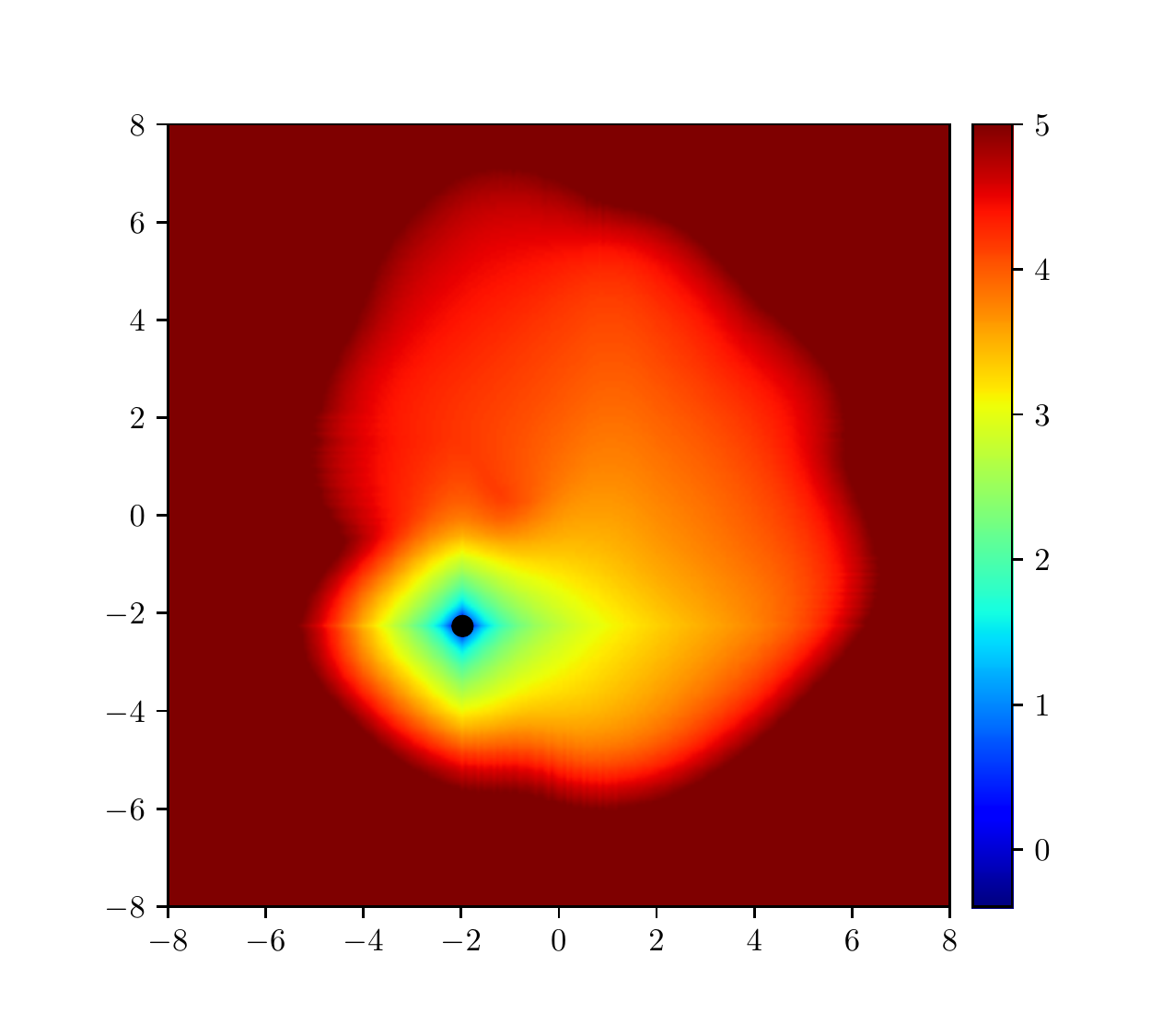}
    \end{minipage}
    \centering
    \vskip -1.em
    \begin{minipage}[c]{0.32\linewidth}
        \centering
         \includegraphics[scale=0.38]{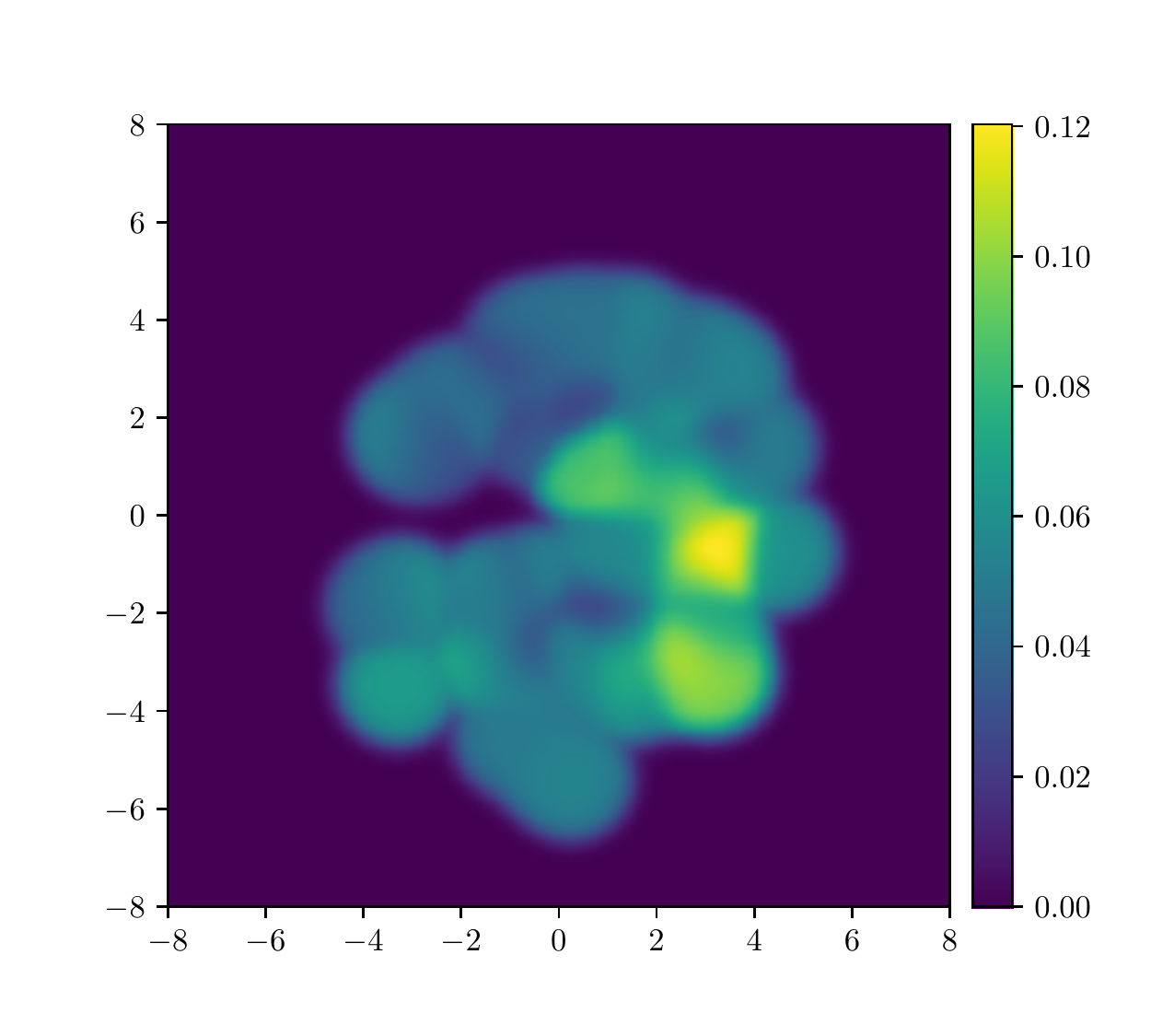}
    \end{minipage}
    \begin{minipage}[c]{0.32\linewidth}
        \centering
         \includegraphics[scale=0.38]{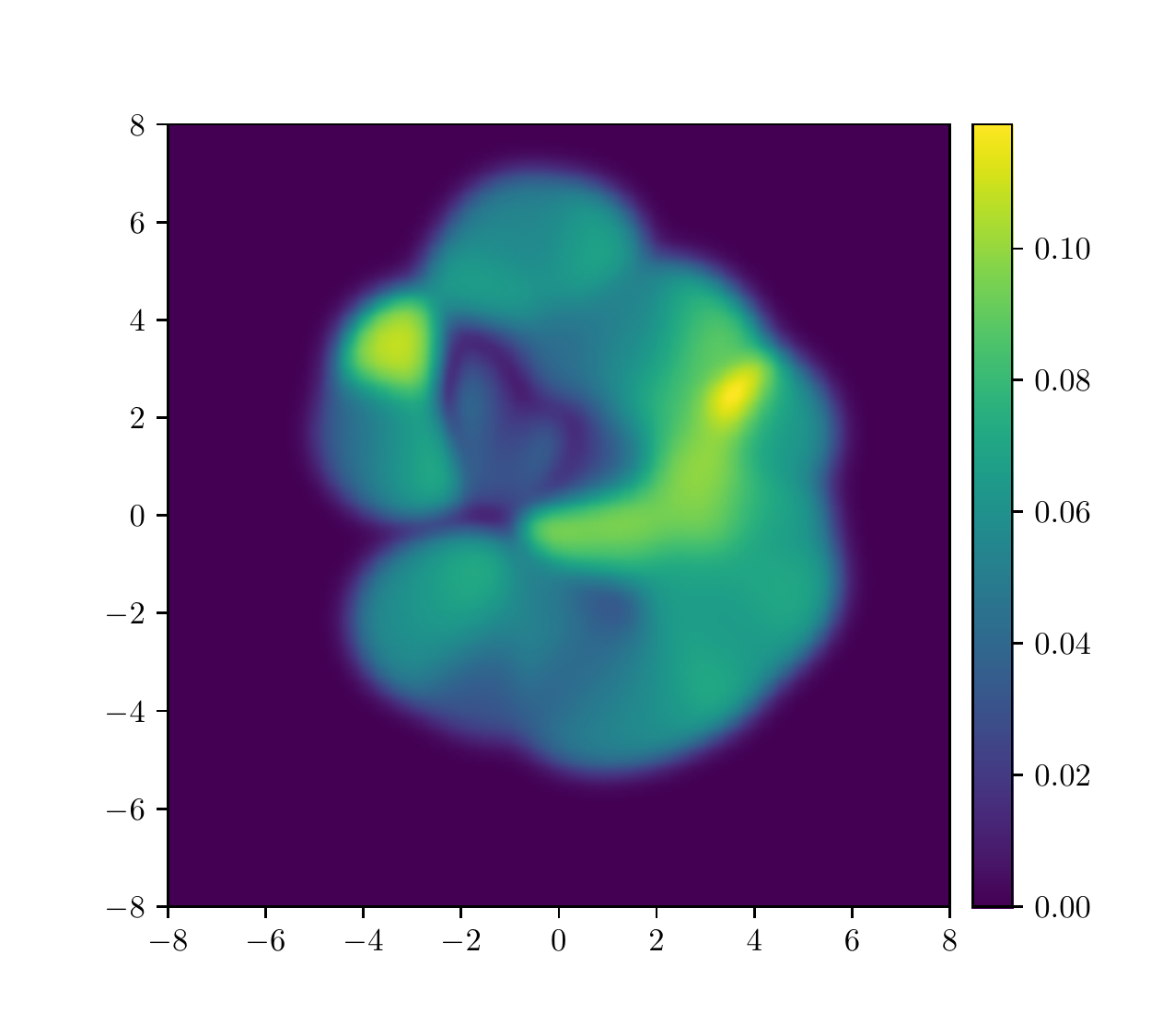}
    \end{minipage}
    \begin{minipage}[c]{0.32\linewidth}
         \centering
         \includegraphics[scale=0.38]{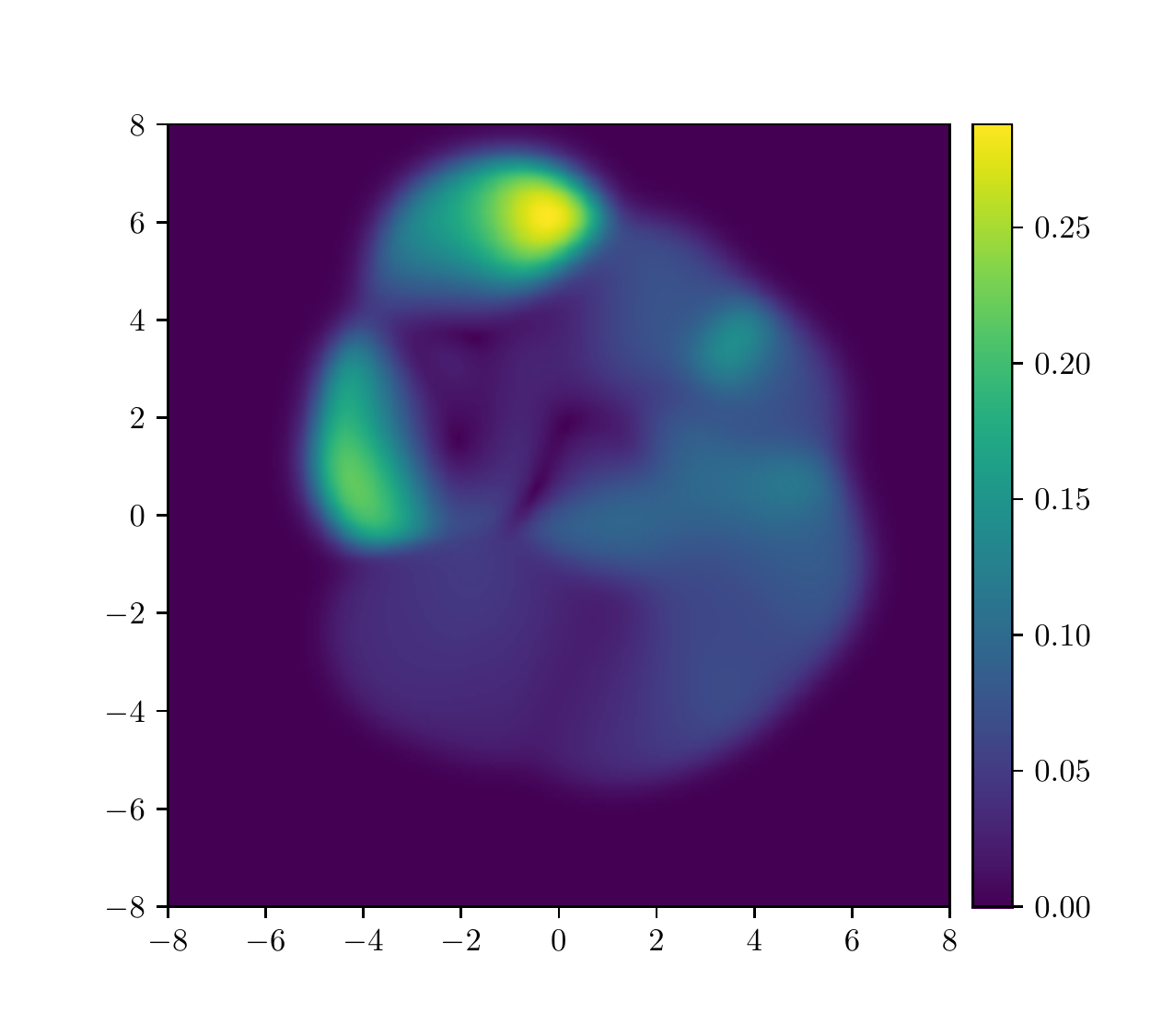}
    \end{minipage}
    \caption{{\it Top:} Log-distance maps to a point in the latent space computed using Dijkstra algorithm. \textit{Bottom:} Metric's anisotropy  $ A(z) =\frac{\lambda_{\max}(z) - \lambda_{\min}(z)}{\lambda_{\max}(z) + \lambda_{\min}(z)}$.}
    \label{fig: metric anisotropy distance map}
\end{figure}

    First of all, the \textit{shape} and the influence of the metric temperature $T$ is studied. To do so, we train a RHVAE model with 3 classes of the MNIST data set. In order to stick to the \textit{small } data set framework, we only select 50 samples of each class and train the model on 80\% randomly chosen from the data set ensuring balanced classes. The regularization factor of the metric is set to $\lambda=10^{-2}$ for each experiment and we consider a fixed $\varepsilon_{\mathnormal{lf}}=10^{-2}$ along with $n_{\mathnormal{lf}}=5$. The models are trained with 300 epochs with different temperatures $T$ ranging from $T=0.6$ to $T=1$. In Figure~\ref{fig: metric} (top row), we display the learned latent space for these 3 temperatures. The coloured dots represent the mean $\mu(x_i)$ of the distribution associated to the latent variable $z_i \sim \mathcal{N}(\mu(x_i), \Sigma(x_i))$ for each class while the log of the volume element of the learned manifold $\sqrt{\det \mathbf{G}(z)}$ is displayed in the background. We also present the eigenvalues and eigenvectors of the learned metric thanks to ellipses (bottom row). Interestingly, even with the proposed metric, the model is apparently able to distinguish the data points belonging to the same class and group them together. In addition, the volume element $\sqrt{\det \mathbf{G}(z)}$ is far smaller where samples are located than where it is not. By construction the regularization factor scales its value far from the data and so has a strong impact on geodesic paths. This is interesting since the metric gives strong information about the location of the data allowing for potentially better clustering, interpolation or generation. In Figure~\ref{fig: metric anisotropy distance map}, the distance maps to a given point in the latent space are presented for the same learned metrics (top row) along with the metrics' anisotropy (bottom row) $ A(z) =\frac{\lambda_{\max}(z) - \lambda_{\min}(z)}{\lambda_{\max}(z) + \lambda_{\min}(z)}$ where $\lambda_{\min}$ (resp. $\lambda_{\max}$) is the minimum (resp. maximum) eigenvalue of the metric tensor $\mathbf{G}$. The distance maps are estimated using a latent space discretization (200x200) and the Dijkstra algorithm \citep{dijkstra1959note}. We refer the reader to \citep{peyre2010geodesic} for example. These maps show that the geodesic curves are designed to follow the learned manifold and so stay close to the data which reveals very useful to perform meaningful interpolation. We refer the reader to Section~\ref{Sec: Geodesics} for a more detailed analysis on geodesic interpolation.
    
\subsubsection{Geodesics Path Computation} \label{Sec: Geodesics}

One good way to apprehend geometrical aspects provided by the learned metric is
to compute the geodesic curves. Recall that the length of a curve $\gamma: [0, 1] \to \mathcal{M}$ from $z_1$ to $z_2$ living in a Riemannian manifold
$\mathcal{M}$ endowed with a metric $\mathbf{G}$ is given by
\begin{equation}\label{eq: length}
    L(\gamma) = \int \limits_0^1 \sqrt{\langle \gamma'(t), \gamma'(t)\rangle_{\gamma(t)}} dt \hspace{5mm} \gamma(0)=z_1 , \gamma(1)=z_2\,.
\end{equation}
Computing geodesic paths consists in finding the curve $\gamma$ minimizing Eq.~\eqref{eq: length} or equivalently the curve's energy \citep*{carmo1992riemannian}
\[
    E(\gamma) = \int \limits_0 ^1 \langle \gamma'(t), \gamma'(t)\rangle_{\gamma(t)} dt \hspace{5mm} \gamma(0)=z_1 , \gamma(1)=z_2\,.
\]
Since these two optimization problems are rather hard to solve we decide to use a method proposed by \citet{chen2018metrics} allowing for relatively "fast" geodesic computation. The main idea of their method consists in discretizing the integral as such:
\[\begin{aligned} L(\gamma) &\approx \frac{1}{n} \sum_{i=1}^n \sqrt{\langle
    \gamma'(t_i), \gamma'(t_i)\rangle_{\gamma(t_i)}}\\
              &\approx \frac{1}{n} \sum_{i=1}^n \sqrt{\gamma'(t_i)^{\top}
\mathbf{G}(\gamma(t_i)) \gamma'(t_i)} \,, \end{aligned}\] where $n$ is the
granularity of the curve. The curve $\gamma$ is then parametrized using a neural network of parameters $\omega$. The authors also proposed to add a regularization factor which according to them would "ensure that the geodesic remains close to the data" leading to the following loss function:
\[
    \mathcal{L}_{\mathnormal{geo}} = L(\gamma_{\omega}) + \lambda \lVert \mathbf{G}(\gamma_{\omega})\rVert_{\mathnormal{2}}\,.
\]

This regularization was motivated by the fact that according to the author this method can lead to local minimums. However, we do not see any apparent reason to set the regularization different from zero to compute the true geodesic paths and so we decide to only consider the loss function with $\lambda =0$. The curve $\gamma_{\omega}$ is then optimized using gradient descent. We use a 3 layers MLP network ((1, 100, $\tanh$), (100, 100, $\tanh$), (100, d, $Linear$)) for $\gamma_{\omega}$. The result of these geodesic paths and their relevance will be assessed below.

\subsubsection{Synthetic Data Set}\label{Sec: Geodesics synthetic}  
\begin{figure}[ht]
    \centering
    \begin{minipage}[c]{0.32\linewidth}
        \centering
        \subcaption*{}
         \includegraphics[scale=0.38]{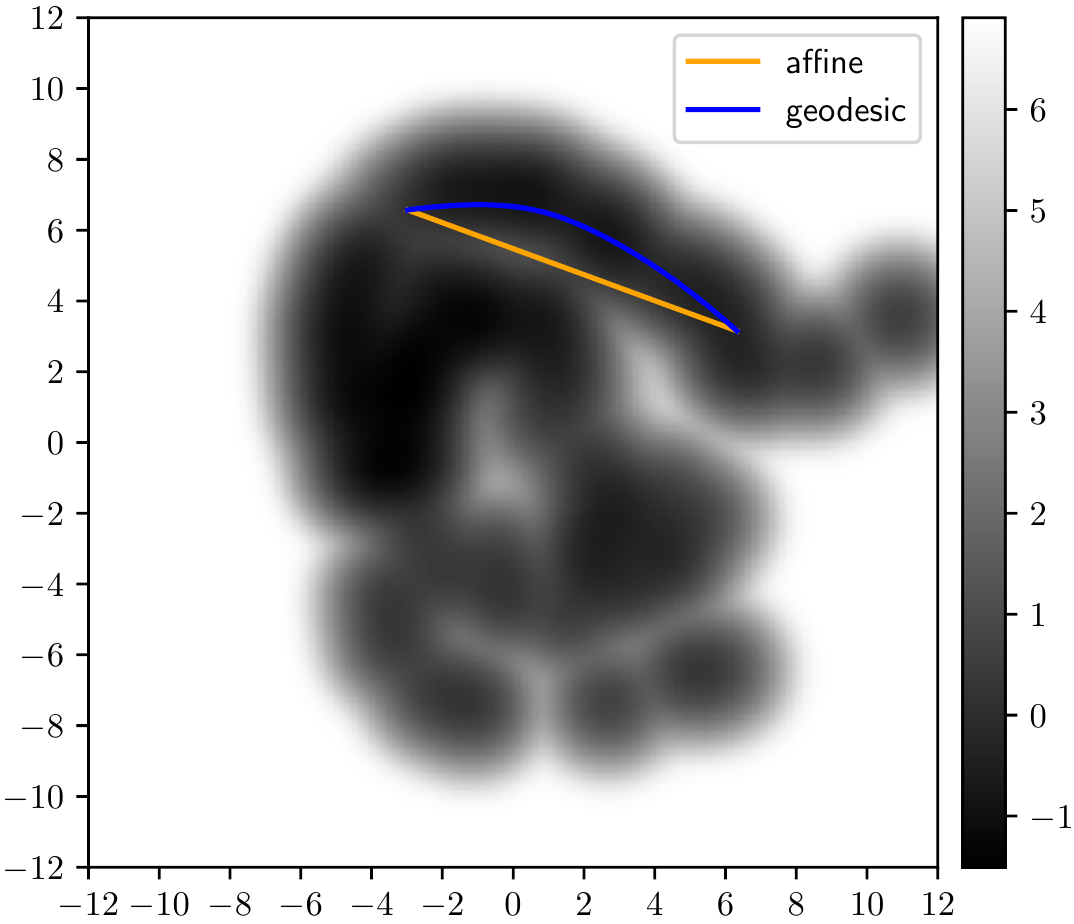}
    \end{minipage}
    \begin{minipage}[c]{0.32\linewidth}
        \centering
        \subcaption*{RHVAE (Ours)}
         \includegraphics[scale=0.38]{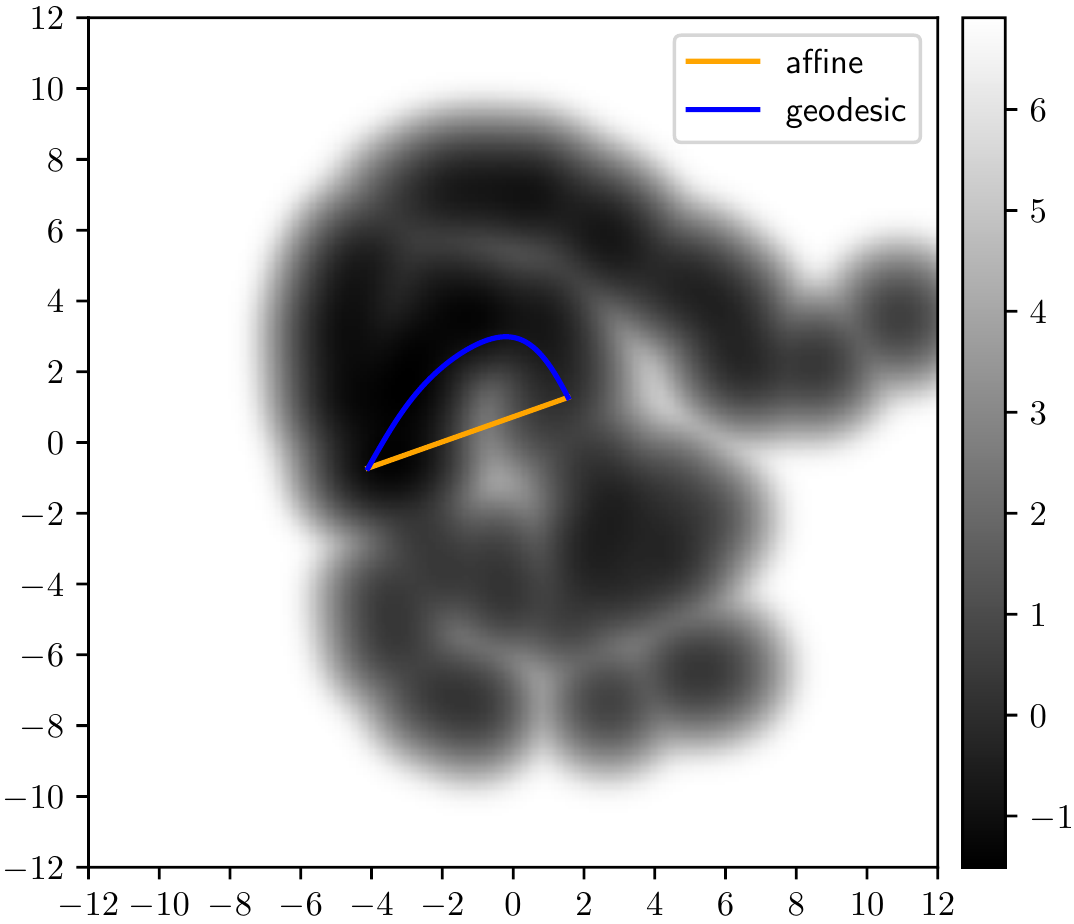}
    \end{minipage}
    \begin{minipage}[c]{0.32\linewidth}
        \centering
        \subcaption*{}
         \includegraphics[scale=0.38]{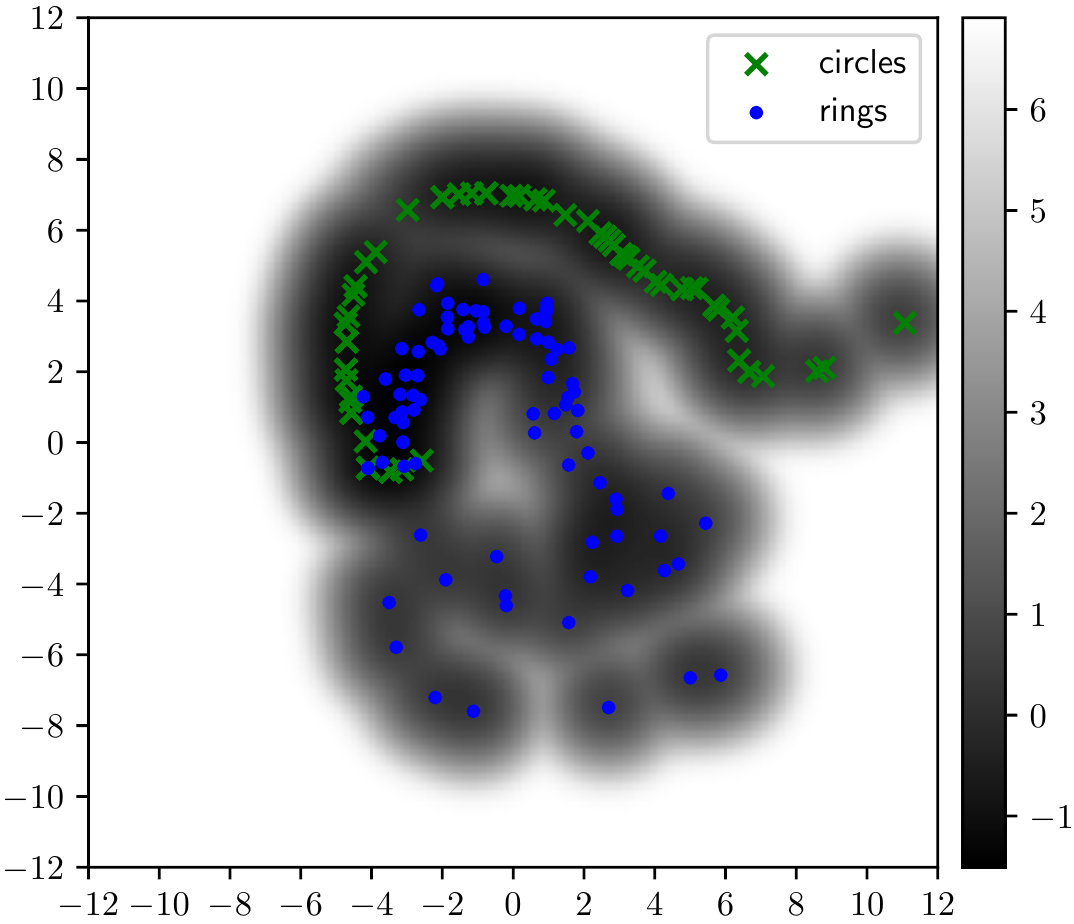}
    \end{minipage}
    \vskip 0.5em
    \centering
    \begin{minipage}[c]{\linewidth}
        \centering
         \includegraphics[scale=0.28]{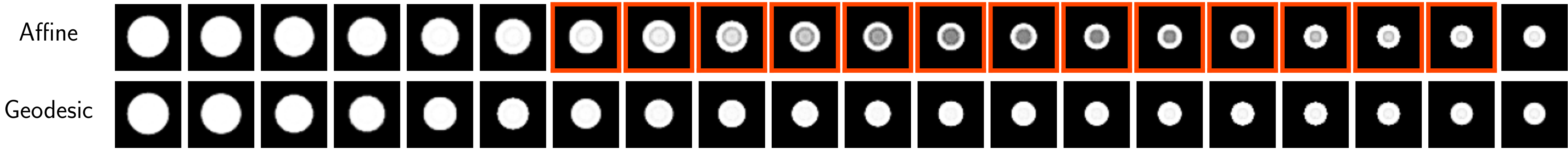}
    \end{minipage}
    \vskip 0.5em
    \centering
    \begin{minipage}[c]{\linewidth}
        \centering
         \includegraphics[scale=0.28]{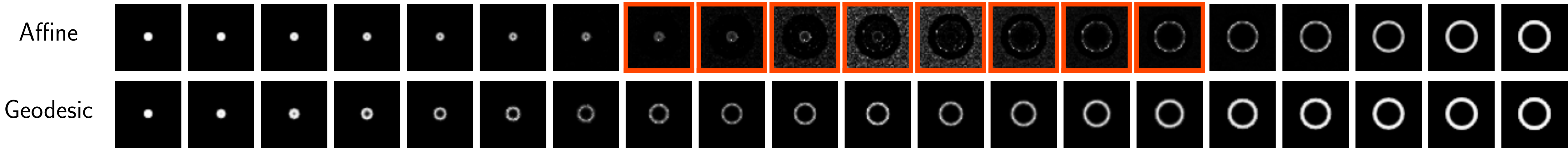}
    \end{minipage}
    \caption{ \textit{Top:} The latent space along with the logarithm of the volume element and interpolation curves. \textit{Bottom:} The decoded samples along the curves (granularity of 5 time steps). The model is trained on 80\% of a synthetic data set composed by 100 circles and 100 rings with different diameters and thicknesses.}
    \label{fig: Geodesic Interpolation Shapes}
\end{figure}

In order to assess the usefulness of the proposed metric we first try to perform some geodesic interpolations on a hand-made synthetic data set. The data set we consider consists in 200 binary images, 100 of which represent circles and the others rings. For each shape, we consider different diameters and thicknesses. An extract of the training samples is available in Appendix~\ref{app: Geodesic training samples}. The interpolations are performed such that the starting and ending points are the mean values of the distribution of two encoded samples from the training set i.e. $z_1 = \mu(x_1)$ and $z_2 = \mu(x_2)$ where we recall $z_i \sim \mathcal{N}(\mu(x_i), \Sigma(x_i))$, $i \in \{1, 2\}$. We 
compare the resulting curves under the euclidean metric (i.e. affine interpolation) and the proposed metric (i.e. geodesic interpolation) in Figure~\ref{fig: Geodesic Interpolation Shapes}. Two interpolations are presented with the logarithm of the volume element (top left and middle) along with the learned latent space (top right). The affine and geodesic curves are then discretized in 100 time steps and we present the decoded samples all along the curves with a granularity of 5 time steps (bottom). Impressively, using the proposed metric allows for far more meaningful interpolations. On the two first rows we try to interpolate between two points of the latent space corresponding to circles of different diameters when decoded. While the affine interpolation fails to keep the intrinsic topology of the data (see orange frames), the geodesic interpolation seems to be able to constraint the curve so that it stays within the learned manifold and so each point along the curve remains a circle the diameter of which is smoothly decreased. The second row consists in the interpolation between latent variables leading to a small dot and a larger ring when decoded. Again, the affine curve leaves the manifold and so is not able to produce a meaningful interpolation since some samples along the curve do not even look like either a ring or a circle (see orange frames) whereas the geodesic curve is again able to do so. Interestingly, the latent space with the proposed metric seems to highlight an underlying structure since the circles seem "grouped" together and ordered by size when one considers geodesic distances. The clustering ability of the model will be discussed into details in Section~\ref{Sec: Clustering}.

\subsubsection{Real Data Sets}\label{Sec: Geodesics real}  
\begin{figure}[t]
    \centering
    \begin{minipage}[c]{\linewidth}
        \centering
        \subcaption*{VAE}
        \vskip -0.5em
         \includegraphics[scale=0.5]{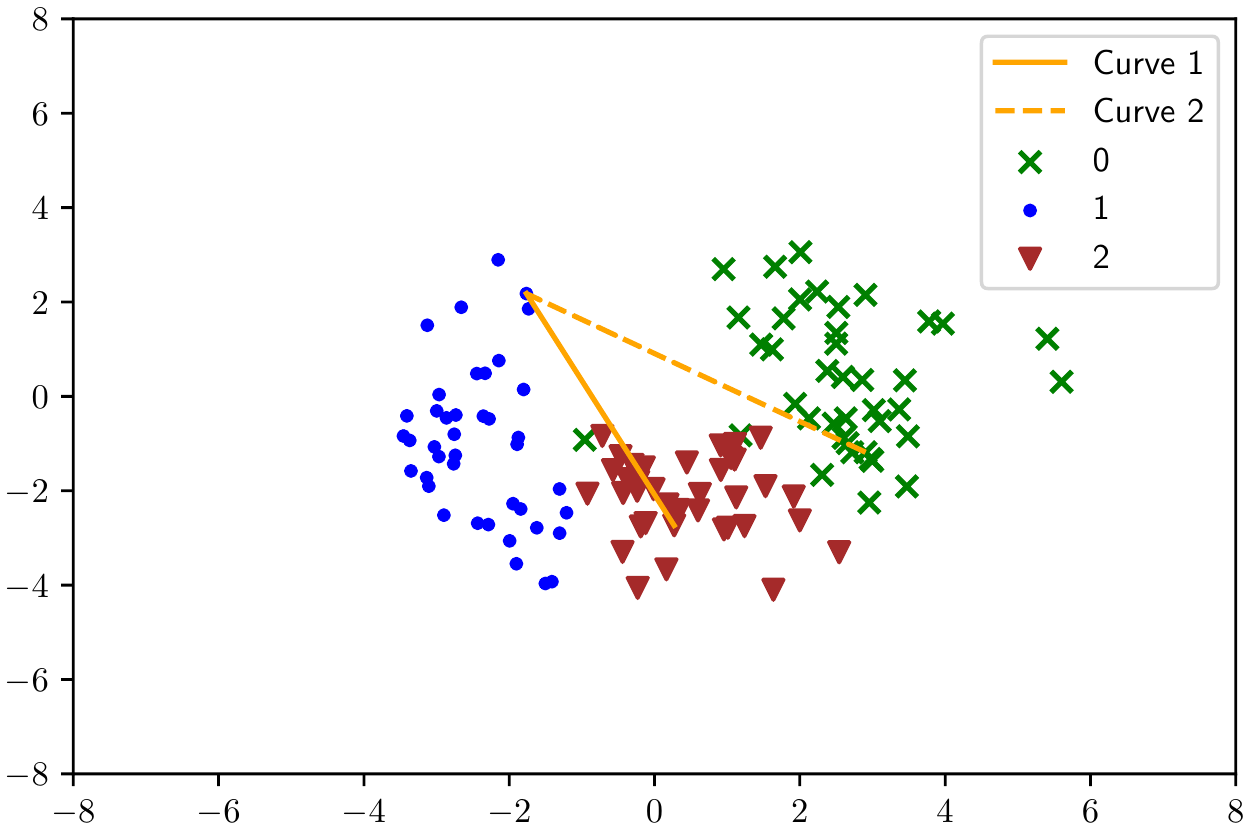}
    \end{minipage}
    \centering
    \vskip 0.5em
    \begin{minipage}[c]{\linewidth}
        \centering
         \includegraphics[scale=0.28]{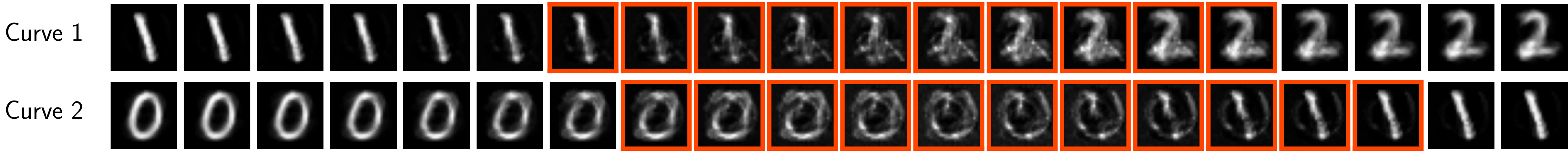}
    \end{minipage}
    \caption{Affine interpolations with a classic VAE trained with 3 classes of 50
    elements each. The model is trained with 300 epochs on 80\% of the data set
    randomly chosen. \textit{Top:} The latent space along with the  encoded data points and interpolation curves. \textit{Bottom}: The decoded samples along the curves (granularity of 5 time steps).}
    \label{fig: VAE Interpolation MNIST}
\end{figure}

\begin{figure}[p]
    \centering
    \begin{minipage}[c]{0.32\linewidth}
        \centering
         \subcaption*{}
         \vskip -0.5em
         \includegraphics[scale=0.38]{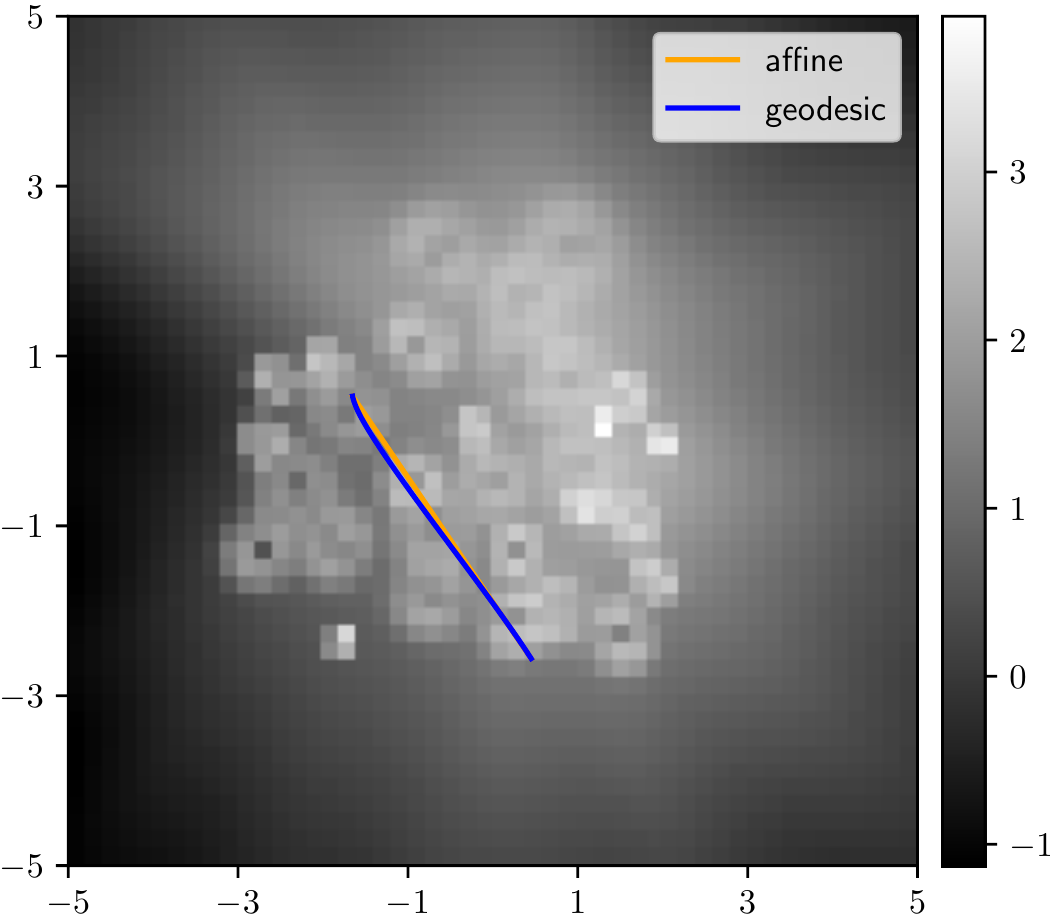}
    \end{minipage}
    \begin{minipage}[c]{0.32\linewidth}
        \centering
        \subcaption*{VAE  \citep{arvanitidis2017latent}}
        \vskip -0.5em
         \includegraphics[scale=0.38]{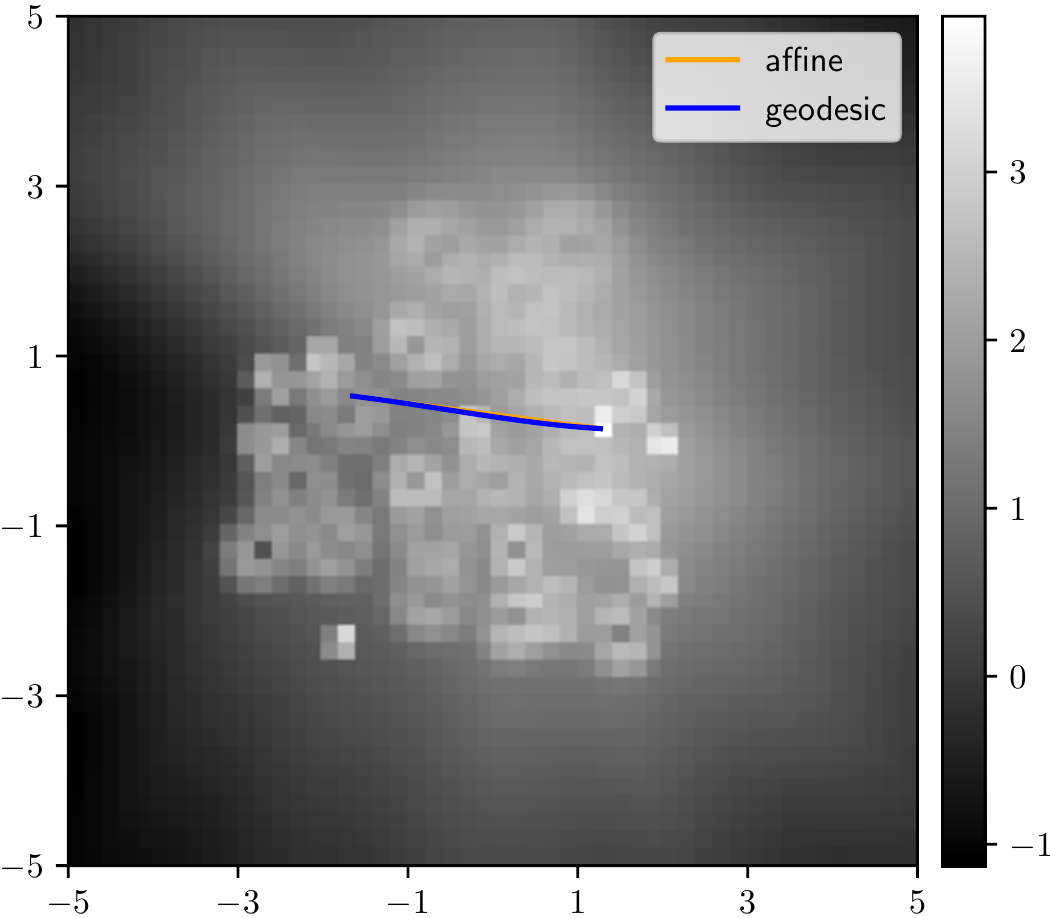}
    \end{minipage}
    \begin{minipage}[c]{0.32\linewidth}
        \centering
        \subcaption*{}
        \vskip -0.5em
         \includegraphics[scale=0.38]{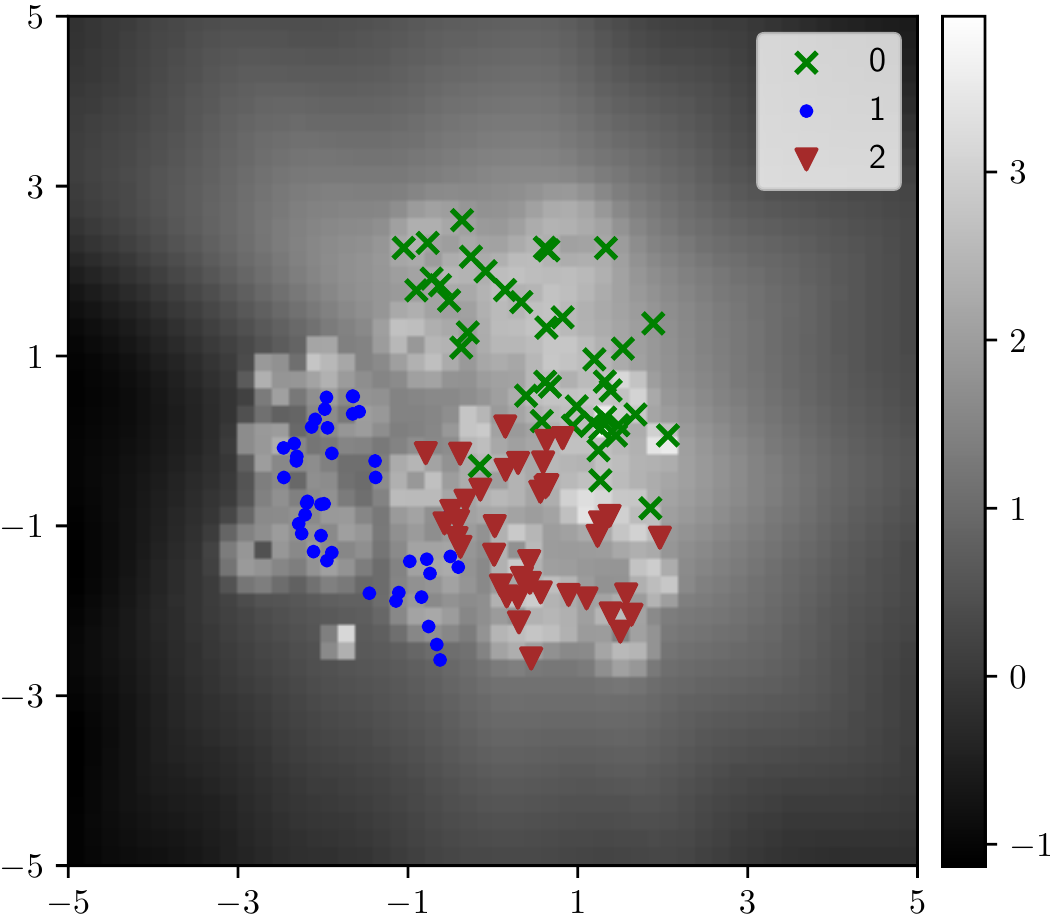}
    \end{minipage}
    \centering
    \vskip 0.5em
    \begin{minipage}[c]{\linewidth}
        \centering
         \includegraphics[scale=0.28]{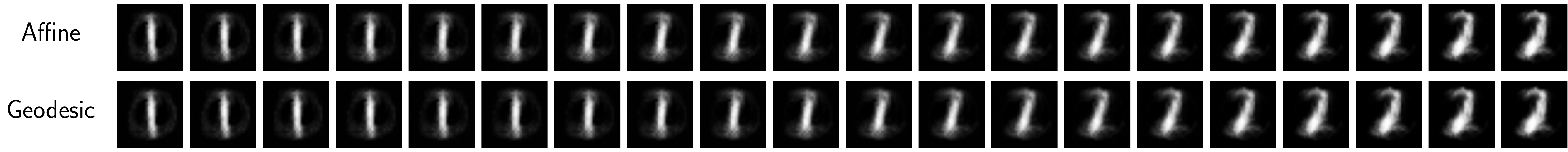}
    \end{minipage}
    \vskip 0.5em
    \centering
    \begin{minipage}[c]{\linewidth}
        \centering
         \includegraphics[scale=0.28]{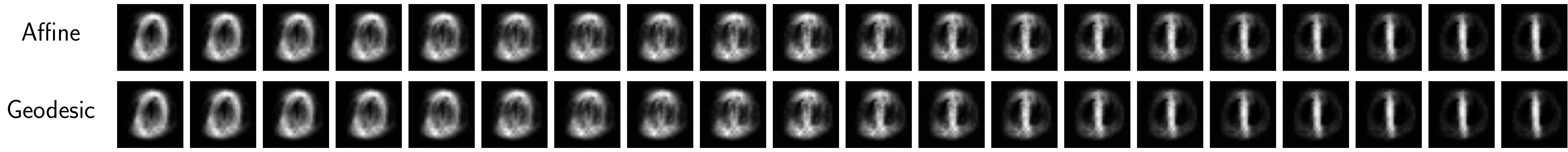}
    \end{minipage}
    \caption{Affine and geodesic interpolations with a VAE trained as specified in \citep{arvanitidis2017latent} with 3 classes of 50 elements each. The model is trained with 300 epochs on 80\% of the data set randomly chosen. \textit{Top:} The latent space along with the logarithm of the volume element and interpolation curves. \textit{Bottom:} The decoded samples all along the curves (granularity of 5 time steps).}
    \label{fig: Geodesic Interpolation Arvanitidis MNIST}
\end{figure}

\begin{figure}[p]
    \centering
    \begin{minipage}[c]{0.32\linewidth}
        \centering
         \subcaption*{}
         \vskip -0.5em
         \includegraphics[scale=0.38]{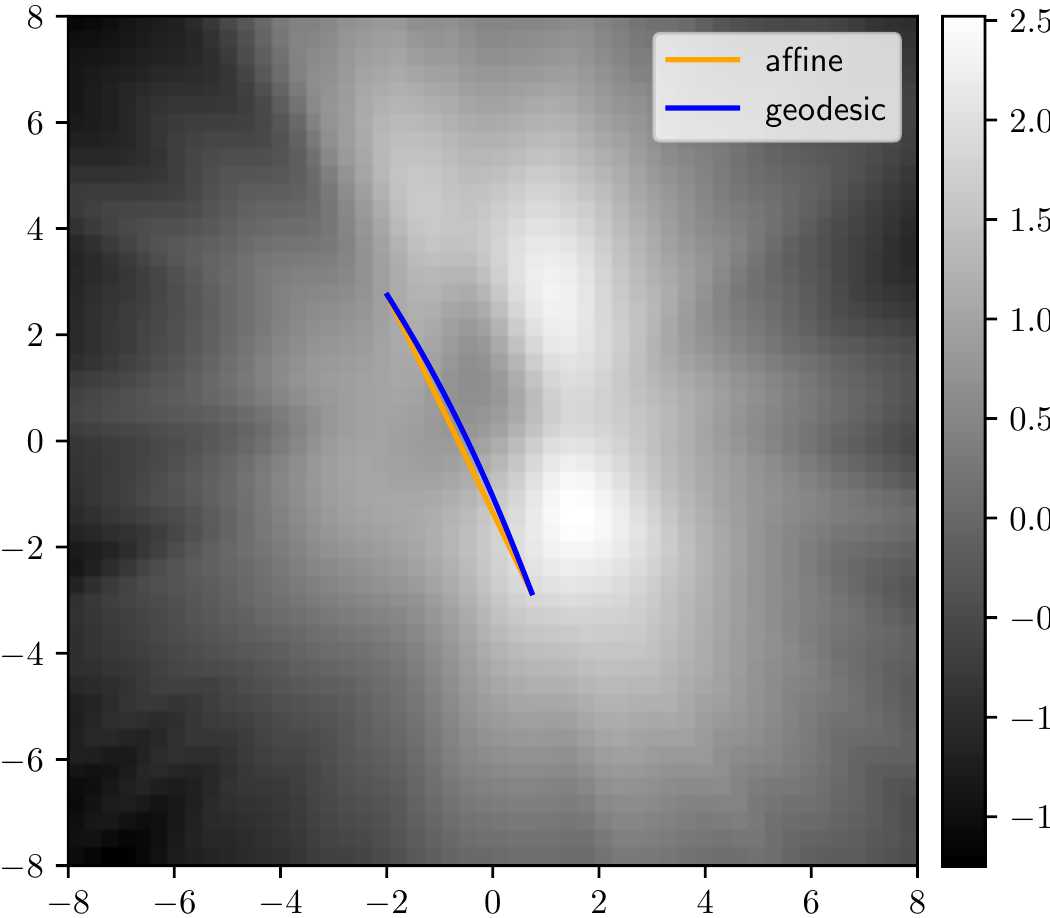}
    \end{minipage}
    \begin{minipage}[c]{0.32\linewidth}
        \centering
         \subcaption*{VAE \citep{chen2018metrics}}
         \vskip -0.5em
         \includegraphics[scale=0.38]{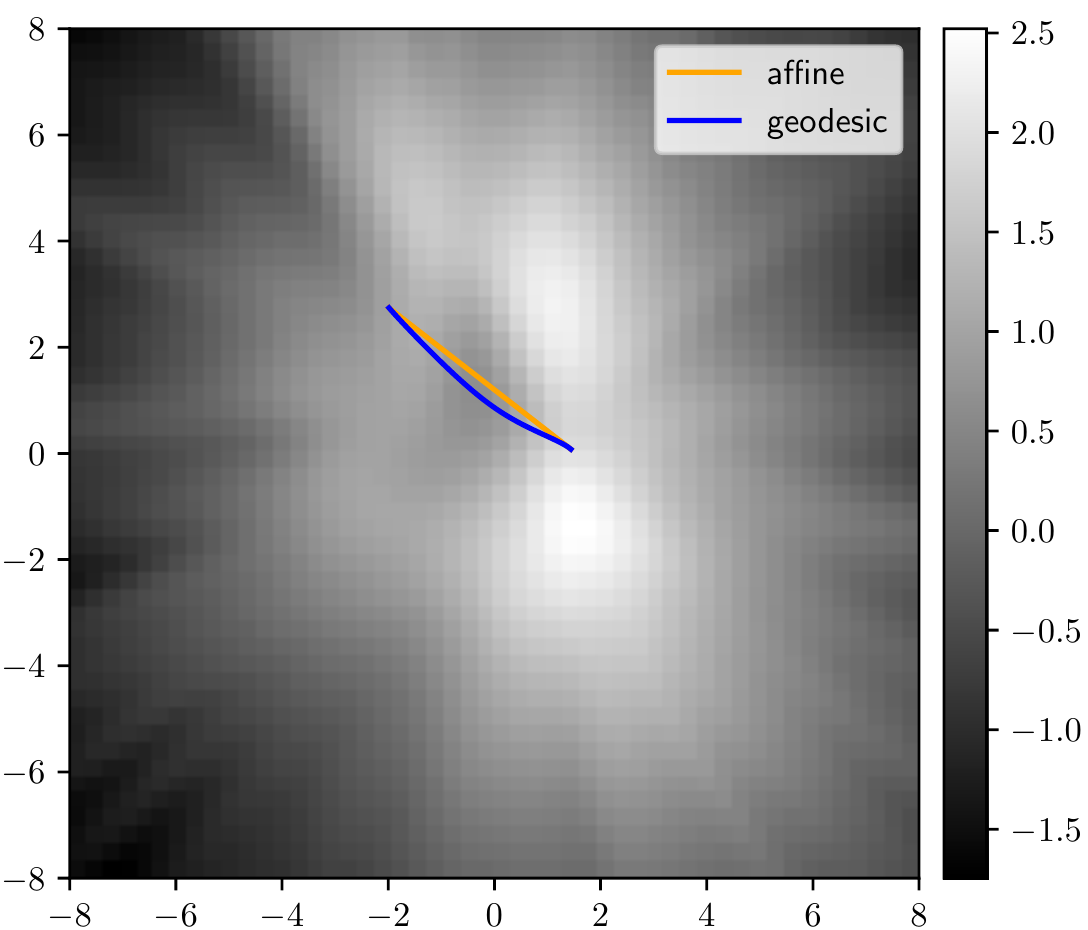}
    \end{minipage}
    \begin{minipage}[c]{0.32\linewidth}
        \centering
         \subcaption*{}
         \vskip -0.5em
         \includegraphics[scale=0.38]{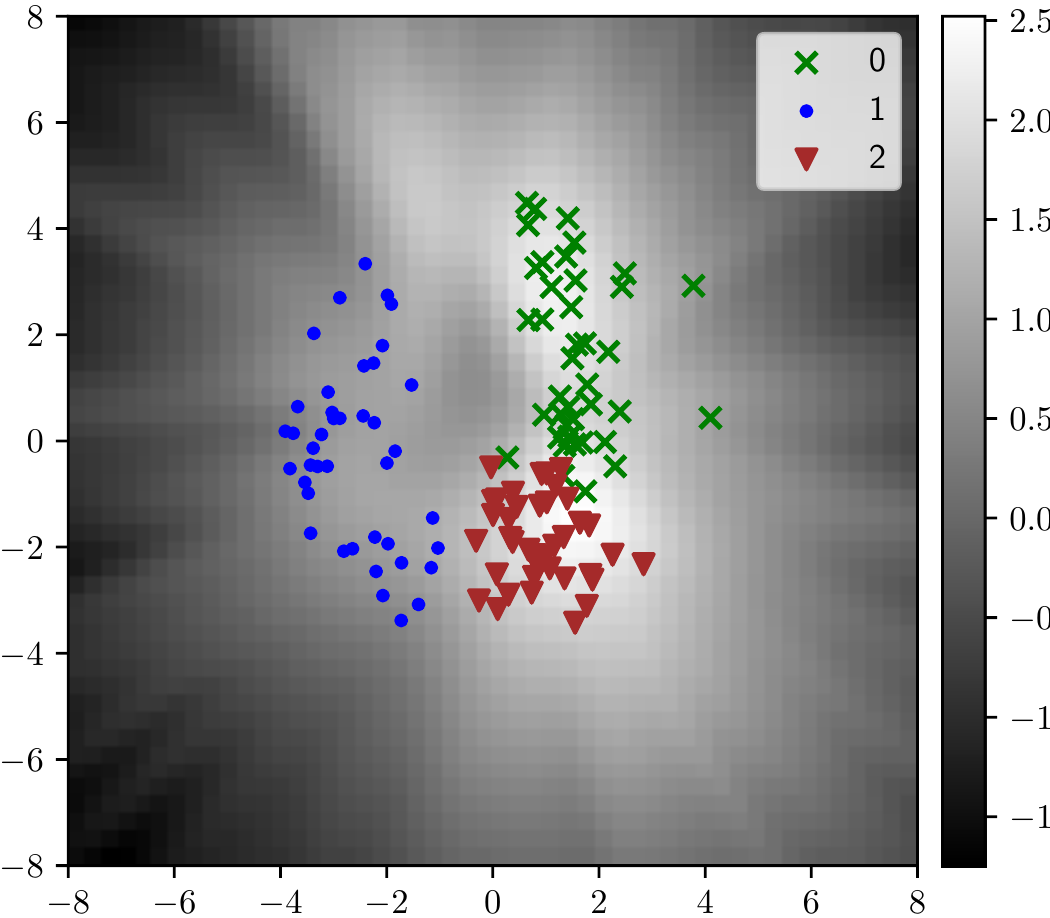}
    \end{minipage}
    \centering
    \vskip 0.5em
    \begin{minipage}[c]{\linewidth}
        \centering
         \includegraphics[scale=0.28]{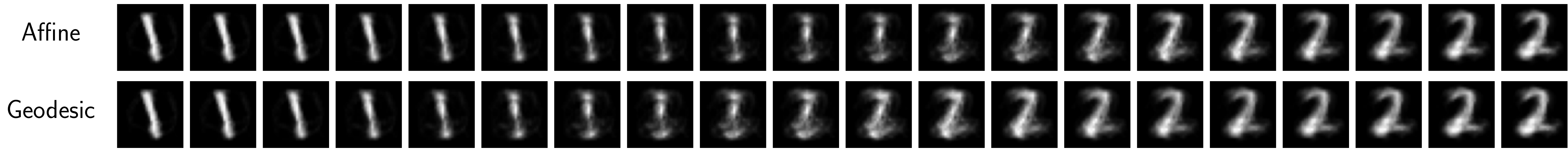}
    \end{minipage}
    \vskip 0.5em
    \centering
    \begin{minipage}[c]{\linewidth}
        \centering
         \includegraphics[scale=0.28]{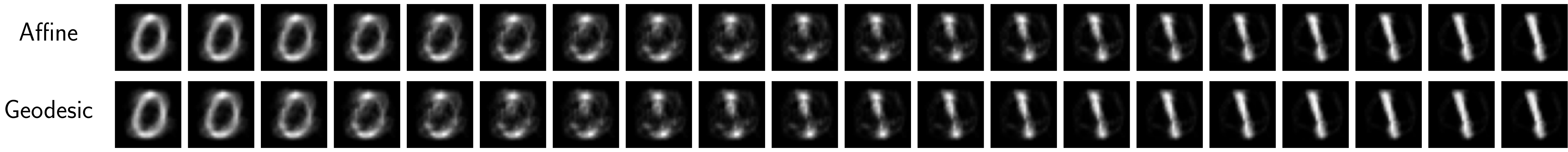}
    \end{minipage}
    \caption{Affine and geodesic interpolations with a VAE trained as specified in \citep{chen2018metrics} with 3 classes of 50 elements each. The model is trained with 300 epochs on 80\% of the data set randomly chosen. \textit{Top:} The latent space along with the logarithm of the volume element and interpolation curves. \textit{Bottom:} The decoded samples all along the curves (granularity of 5 time steps).}
    \label{fig: Geodesic Interpolation Chen MNIST}
\end{figure}

\begin{figure}[p]
    \centering
    \begin{minipage}[c]{0.32\linewidth}
        \centering
        \subcaption*{}
        \vskip -0.5em
         \includegraphics[scale=0.38]{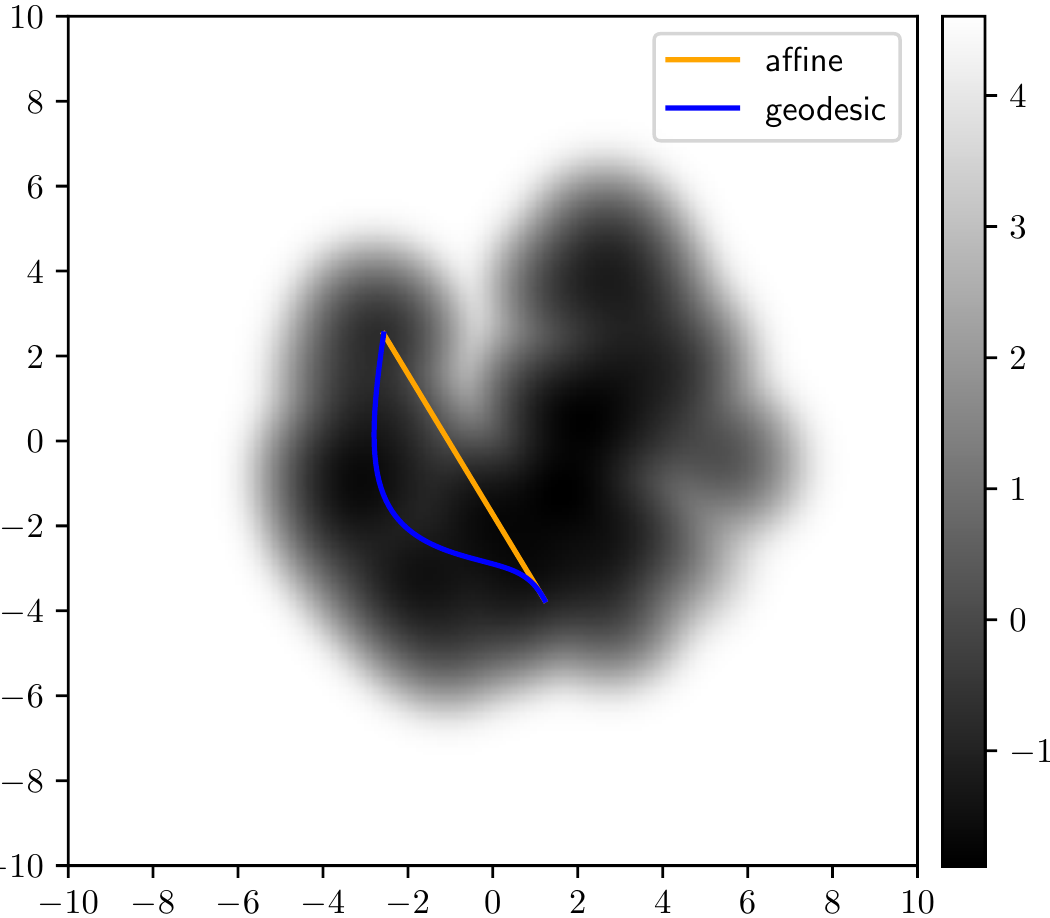}
    \end{minipage}
    \begin{minipage}[c]{0.32\linewidth}
        \centering
        \subcaption*{RHVAE (Ours)}
        \vskip -0.5em
         \includegraphics[scale=0.38]{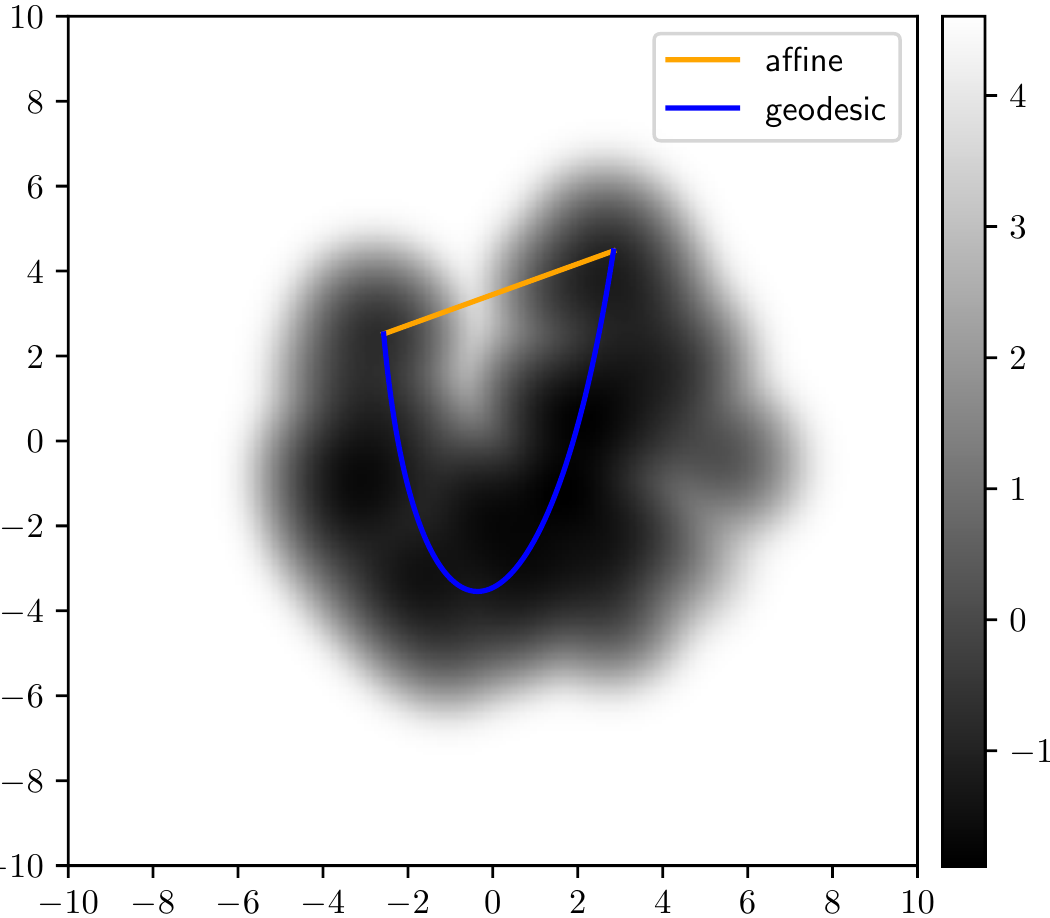}
    \end{minipage}
    \begin{minipage}[c]{0.32\linewidth}
        \centering
        \subcaption*{}
        \vskip -0.5em
         \includegraphics[scale=0.38]{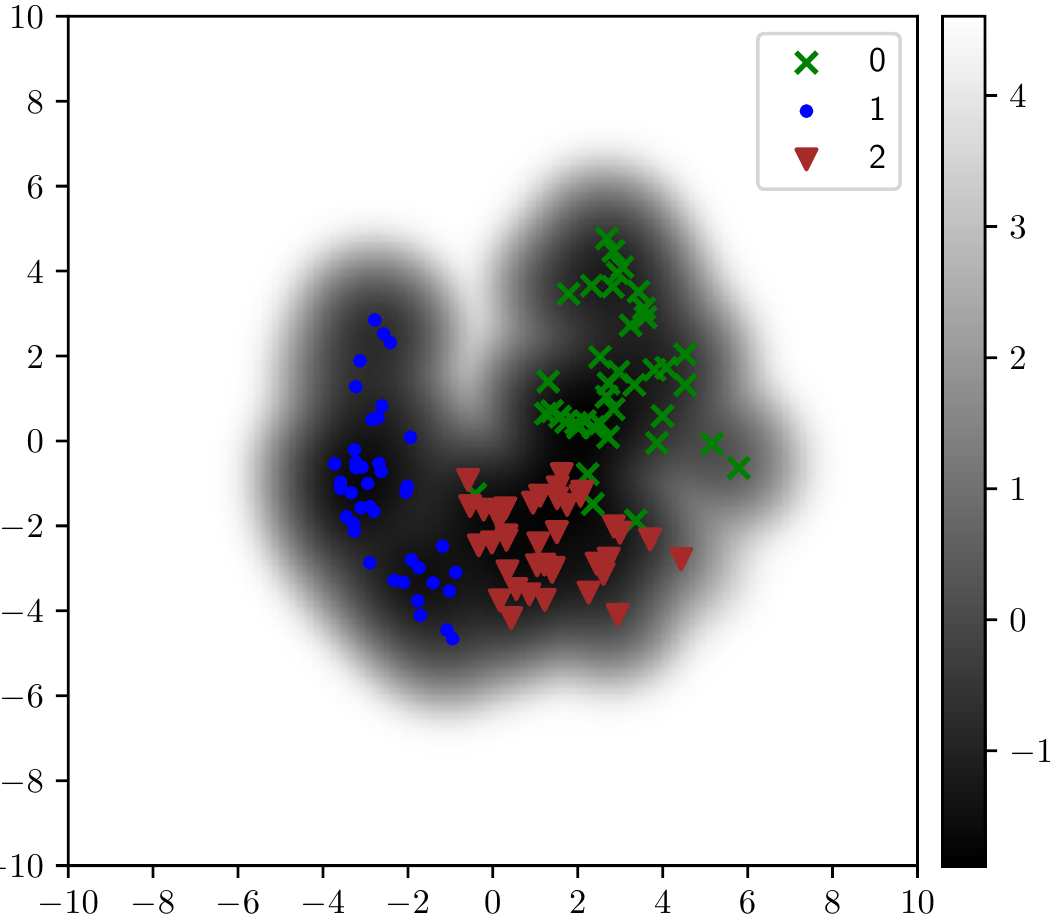}
    \end{minipage}
    \vskip 0.5em
    \centering
    \begin{minipage}[c]{\linewidth}
        \centering
         \includegraphics[scale=0.28]{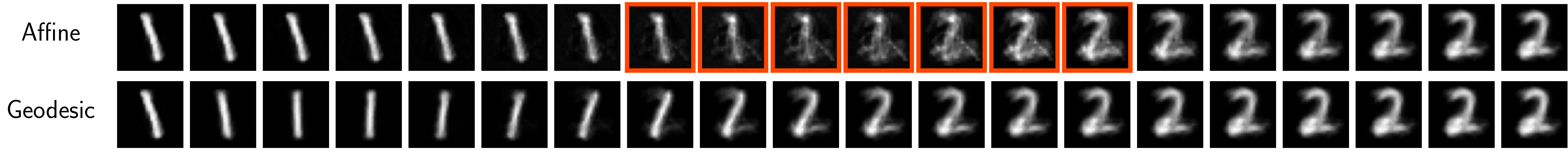}
    \end{minipage}
    \vskip 0.5em
    \centering
    \begin{minipage}[c]{\linewidth}
        \centering
         \includegraphics[scale=0.28]{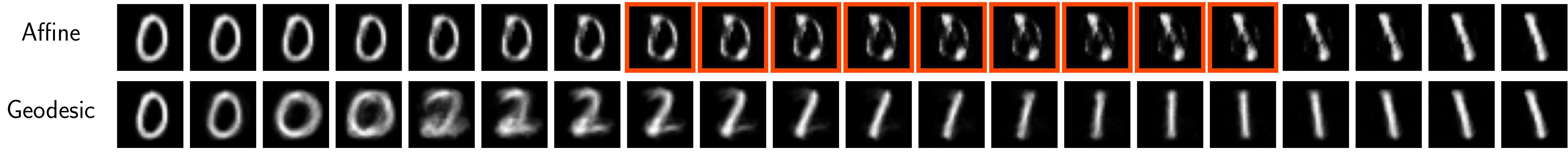}
    \end{minipage}
    \caption{Affine and geodesic interpolation with the proposed RHVAE trained with 3 classes of 50 elements each. The model is trained with 300 epochs on 80\% of the data set randomly chosen. \textit{Top:} The latent space along with the logarithm volume element and interpolation curves. \textit{Bottom:} The decoded samples along the curves (granularity of 5 time steps).}
    \label{fig: Geodesic Interpolation MNIST}
\end{figure}

\emph{Implementation details:} To asses the quality of the geodesic interpolations obtained with the proposed metric, we also propose to perform a comparison between 1) a vanilla VAE, 2) the metric proposed in \citep{arvanitidis2017latent}, 3) the one exposed in \citep{chen2018metrics} and 4) a RHVAE using the metric as defined in Eq.~\eqref{eq: metric} on two data sets created from MNIST and FashionMNIST. An overview of the training samples is available in Appendix~\ref{app: Geodesic training samples}. Since we have to compute the Jacobian of the generator function and so need differentiable generator functions for 2) and 3) we amend a little bit the model architectures used in Table~\ref{Table: Model architectures} as presented in Appendix~\ref{app: VAE models architectures}. $K$-means algorithm from \citep{scikit-learn} is used to find the centroids in Eq.~\eqref{eq: arvanitidis centroids} of the method proposed by \citet{arvanitidis2017latent}. For each model, we propose to compare the affine interpolation method and the geodesic interpolation with respect to the metric associated to the model. The starting and ending points are the mean values of two encoded samples from the training set i.e. $z_1 = \mu(x_1)$ and $z_2 = \mu(x_2)$ where $(x_1, x_2)$ is rigorously the same for each model. 

\noindent \\
\emph{MNIST:} First, we train the models on a small data set created from 50 samples of the classes \{"0", "1", "2"\} of the MNIST data set. Again the train set is created by selecting randomly 80\% of the data ensuring balanced classes. To allow a fair comparison, each model is trained with the same number of epochs set to 300. Since no clear procedure of training is made available by the authors, model 2) is trained as follows: three quarters of the training time is allocated to fit the inference network and the generator's mean function $\mu_{\theta}$ and the remaining time is used to fit the variance function $\sigma_{\psi}$. As to the parameter $a$ of Eq.~\eqref{eq: arvanitidis centroids} and the number of centroids, again no clear indication is stated and so we use the same values as those they used in their paper ($a=1$ and $K=32$). The latent space learned by the vanilla VAE can be observed in Figure~\ref{fig: VAE Interpolation MNIST} along with two affine interpolations (top). The classic VAE seems to perform very poorly in terms of interpolation as the points along the curve are only a superposition of digits (see orange frames). Ideally, we would expect the starting point to progressively undergo deformation towards the ending point while looking like a digit all along the path. The same experiment is conducted with the models and metrics proposed by \citet{arvanitidis2017latent} and \citet{chen2018metrics}. Figure~\ref{fig: Geodesic Interpolation Arvanitidis MNIST} and Figure~\ref{fig: Geodesic Interpolation Chen MNIST} illustrate the affine and geodesic interpolation obtained with each model along with the learned latent space (top) and the decoded samples along each of the paths (bottom). Interestingly, using the metrics involving the Jacobian of the generator function of the VAE seems to conduct to geodesic paths that are very close to straight lines. This aspect was also noted by \citet{shao2018riemannian} who concluded that the learned manifold has a "surprisingly small curvature" on the data sets they studied. Finally, we compare the former results with those of a RHVAE trained on the same data and with the same number of epochs. The metric temperature is set to $T=1$ and metric regularization to $\lambda = 10^{-2}$. Affine and geodesic interpolations are compared in Figure~\ref{fig: Geodesic Interpolation MNIST}.  Impressively, the curves obtained by geodesic interpolation and using the learned metric are far more meaningful. They clearly remain into the learned manifold since each of the decoded samples can be interpreted as a digit which is progressively distorted. Figure~\ref{fig: Geodesic Interpolation MNIST} (middle) clearly demonstrates that the euclidean distance is not suited to perform such a task. This experiment underlines the usefulness of the introduction of a meaningful metric in the latent space and justifies the modelling of the latent space as a Riemannian manifold.

\noindent \\
\emph{FashionMNIST:} The same models are trained on a \textit{small} data set created with 200 samples from one class of the FashionMNIST data set (\{"sandals"\}) consisting in even more complex shapes. In this experiment, the same number of epochs is used as well and set to 1000. The RHVAE is trained with a temperature fixed to $T=0.5$, $\lambda = 10^{-1}$ and $n_{\mathnormal{lf}}=3$. The geodesic curves obtained using the vanilla VAE and metrics proposed by \citet{arvanitidis2017latent} and \citet{chen2018metrics} are available in Appendix~\ref{ap: Geodesic interpolation Fashion}. Likewise the experiment conducted on MNIST data set, the affine interpolations are not visually satisfying since they always involve sharp transitions leading to samples not having the expected shape. Geodesic paths computed using the metrics proposed by both \citet{arvanitidis2017latent} and \citet{chen2018metrics} are again close to straight lines. 

\begin{figure}[p]
    \centering
    \begin{minipage}[c]{0.32\linewidth}
        \centering
        \subcaption*{}
        \vskip -0.5em
         \includegraphics[scale=0.38]{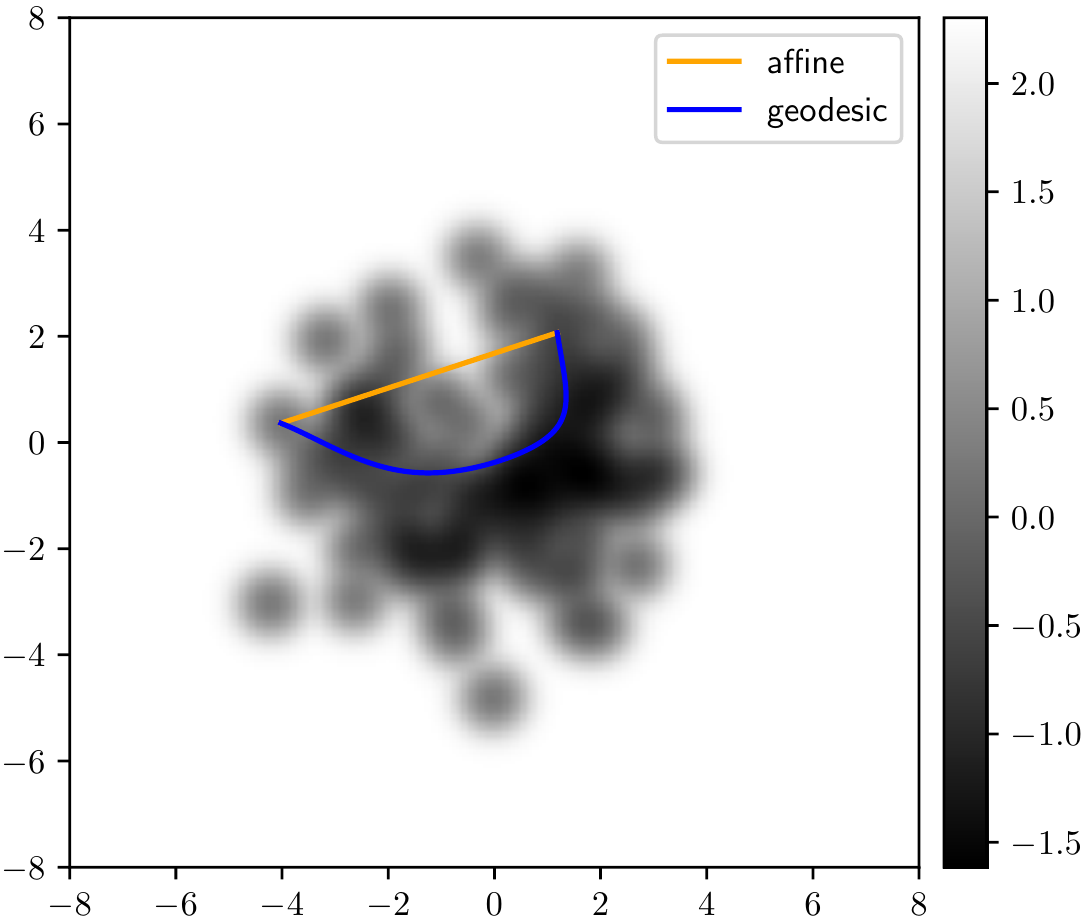}
    \end{minipage}
    \begin{minipage}[c]{0.32\linewidth}
        \centering
        \subcaption*{RHVAE (Ours)}
        \vskip -0.5em
         \includegraphics[scale=0.38]{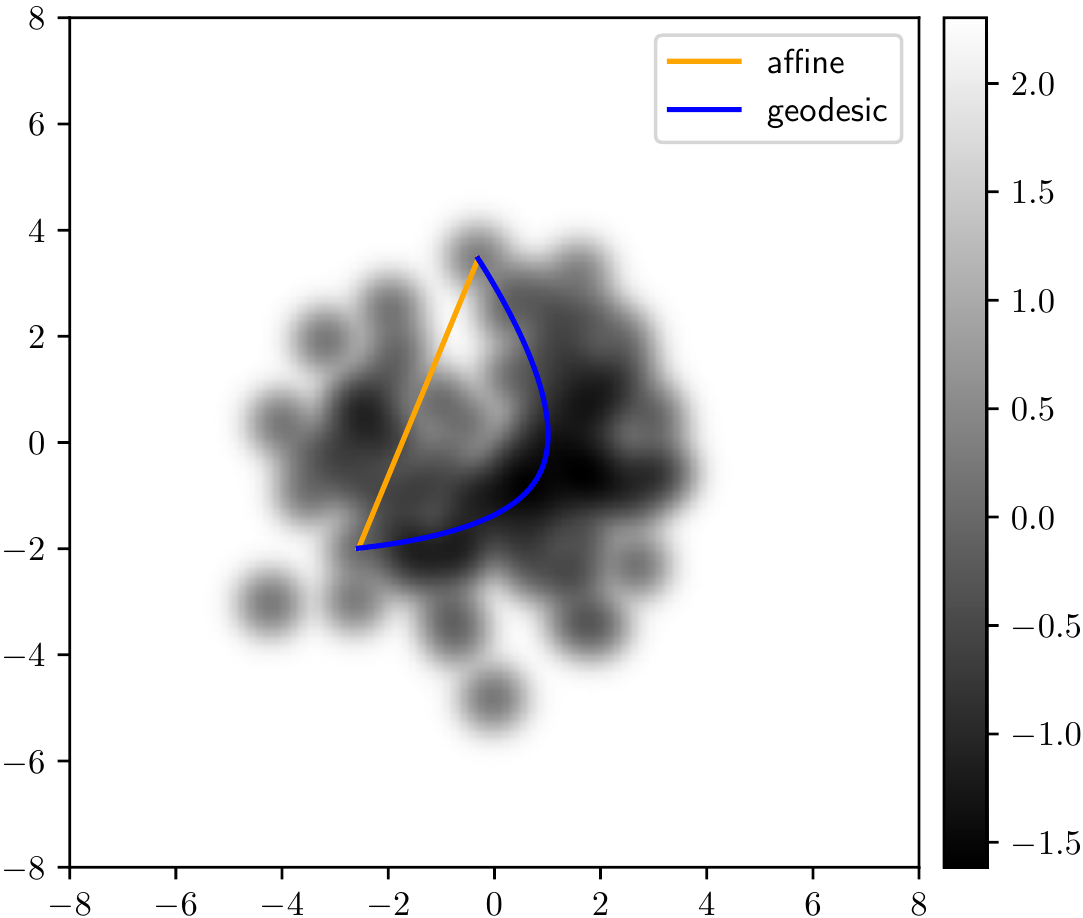}
    \end{minipage}
    \begin{minipage}[c]{0.32\linewidth}
        \centering
        \subcaption*{}
        \vskip -0.5em
         \includegraphics[scale=0.38]{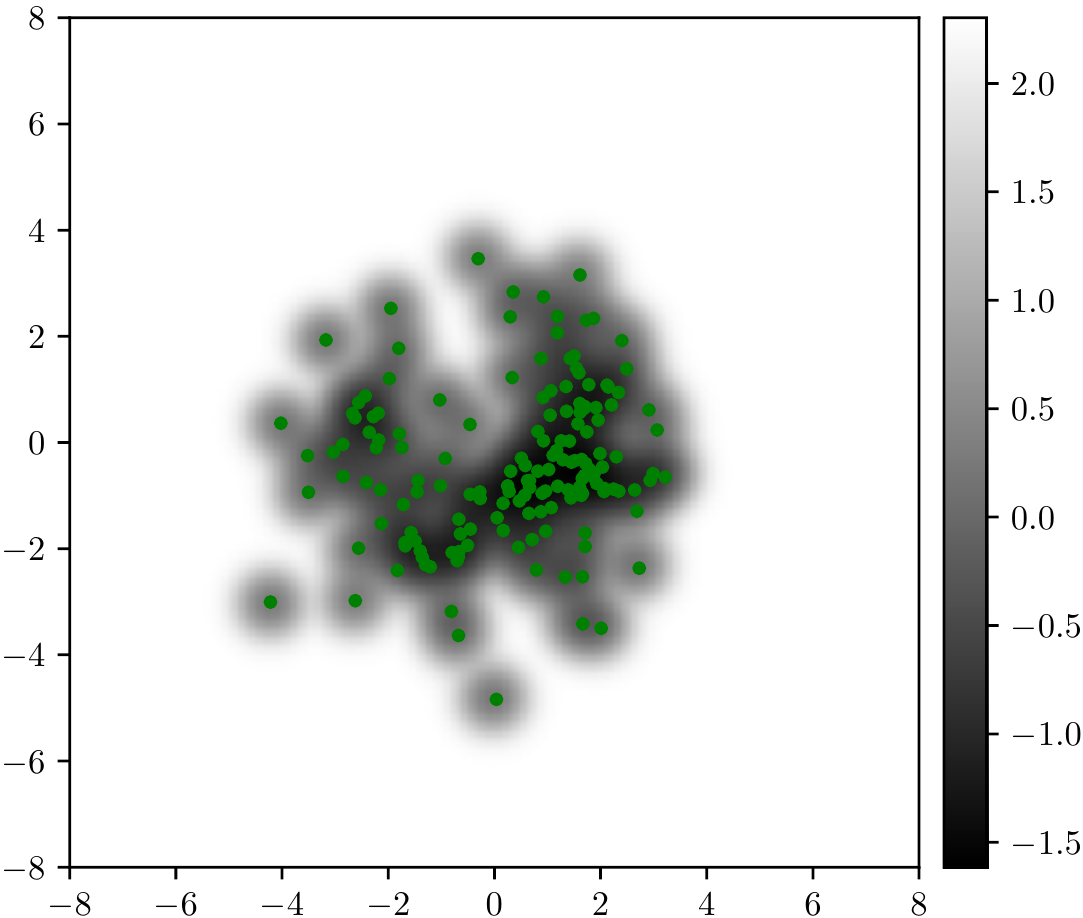}
    \end{minipage}
    \centering
    \vskip 0.5em
    \begin{minipage}[c]{\linewidth}
        \centering
         \includegraphics[scale=0.28]{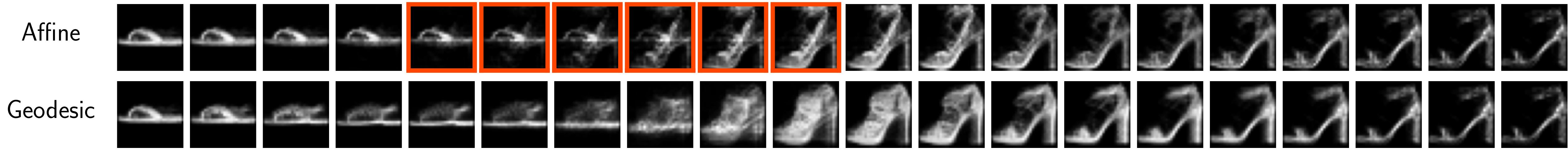}
    \end{minipage}
    \vskip 0.5em
    \centering
    \begin{minipage}[c]{\linewidth}
        \centering
         \includegraphics[scale=0.28]{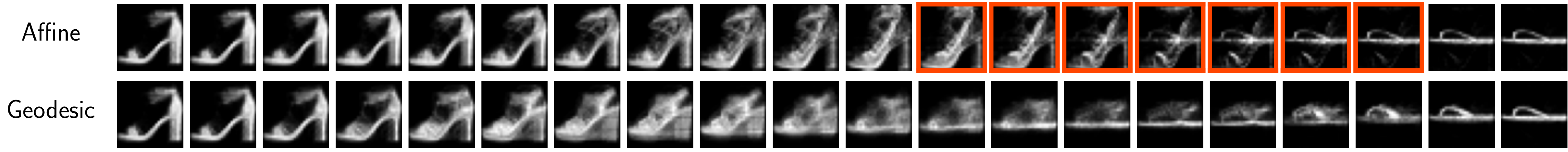}
    \end{minipage}
    \caption{Affine and geodesic interpolations with the proposed RHVAE trained with 160 samples of a single class extracted from the FashionMNIST data set and with 1000 epochs. \textit{Top:} The latent space along with the volume element and interpolation curves. \textit{Bottom:} The decoded samples along the curves (granularity of 5 time steps).}
    \label{fig: Geodesic Interpolation FashionMNIST}
\end{figure}

\begin{remark}
    It can also be noted that these models perform poorly in terms of reconstruction on this data set as highlighted by the last decoded sample of the top row in Figure~\ref{fig: Geodesic Interpolation Arvanitidis FashionMNIST} and Figure~\ref{fig: Geodesic Interpolation Chen FashionMNIST} for example. Recall that this decoded sample is obtained by decoding the mean $\mu(x_i)$ of the distribution $\mathcal{N}(\mu(x_i), \Sigma(x_i))$ associated to the data point $x_i$ extracted from the training set. Hence, the decoded sample is expected to be close to $x_i$ likewise the vanilla VAE (see Figure~\ref{fig: VAE Interpolation FashionMNIST}). We note that this is due to the slight change in the activation function. Recall that the only change we made between the vanilla VAE and the one used to reproduce \citet{chen2018metrics} model is to use the $Softplus$ activation function for the generator function instead of $ReLu$. Finally constraining the model architecture by imposing twice differentiable functions may have a strong impact on the overall model quality as well.
\end{remark}
Then, the interpolations obtained using the proposed RHVAE are presented in Figure~\ref{fig: Geodesic Interpolation FashionMNIST}. Again, all along the geodesic path, the starting image is progressively distorted and impressively even in a very small latent space dimension (2) each image looks like a shoe whereas affine interpolations still perform poorly (see orange frames). Finally, the proposed latent space modelling though the learned metric makes possible meaningful interpolations even with quite complex shapes.

We also test the model on the Olivetti faces data set
\citep{olivettifaces} and try to see if geodesic interpolations between faces remain meaningful which is actually the case. Results can be found in Appendix~\ref{app: Geodesic interpolations Olivetti}.

\begin{figure}[t]
    \centering
    \begin{minipage}[c]{0.32\linewidth}
        \centering
        \subcaption*{}
        \vskip -0.5em
         \includegraphics[scale=0.38]{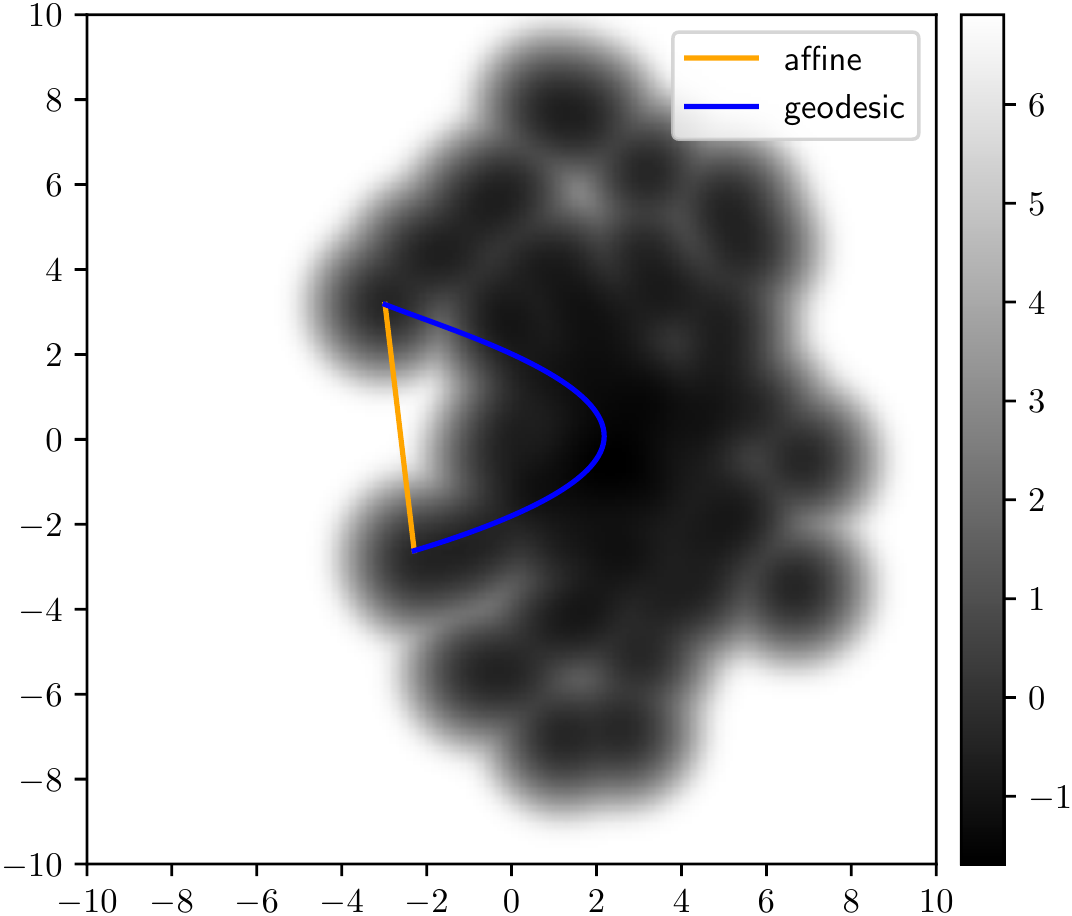}
    \end{minipage}
    \begin{minipage}[c]{0.32\linewidth}
        \centering
        \subcaption*{RHVAE (Ours)}
        \vskip -0.5em
         \includegraphics[scale=0.38]{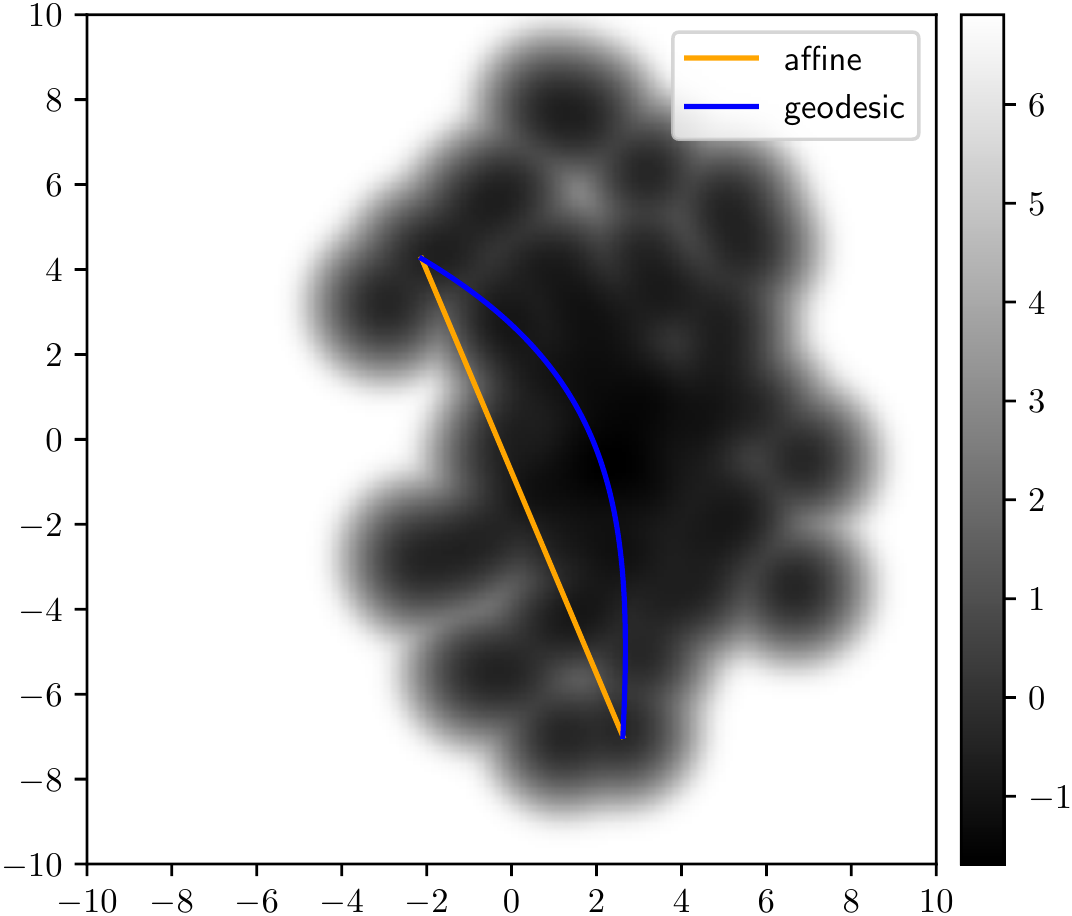}
    \end{minipage}
    \begin{minipage}[c]{0.32\linewidth}
        \centering
        \subcaption*{}
        \vskip -0.5em
         \includegraphics[scale=0.38]{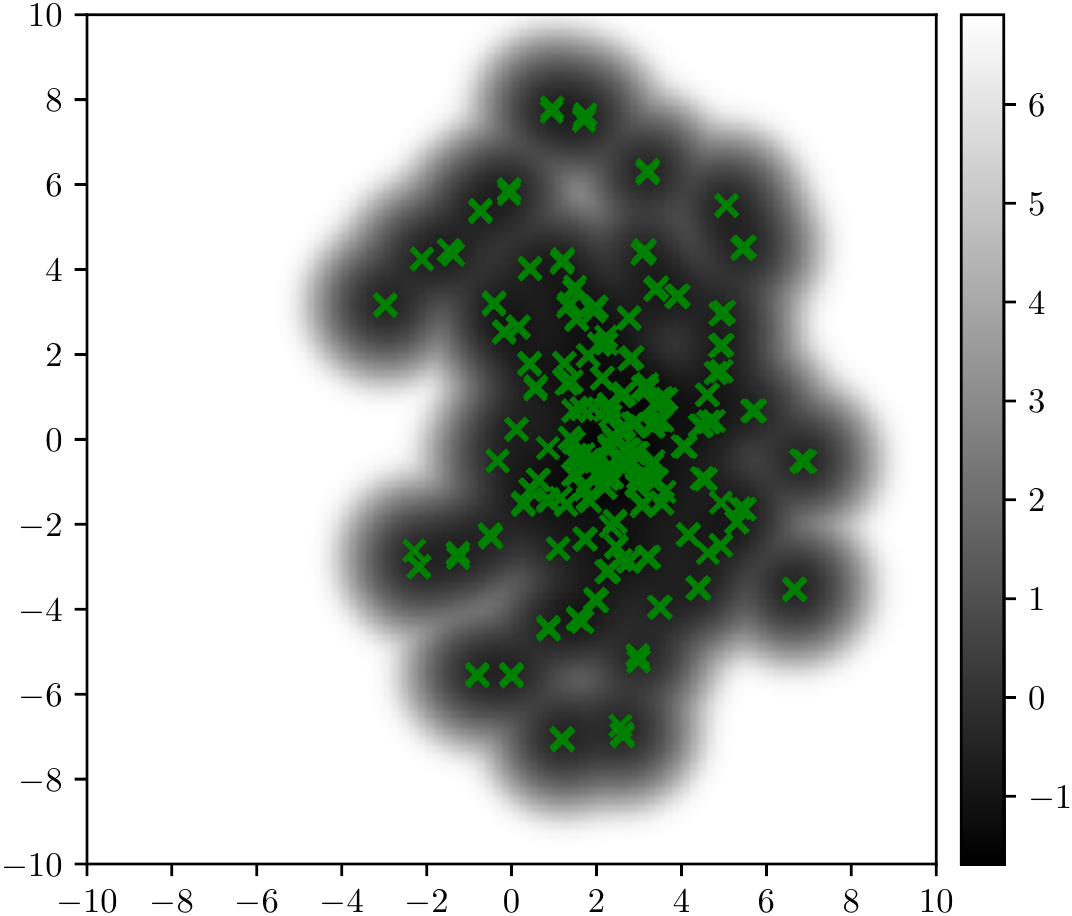}
    \end{minipage}
    \vskip 0.5em
    \centering
    \begin{minipage}[c]{\linewidth}
        \centering
         \includegraphics[scale=0.28]{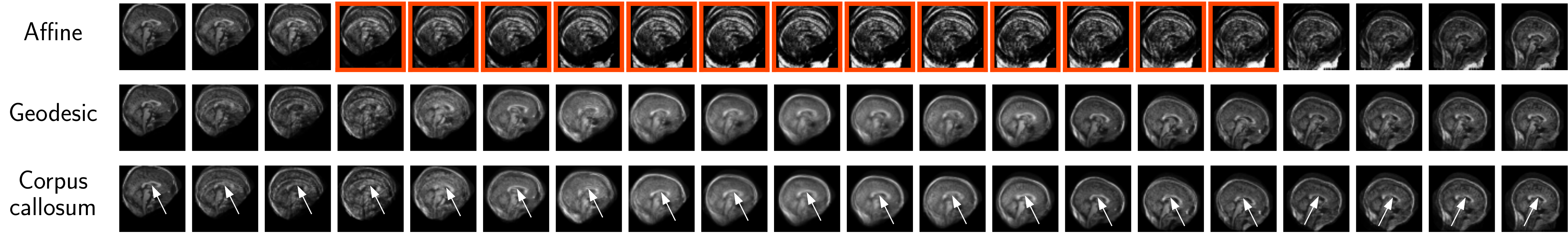}
    \end{minipage}
    \vskip 0.5em
    \centering
    \begin{minipage}[c]{\linewidth}
        \centering
         \includegraphics[scale=0.28]{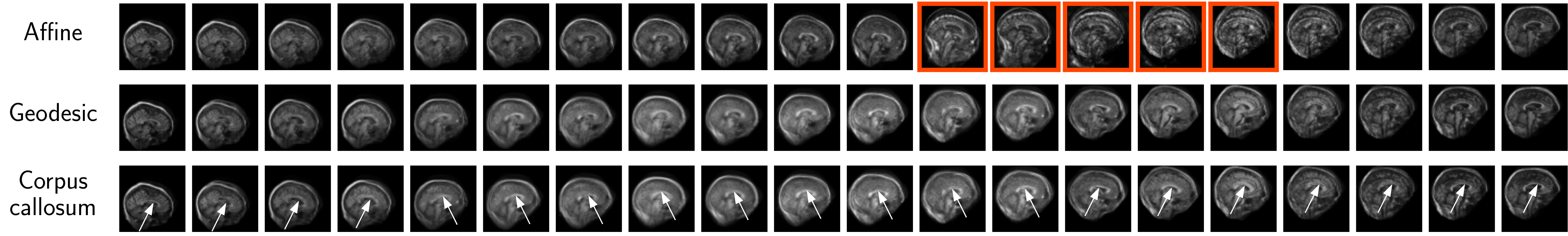}
    \end{minipage}
    \caption{ Affine and geodesic interpolations with the proposed RHVAE trained with raw brain sagital MRI images extracted from OASIS \citep{doi:10.1162/jocn.2007.19.9.1498} database. \textit{Top:} The latent space along with the logarithm volume element and interpolation curves. \textit{Bottom:} The decoded samples along the curves (granularity of 5 time steps).}
    \label{fig: Geodesic Interpolation OASIS}
\end{figure}

\noindent \\
\emph{OASIS Raw Images:} Finally, we try to perform interpolations with the proposed model on a complex database where the variability of shapes is difficult to understand. To do so, we elect the OASIS data set \citep{doi:10.1162/jocn.2007.19.9.1498} and create a data set of 418 raw sagital defaced MRI images. Each image is down-sampled from 256x256 to 100x100 using bi-linear interpolation and fed to the RHVAE. An overview of the training samples is available in Appendix~\ref{app: Geodesic training samples}. This data set is quite challenging since it presents complex shapes which are not always located in the middle of the image. Moreover, we decide not to apply any further pre-processing step such as normalization so we can see how the proposed model would behave with such data. The model is trained with a temperature $T=0.8$, $\lambda=10^{-3}$, $\varepsilon_{\mathnormal{lf}}=10^{-3}$ and $n_{\mathnormal{lf}}=5$. Again, we try to compute the affine and geodesic curves between points in the latent space and display the results in Figure~\ref{fig: Geodesic Interpolation OASIS}. As expected the affine interpolation performs quite poorly since most of the decoded samples within the orange frames are only a superposition of two brains and do not have any physiological meaning. Surprisingly, geodesic interpolations are still able to provide us with satisfying interpolations since each decoded sample along the curve does look like an image which could have been part of the data set. Moreover, the orientation of the brain changes smoothly from an image to another as highlighted in Figure~\ref{fig: Geodesic Interpolation OASIS}. Even more appealing is the fact that we can clearly distinguish the corpus callosum (indicated by the white arrows) on each of the samples along the curves. Finding such a fine detail in the geodesic interpolants shows that a structuring of the latent space is now made possible thanks to the proposed metric and would allow us to produce more realistic samples.

\subsection{Generation} \label{Sec: Generation}

\begin{figure}[p]
    \centering
    \begin{minipage}[c]{0.32\linewidth}
        \centering
        \subcaption*{Training samples}
        \vskip -0.5em
         \includegraphics[scale=0.33]{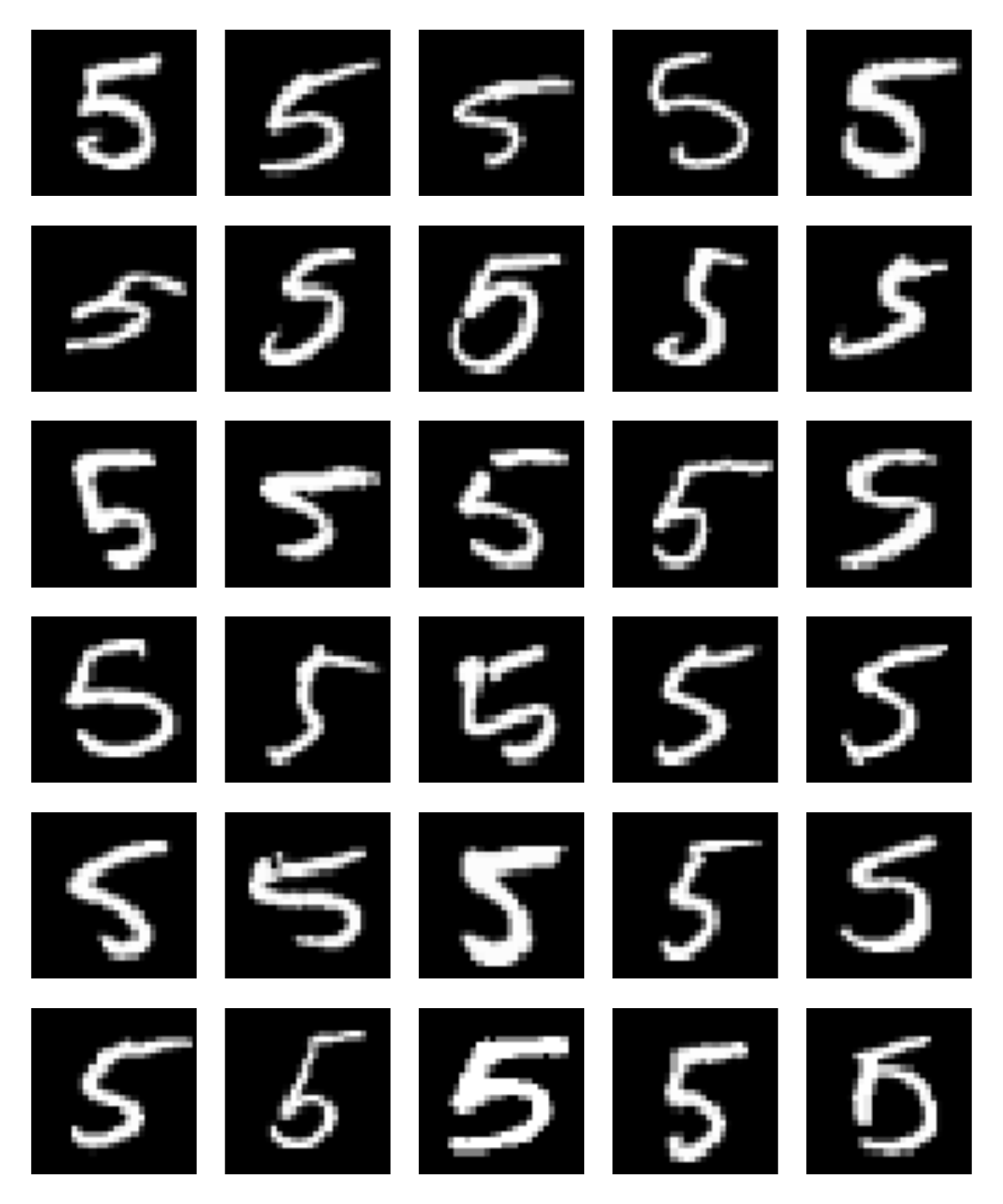}
    \end{minipage}
    \begin{minipage}[c]{0.32\linewidth}
        \centering
        \subcaption*{VAE}
        \vskip -0.5em
         \includegraphics[scale=0.33]{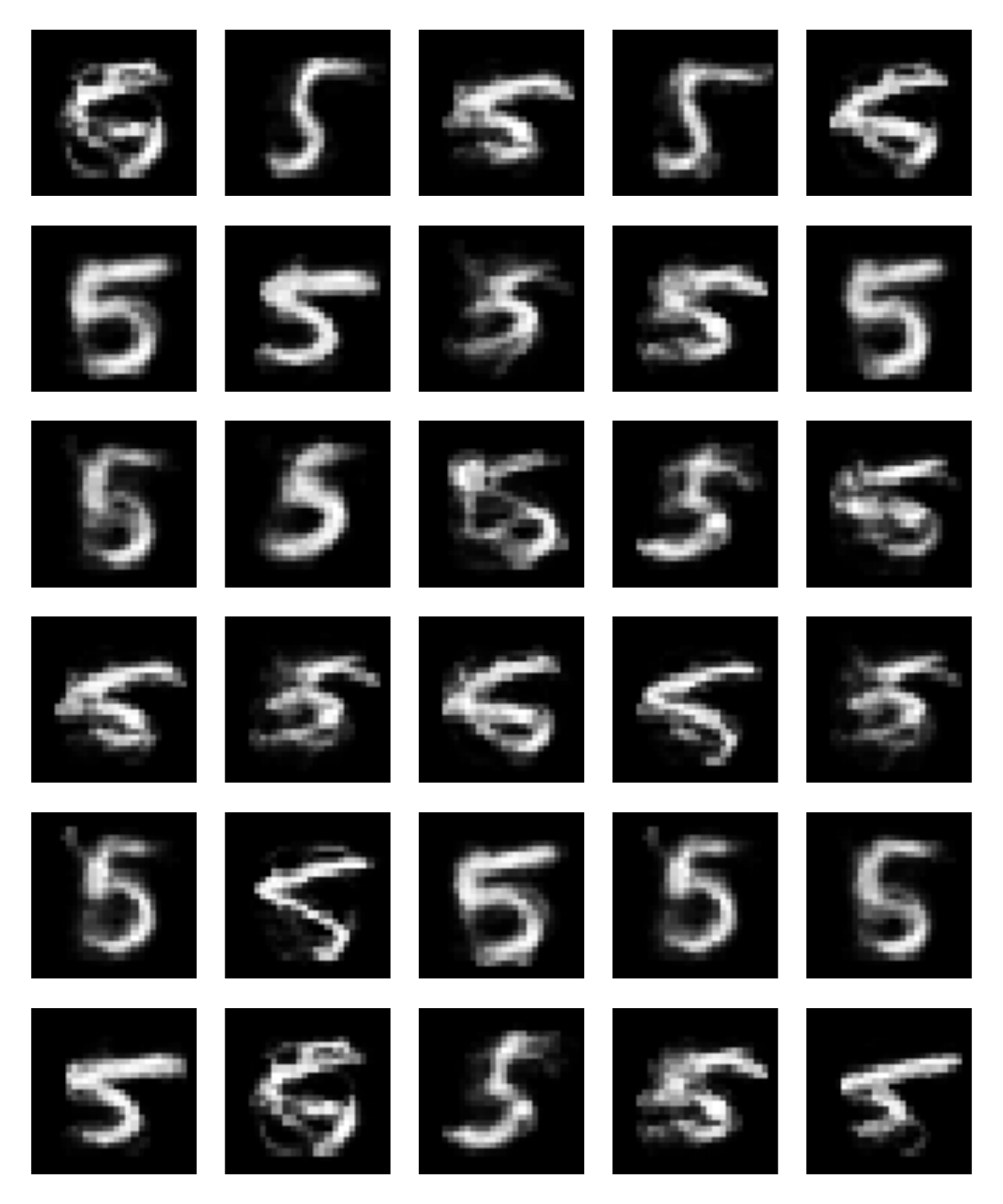}
    \end{minipage}
    \begin{minipage}[c]{0.32\linewidth}
        \centering
        \subcaption*{RHVAE (Ours)}
        \vskip -0.5em
         \includegraphics[scale=0.33]{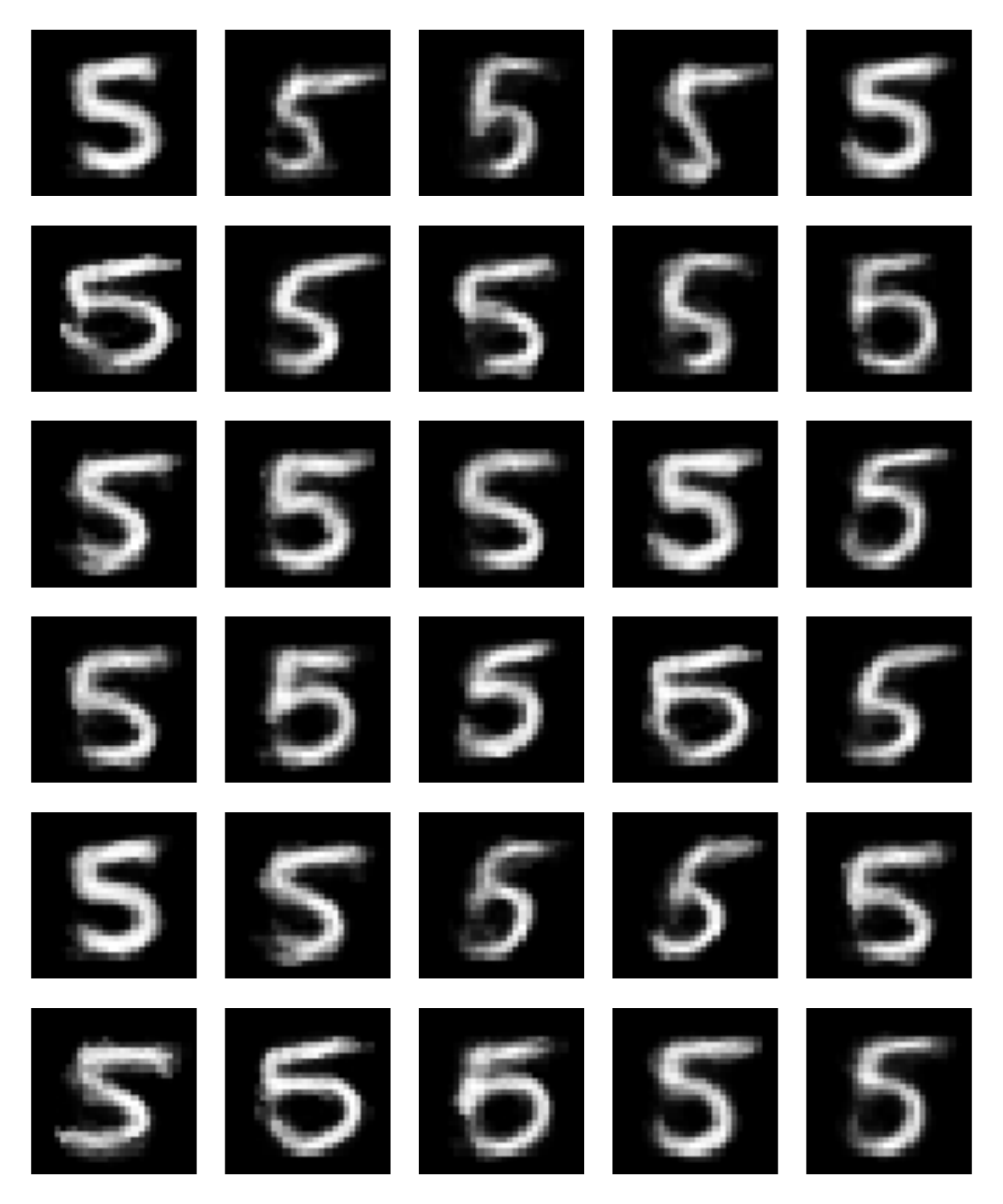}
    \end{minipage}
    \quad
    \begin{minipage}[c]{0.32\linewidth}
        \centering
         \includegraphics[scale=0.33]{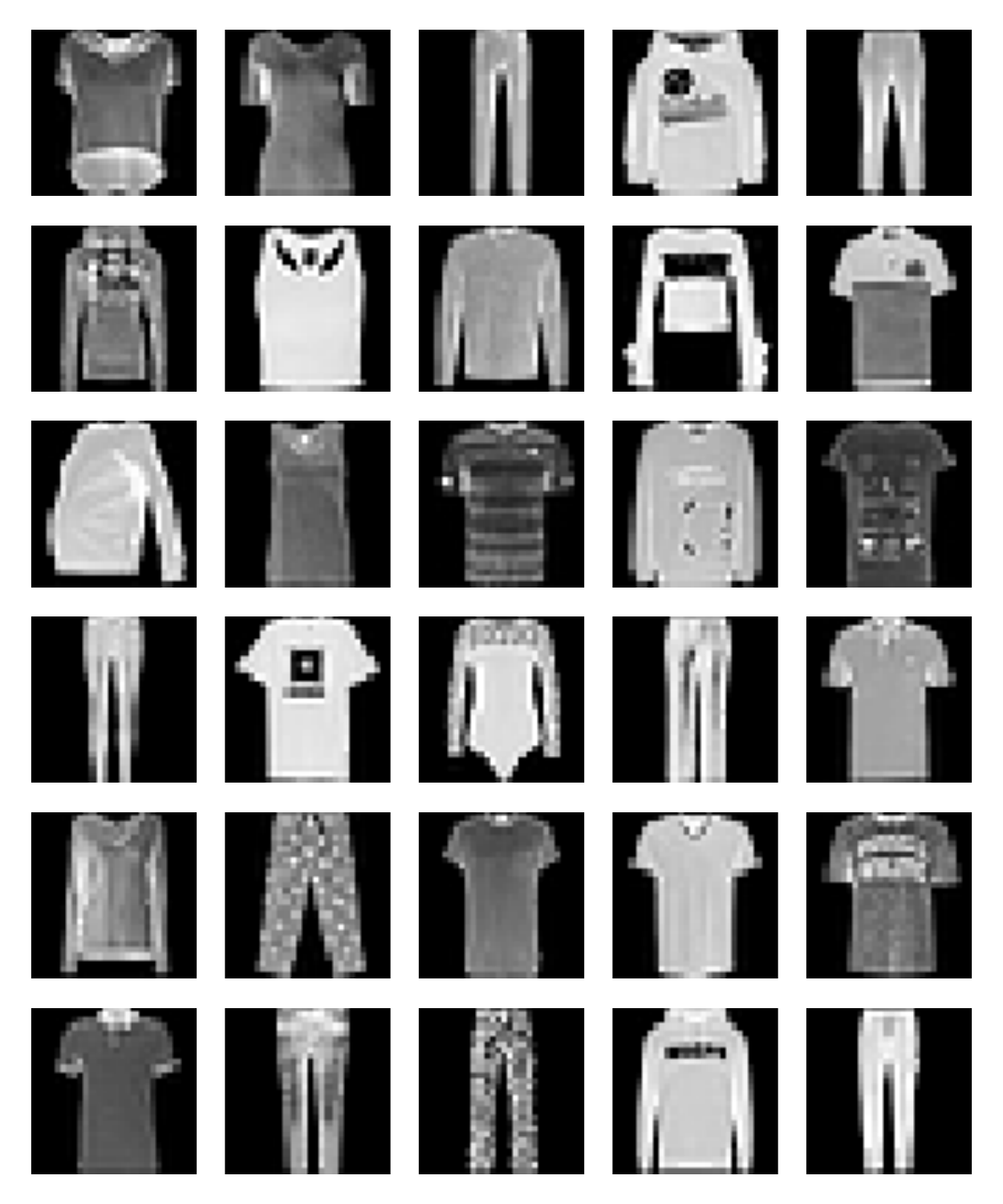}
    \end{minipage}
    \begin{minipage}[c]{0.32\linewidth}
        \centering
         \includegraphics[scale=0.33]{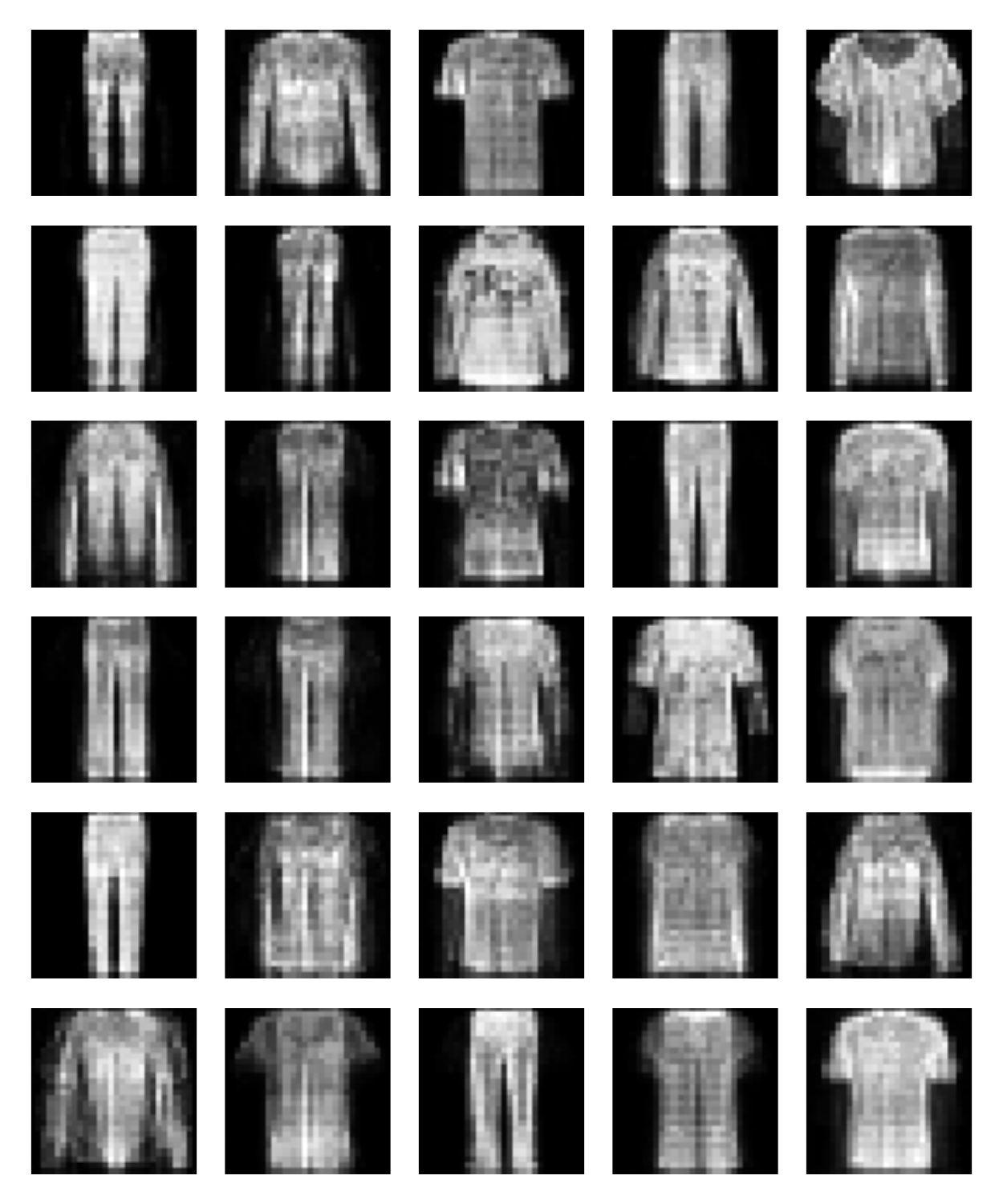}
    \end{minipage}
    \begin{minipage}[c]{0.32\linewidth}
        \centering
         \includegraphics[scale=0.33]{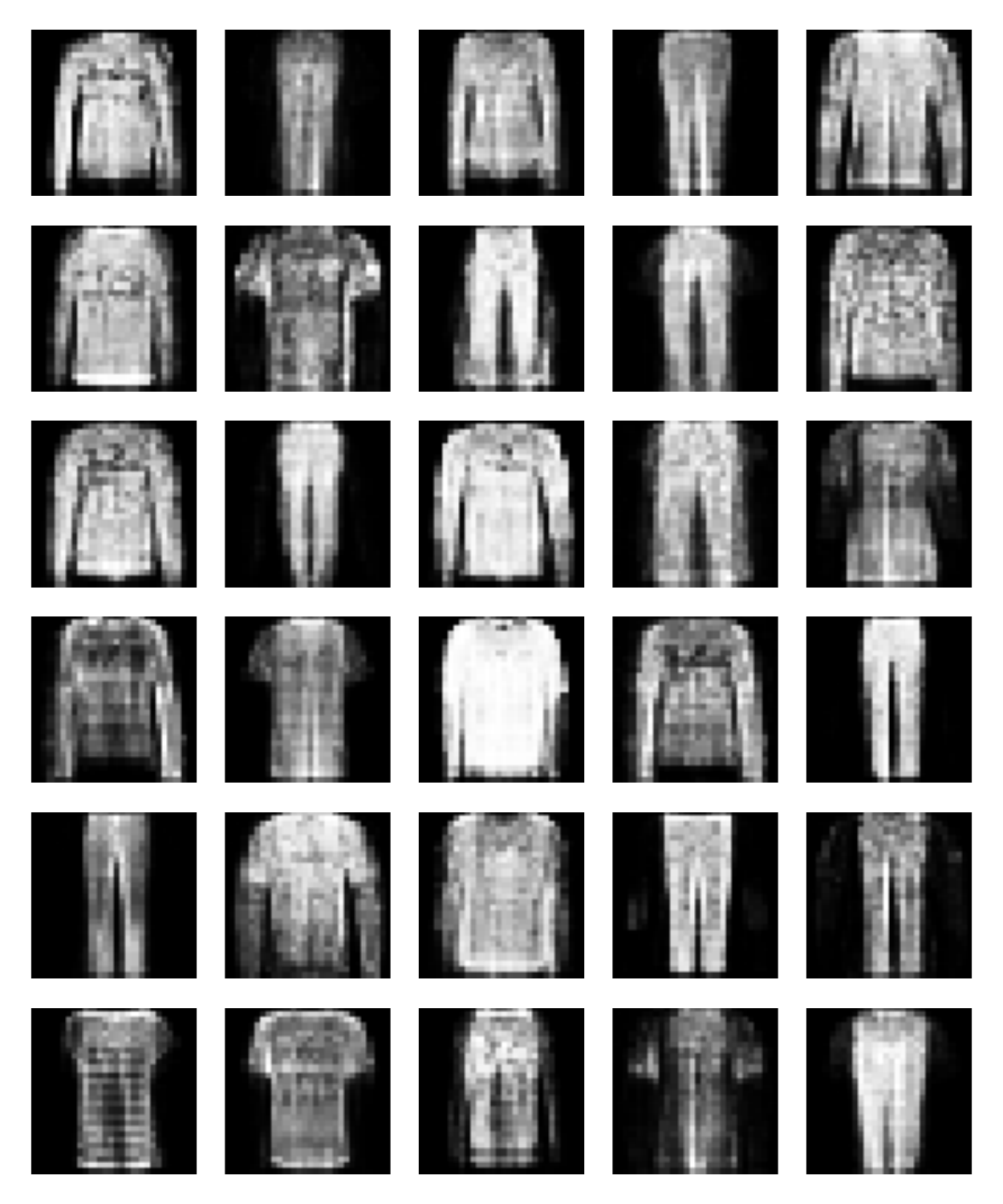}
    \end{minipage}
    \quad
        \begin{minipage}[c]{0.32\linewidth}
        \centering
         \includegraphics[scale=0.33]{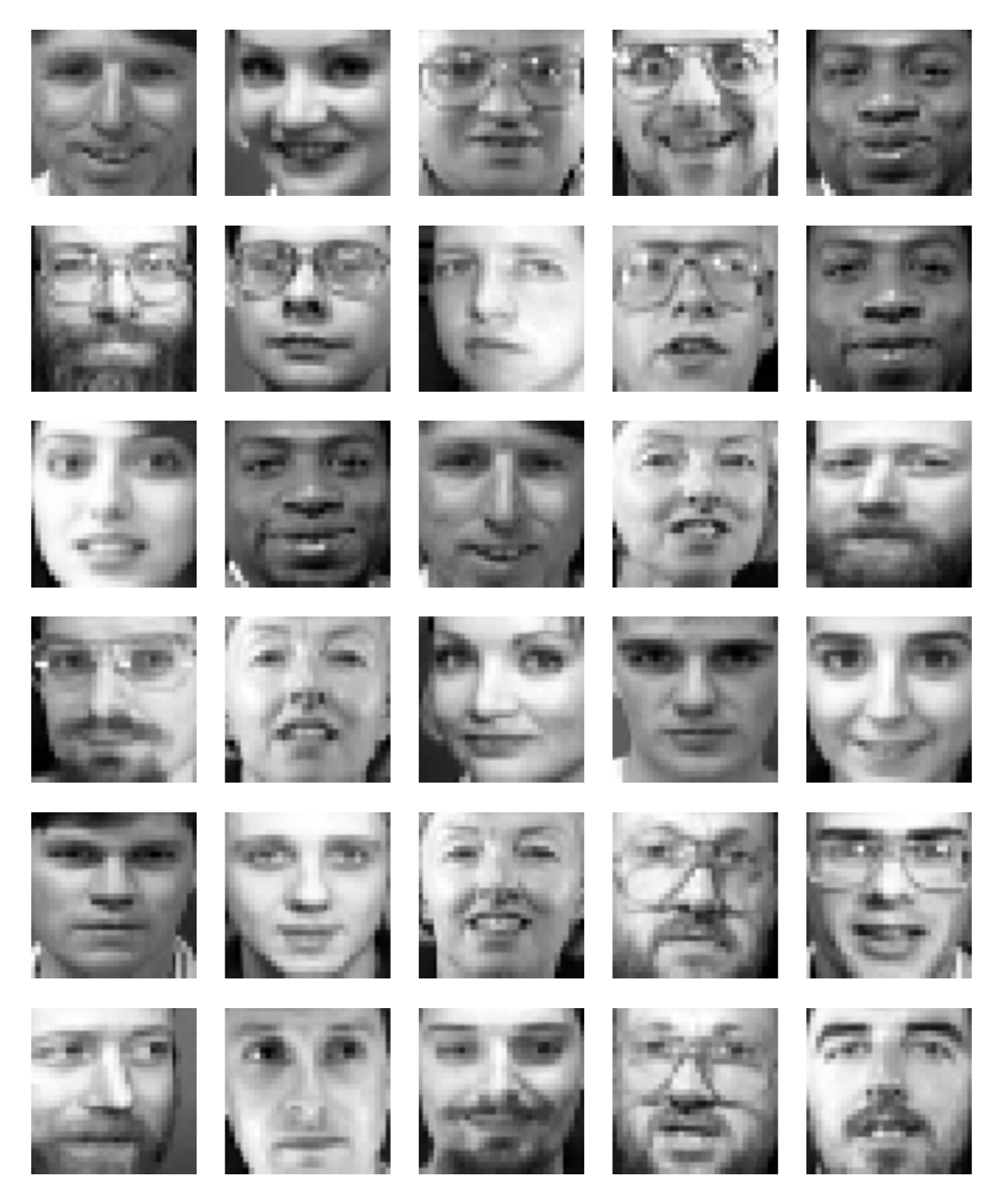}
    \end{minipage}
    \begin{minipage}[c]{0.32\linewidth}
        \centering
         \includegraphics[scale=0.33]{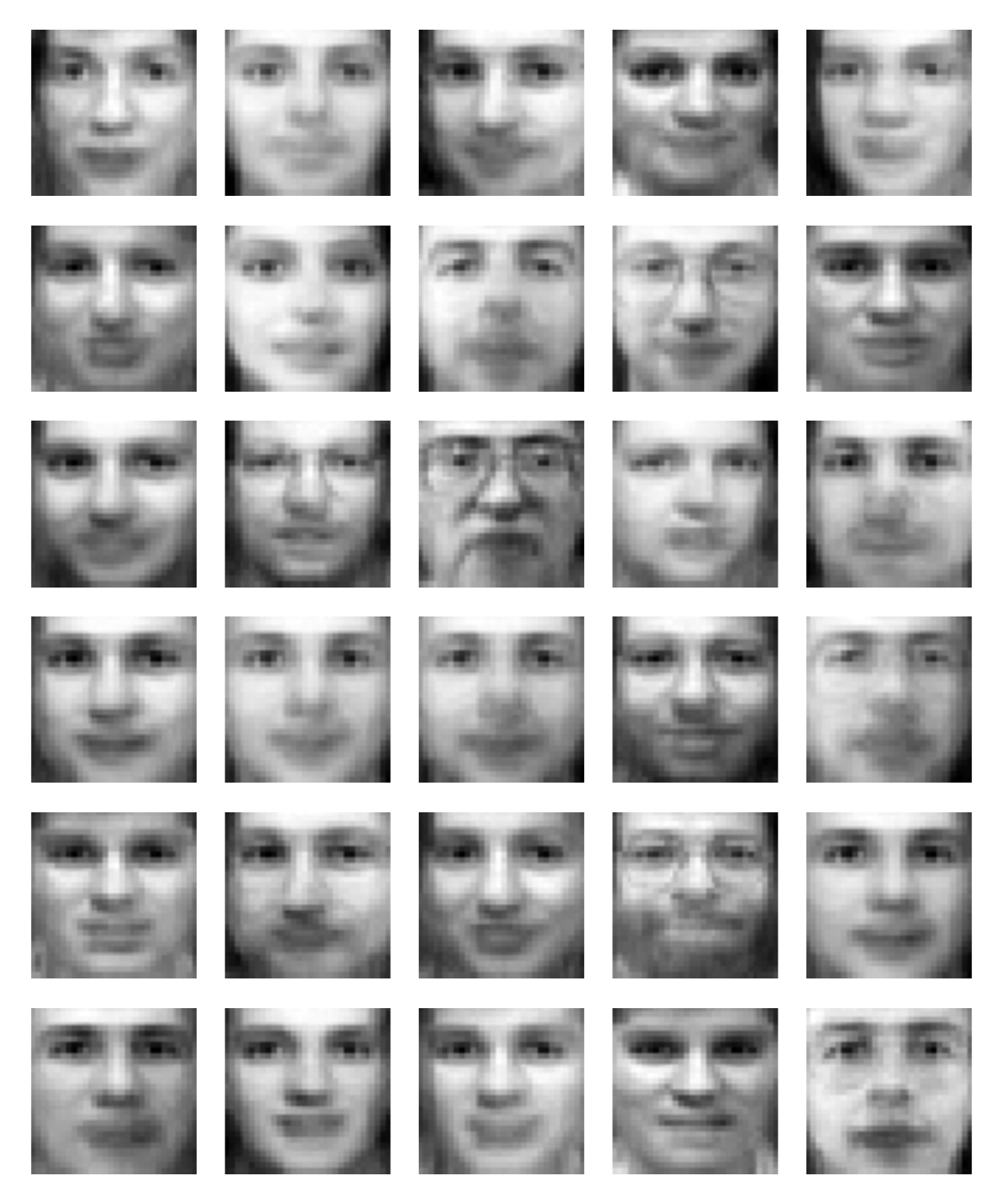}
    \end{minipage}
    \begin{minipage}[c]{0.32\linewidth}
        \centering
         \includegraphics[scale=0.33]{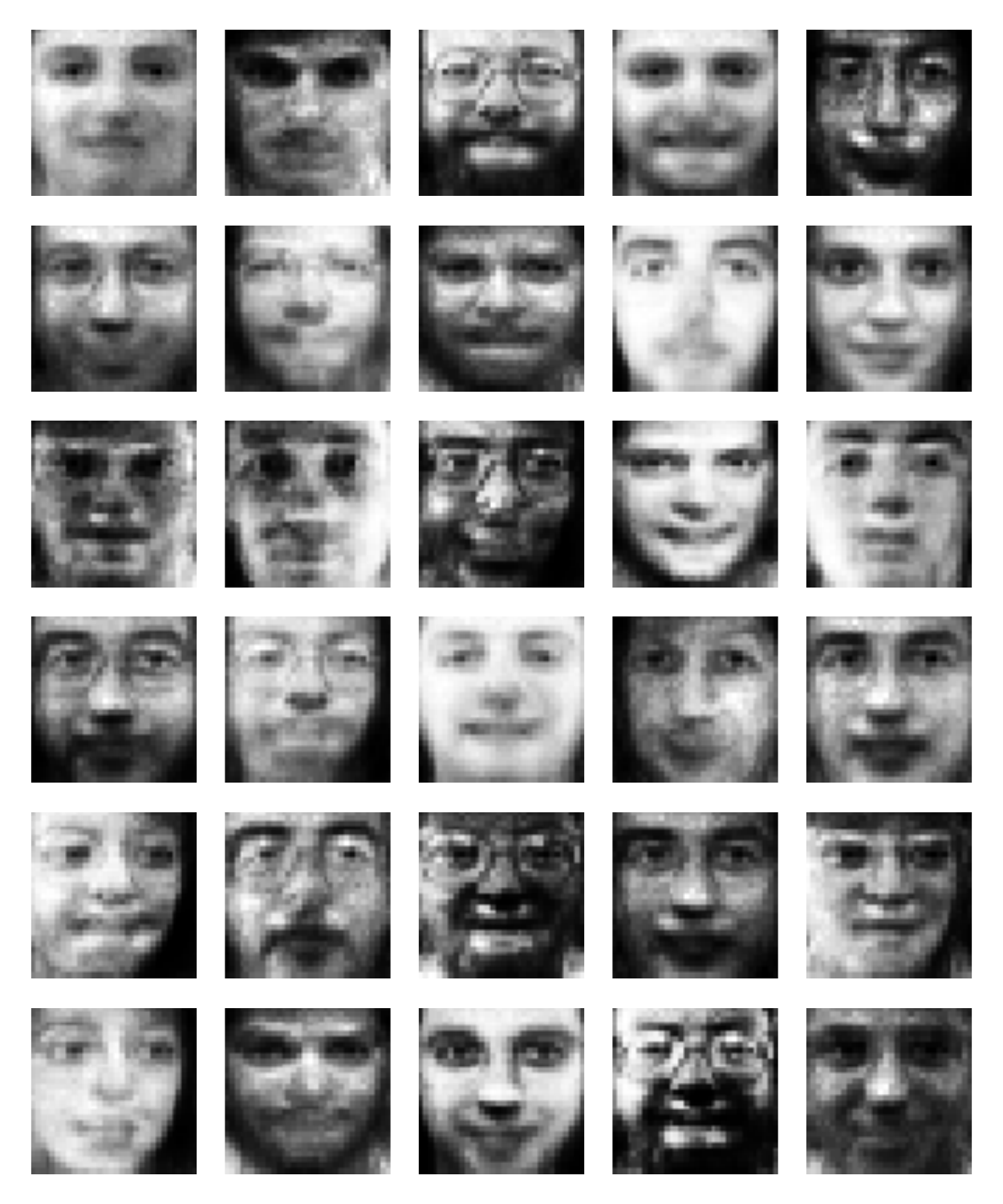}
    \end{minipage}
    \caption{\textit{Top row: } Training samples from a data set created from the class \{"5"\} of the MNIST data set and composed by 160 samples (left) along with samples generated from a vanilla VAE (middle) and our model (right). \textit{Middle row :} Training samples from a data set created by randomly selecting 80\% of 150 samples from the equally distributed classes \{"T-shirt", "Trouser", "Pullover"\} of the FashionMNIST data set (left) along with samples generated from a vanilla VAE (middle) and our model (right). \textit{Bottom row :} Training samples from a data set created by randomly selecting 80\% of the Olivetti data set (left) along with samples generated from a vanilla VAE (middle) and our model (right).}
    \label{fig: Generation}
\end{figure}

In this section, we compare the generative capacity of the proposed method to a vanilla VAE on several data sets. The parameters used are available in Appendix~\ref{app: Generation parameters}.

\noindent \\
\emph{MNIST:} First, we consider a single class small data set extracted from the MNIST data set and composed by 160 randomly selected training samples. An early stopping strategy is employed for each model and consists in stopping the training is the loss does not improve in 20 epochs. The batch size is set to 80 and we display the generated samples in Figure~\ref{fig: Generation} (top). On the left are presented 30 training samples, in the middle samples generated by the vanilla VAE along with samples generated by our model on the right. In this experiment, the vanilla VAE is unable to generate realistic samples since most of them are most of the time very blurry. This is due to the very small number of training samples. Impressively, the RHVAE is still able to generate quite convincing different samples which do not seem to be similar to the training data. This observation goes in the sense of the one done in Section~\ref{Sec: Auto-Encoding comparision} where it was demonstrated that using a {\it geometry-aware} normalizing flow to tweak the approximate posterior distribution indeed improves the model. 

\noindent \\
\emph{FashionMNIST:} The same experiment is realized on a data set created from the FashionMNIST data set. It consists in selecting 50 samples of 3 classes \{"T-shirt", "Trouser", "Pullover"\} from the FashionMNIST data set. We train the model on 80\% of the data randomly selected from the initial data set and ensuring balanced classes. Figure~\ref{fig: Generation} (middle row) highlights 30 training samples (left), 30 samples generated by the classic VAE and 30 samples generated from our model (right). At first sight, since the shape of the data remains quite simple, the VAE is able to generate quite realistic samples so does the RHVAE. However, as highlighted in the Figure~\ref{fig: Generation} (left of middle row) there exists a wide range of colours (shades of grey) and patterns in the training set. Unfortunately, the VAE is unable to generate such details and finally the color of the samples remains quite similar across the generated images. Interestingly, the RHVAE seems to be able to generate various range of colours (see the last row for example) with different patterns matching better the true essence of the training data. This is even more striking in the following experiment. 

\noindent \\
\emph{Olivetti Faces:} Finally, we decide to compare the models on the Olivetti faces data set composed by 400 images of faces. We select randomly 80\% of the initial data set to create the training set and fit the model until the $ELBO$ does not improve for 50 epochs. Training samples are down-sampled from 64x64 to 32x32 using bi-linear interpolation. Generated samples can be observed in Figure~\ref{fig: Generation} (bottom row). Likewise the previous experiment, the VAE seems able to generate faces.  However, it is not able to generate samples with different skin colours, different lightning or diverse facial expressions as in the training set. Moreover, finer details such as the mouth are most of the time blurry. Impressively, the RHVAE we propose is able to generate sharper samples having a wide range of facial expressions (smiles, anger...) and head orientations. Even more appealing, it is able to generate very different images with various lightnings and skin colours.

\subsection{Clustering} \label{Sec: Clustering}
Finally, the clustering ability of the model is assessed on both synthetic and real data sets.
\subsubsection{Synthetic Data}
\begin{figure}[t]
    \centering
    \begin{minipage}[c]{0.32\linewidth}
        \centering
        \subcaption*{True labels}
        \vskip -0.5em
         \includegraphics[scale=0.38]{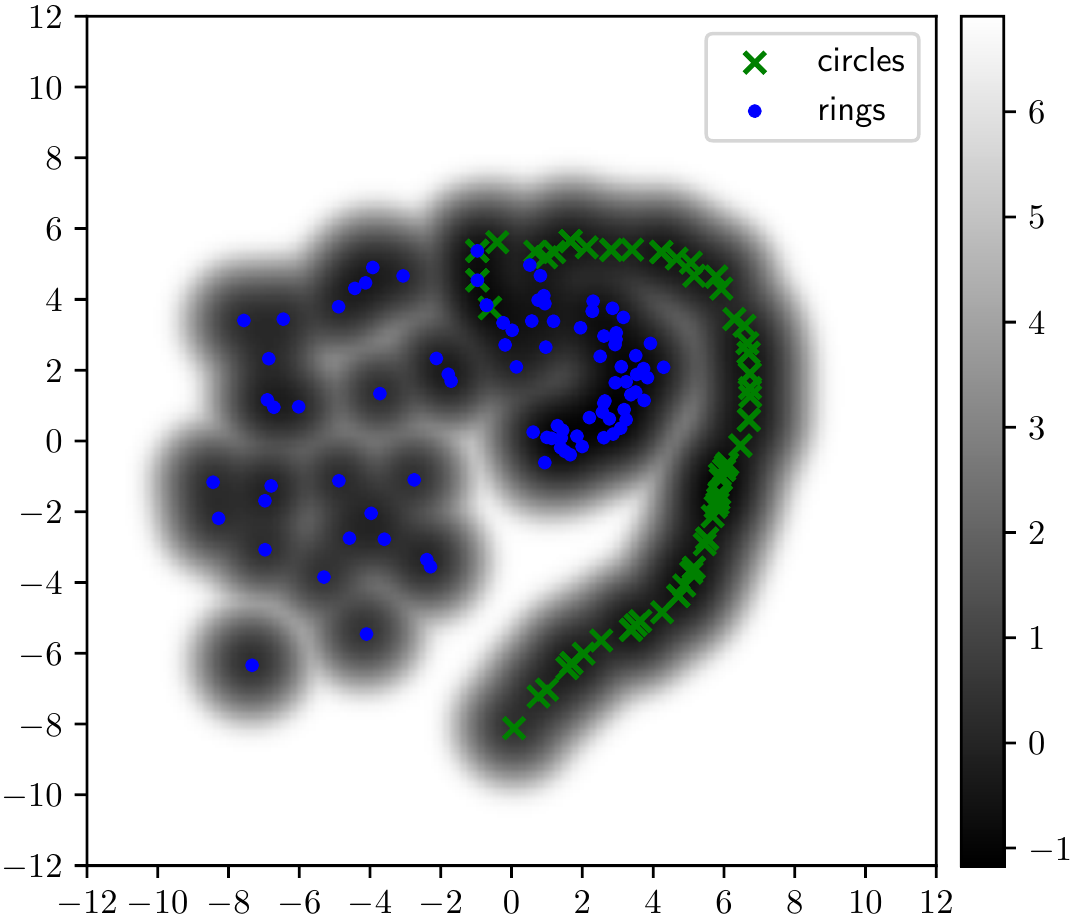}
    \end{minipage}
    \begin{minipage}[c]{0.32\linewidth}
        \centering
        \subcaption*{Euclidean $k$-medoids}
        \vskip -0.5em
        \includegraphics[scale=0.38]{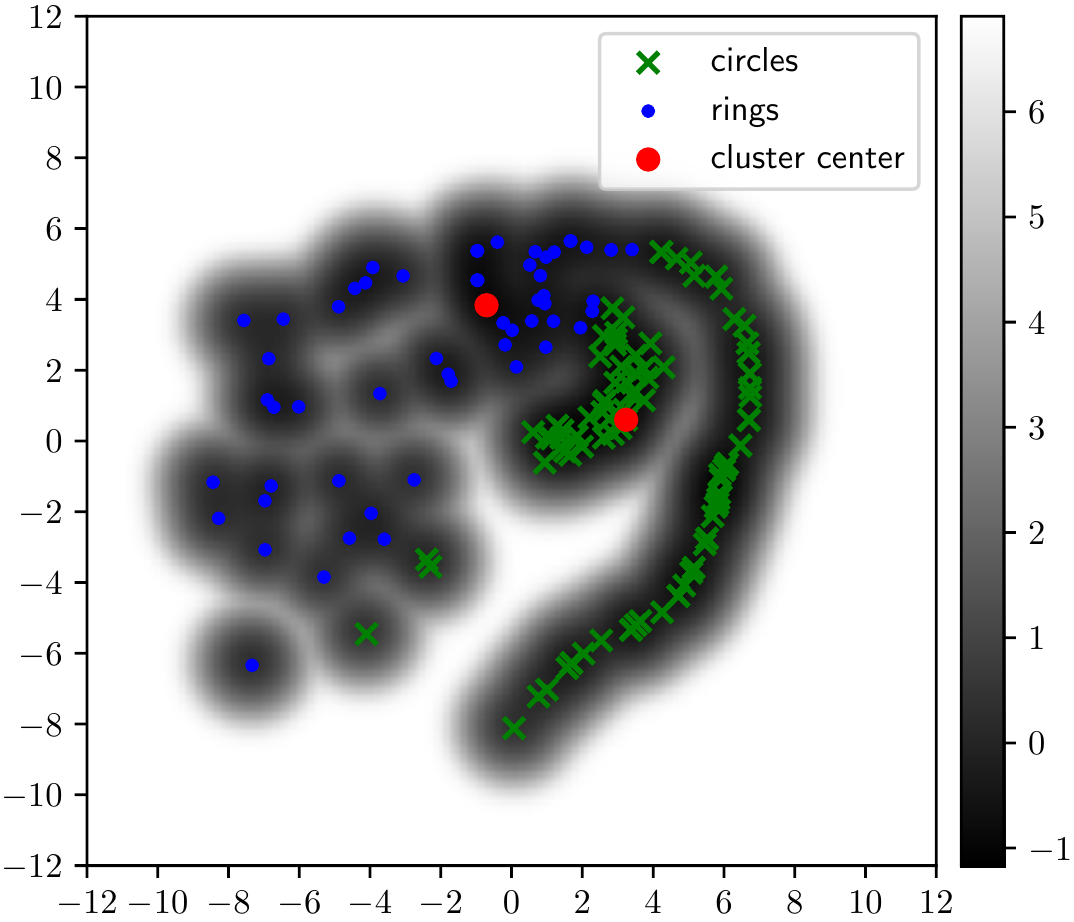}
    \end{minipage}
    \begin{minipage}[c]{0.32\linewidth}
        \centering
        \subcaption*{Riemannian $k$-medoids}
        \vskip -0.5em
         \includegraphics[scale=0.38]{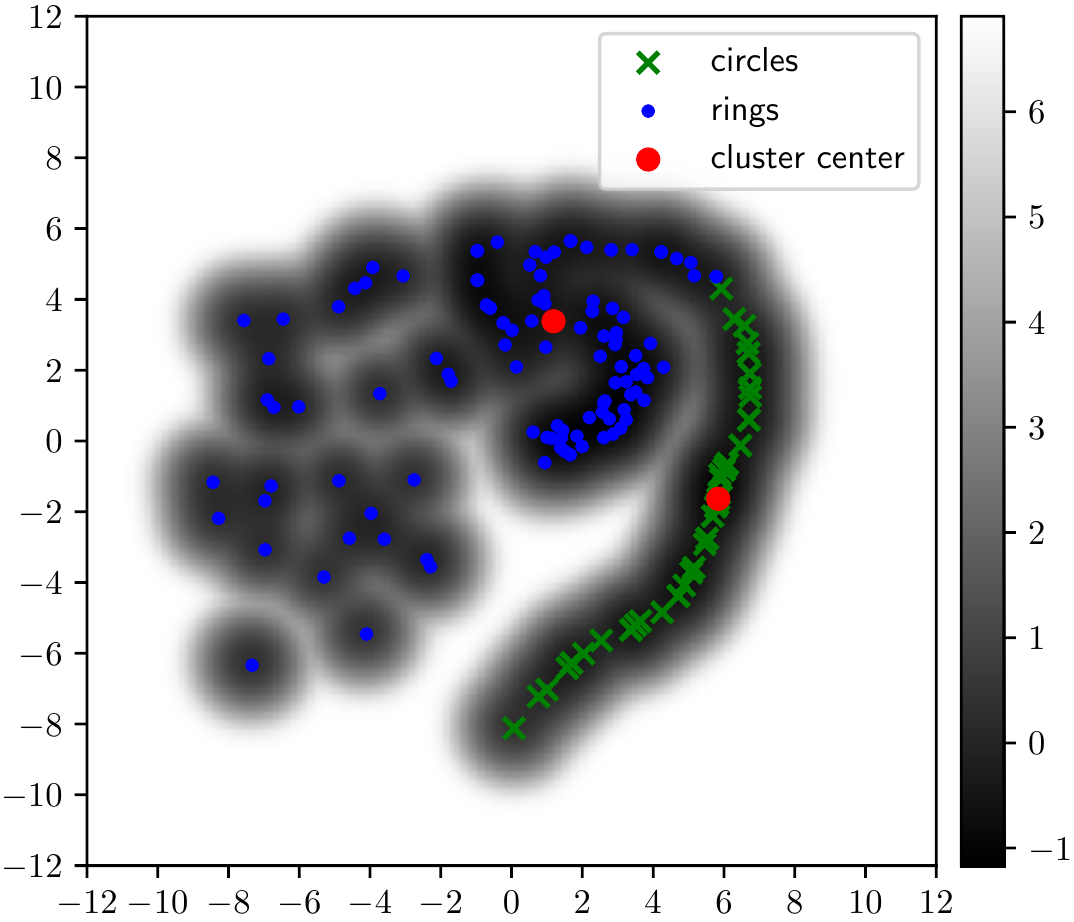}
    \end{minipage}
    \caption{Latent space of a RHVAE trained  with  a synthetic data set composed by 100 rings and  100 circles of different diameters and thicknesses along with the log of the learned volume element (left). The clusters found by a $k$-medoids algorithm using the euclidean metric (middle) and a $k$-medoids algorithm using our Riemannian metric.   The model is trained with $n_{\mathnormal{lf}} =3$, $\varepsilon_{\mathnormal{lf}} = 10^{-2}$, $\lambda = 10^{-3}$ and a metric temperature $T = 0.8$.}
    \label{fig: Clustering shapes}
\end{figure}

\begin{figure}[t]
    \centering
    \begin{minipage}[c]{0.32\linewidth}
        \centering
         \includegraphics[scale=0.38]{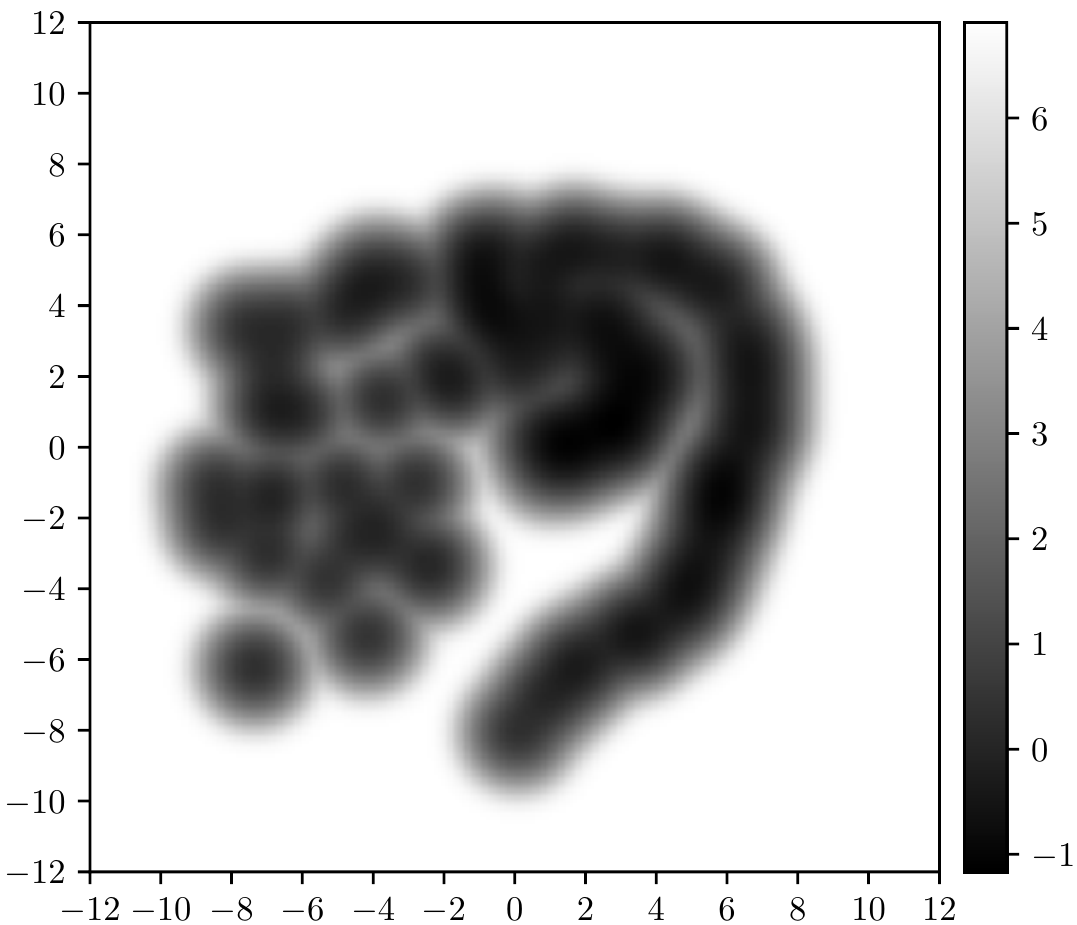}
    \end{minipage}
    \begin{minipage}[c]{0.32\linewidth}
        \centering
        \includegraphics[scale=0.38]{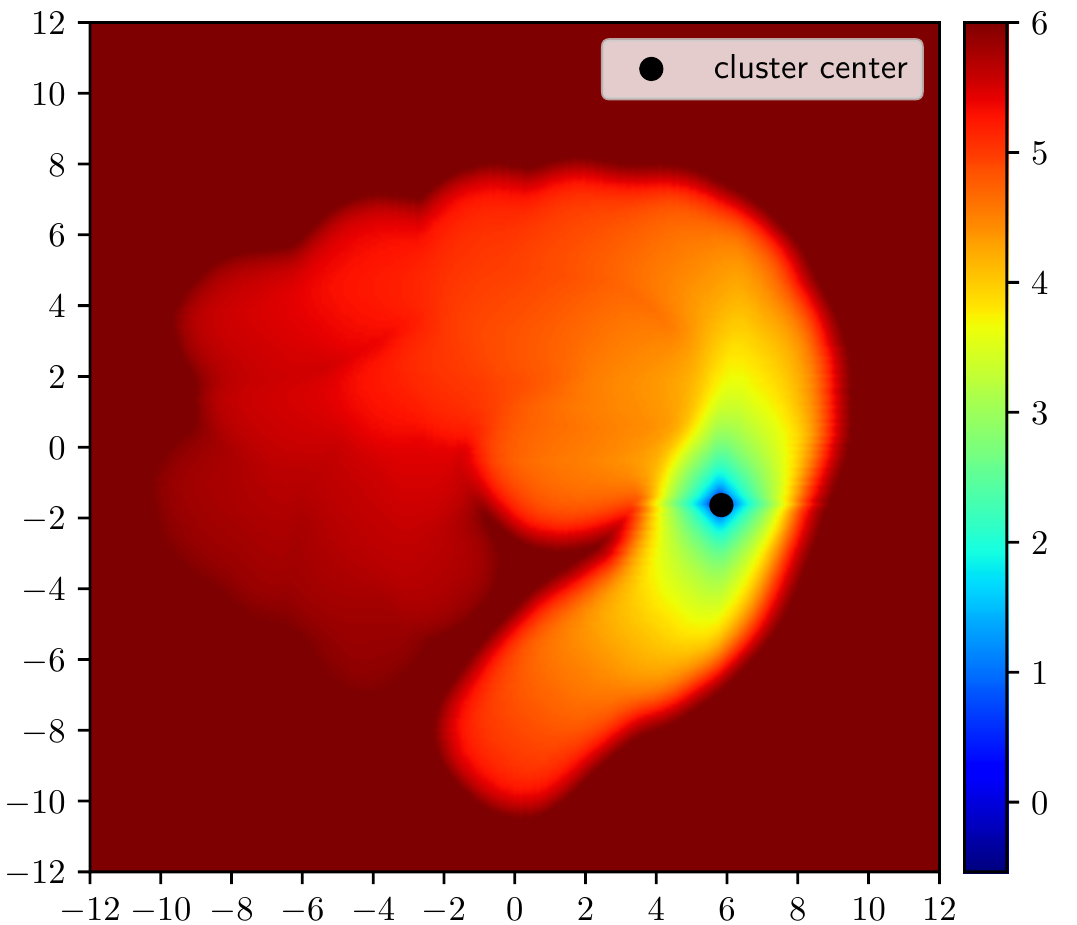}
    \end{minipage}
    \begin{minipage}[c]{0.32\linewidth}
        \centering
         \includegraphics[scale=0.38]{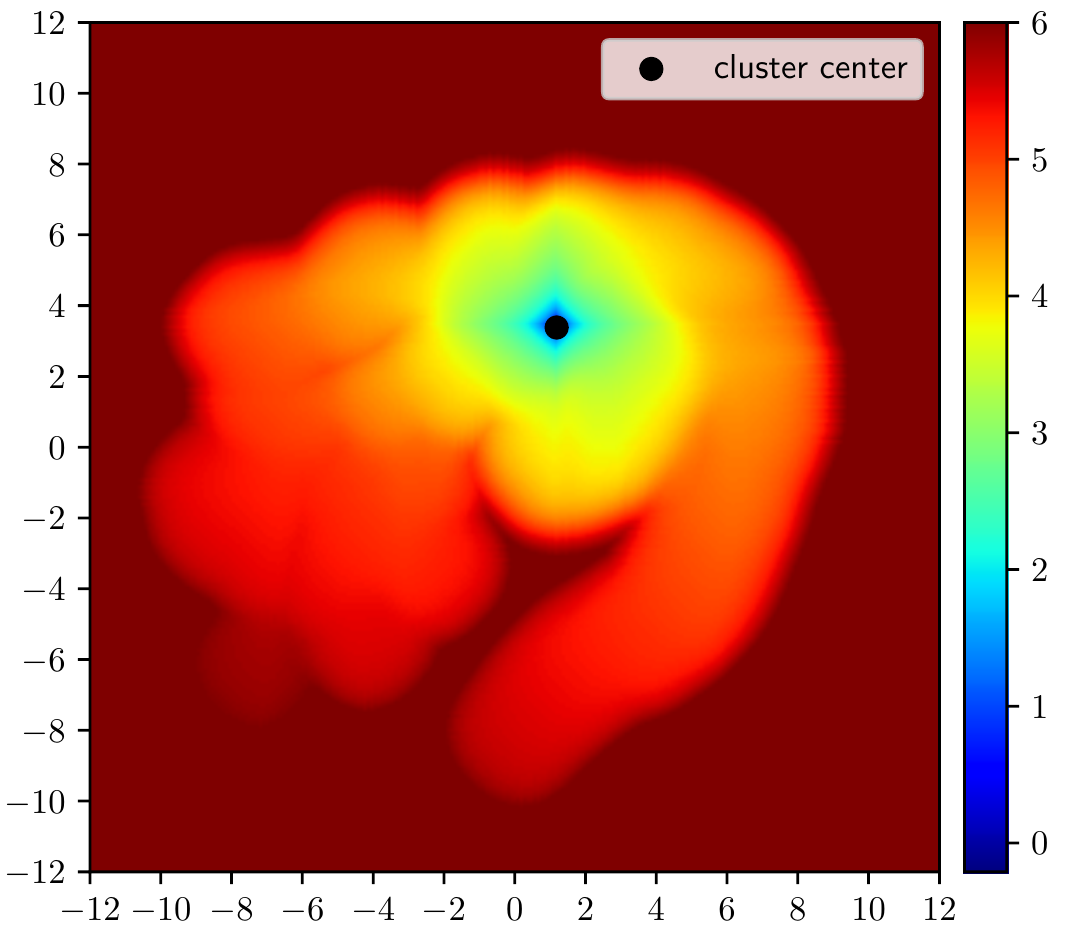}
    \end{minipage}
    \caption{Latent space of a RHVAE trained  with  a synthetic data set composed by 100 rings and  100 circles of different diameters and thicknesses along with the log of the learned volume element (left). The log distance maps from the clusters centers found by a $k$-medoids algorithm using the Riemannian metric.}
    \label{fig: Clustering shapes distance maps}
\end{figure}

 We first consider 3 hand-made synthetic data sets composed by 100 circles and 100 rings to see if the clustering under the Riemannian metric would reveal more accurate. The model is trained employing an early-stopping strategy (i.e. training is stopped if test loss does not improve in 100 epochs) and we use the $k$-medoids algorithm with $k$ set to the true number of classes to compare the clustering accuracy under each distance (i.e. affine and geodesic). Geodesic distances are approximated using Dijkstra algorithm. As highlighted in Table~\ref{tab: clustering scores}, using geodesic distances strongly improves the clustering ability of the model which jumps from 62.68 (affine) to 77.43 (geodesic) on average. An example of the obtained clustering under each distance is available in Figure~\ref{fig: Clustering shapes} and the distance maps to the 2 cluster centers found using the geodesic $k$-medoids method are presented in Figure~\ref{fig: Clustering shapes distance maps}. The use of the Riemannian metric allows us to really take into account the geometry of the latent space since geodesic curves seem to follow the data and so deeply enhance clustering. 

\subsubsection{Real Data Set}
\begin{figure}[ht]
    \centering
    \begin{minipage}[c]{0.32\linewidth}
        \centering
         \includegraphics[scale=0.38]{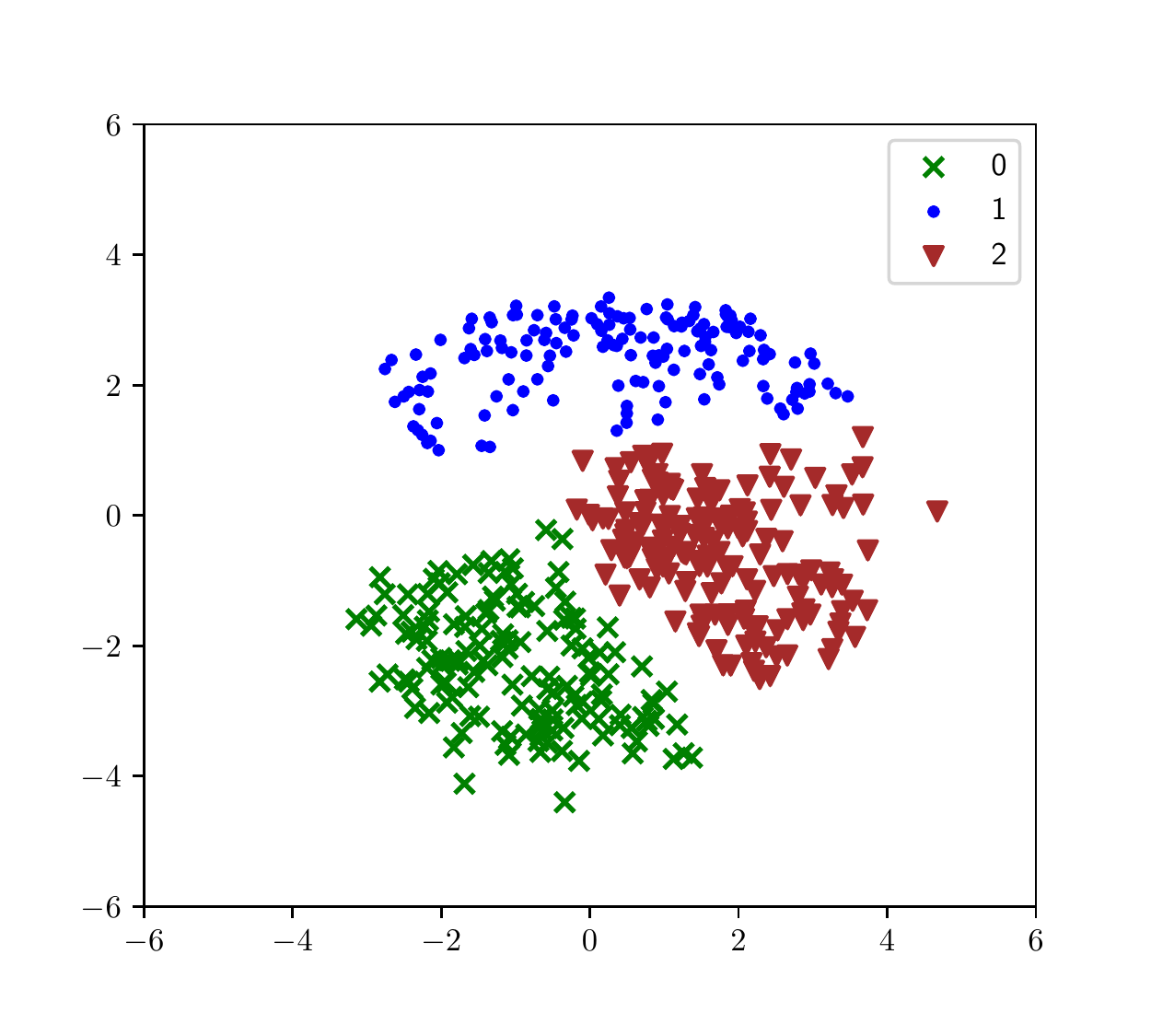}
    \end{minipage}
    \begin{minipage}[c]{0.32\linewidth}
        \centering
        \includegraphics[scale=0.38]{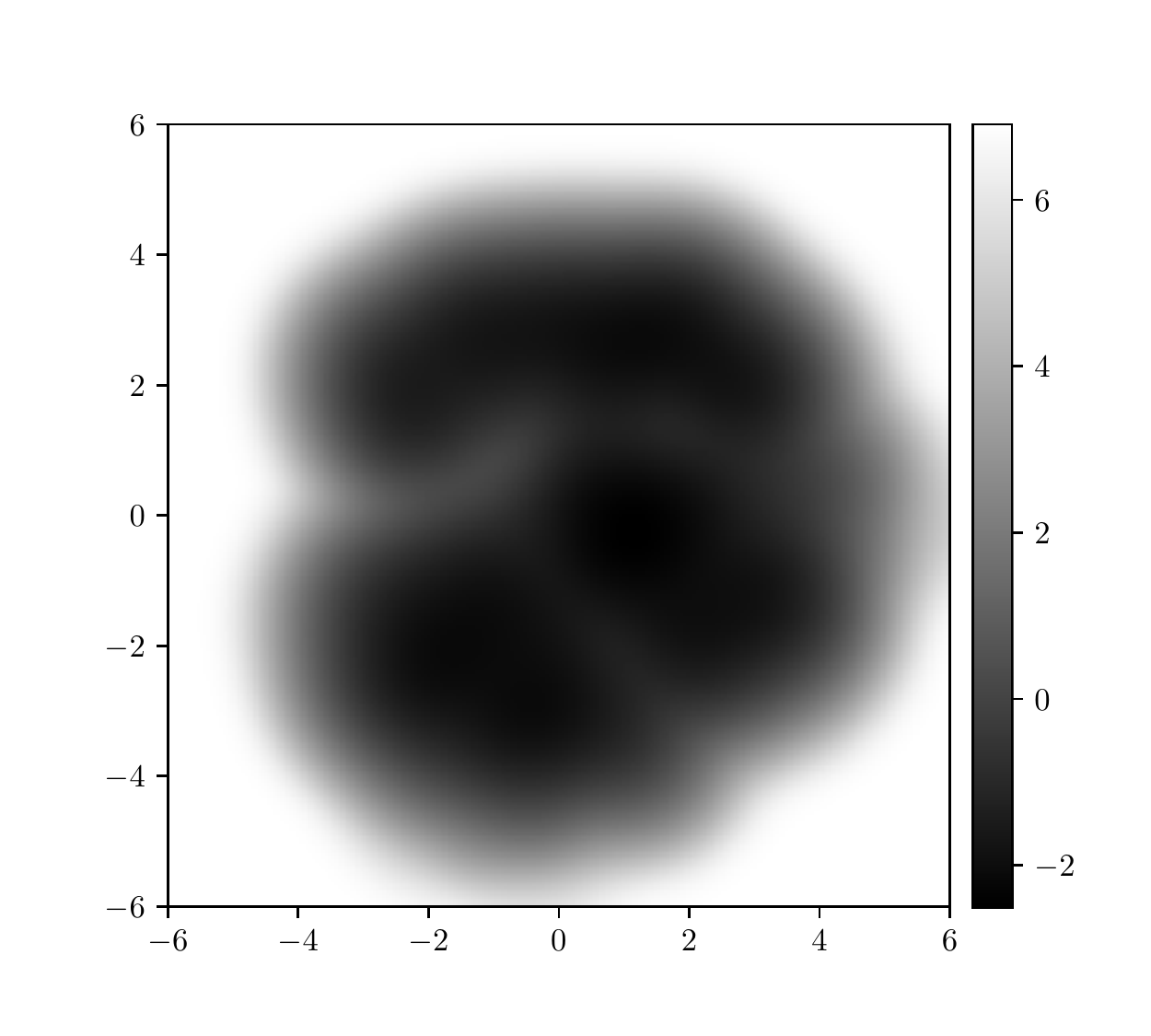}
    \end{minipage}
    \begin{minipage}[c]{0.32\linewidth}
        \centering
         \includegraphics[scale=0.38]{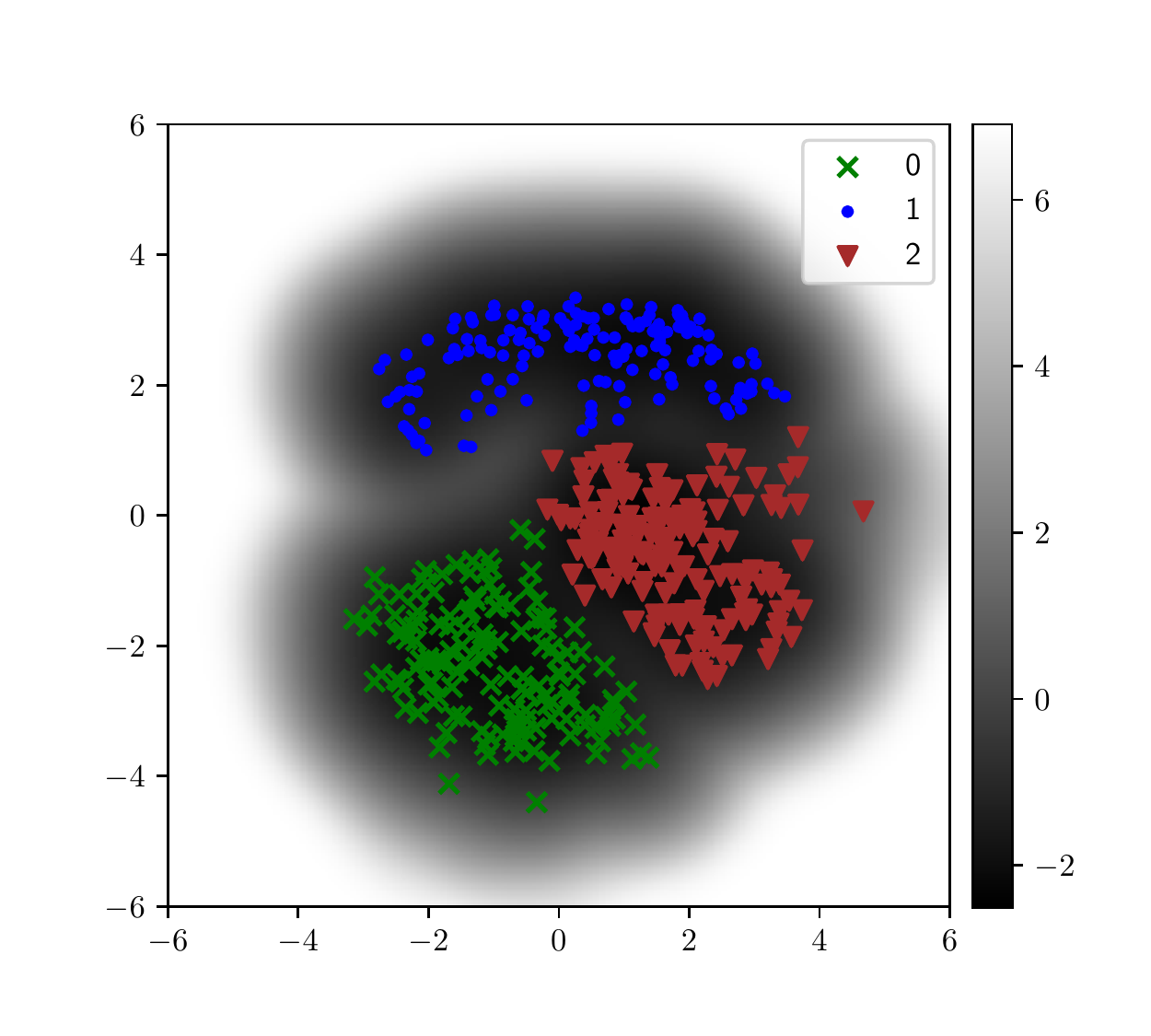}
    \end{minipage}
    \caption{Latent space of a RHVAE trained  with  3  classes  of  200  elements  each (left) along with the logarithm of the learned volume element (middle). The model is trained with $n_{\mathnormal{lf}} = 5$, $\varepsilon_{\mathnormal{lf}} = 10^{-2}$, $\lambda = 10^{-3}$ and a metric temperature $T = 0.8$.}
    \label{fig: Clustering latent space}
\end{figure}

To cross-validate its ability to cluster efficiently we also consider several data sets extracted from the MNIST and FashionMNIST data sets and report all of the results in Table~\ref{tab: clustering scores}. We create 3 groups respectively composed by the classes \{"0", "1", "2"\} (MNIST 1), \{"3", "4", "5"\} (MNIST 2) and \{"7", "8", "9"\} (MNIST 3). We select 450 samples for each class within each group and split each group into 3 subsets of 150 samples per class. We perform the same processing to create the data sets extracted from FashionMNIST. We then train the same RHVAE model on these 18 data sets and report the F$1$-scores in Table~\ref{tab: clustering scores}.  Interestingly, using geodesic distances improves the clustering ability of the model by 1 to 2\% on average with these two databases. In Figure~\ref{fig: Clustering latent space} we also provide an example of learned latent space along with the induced metric. Again, the metric seems to provide very useful information since we can clearly distinguish 3 clusters in Figure~\ref{fig: Clustering latent space} (middle) corresponding to the true clusters. The parameters used for the proposed model are available in Appendix~\ref{app: Clustering paramters}. We believe that further change on the metric shape like choosing an adaptive temperature could lead to even more powerful clustering but this will be part of future work. 
    
\begin{table}[ht]
    \centering
    \begin{tabular}{c|c|ccc|c}
    \toprule
        data set &  Model    &  Subset 1 & Subset 2 & Subset 3 & Mean\\
        \hline \hline
         \multirow{2}{*}{Synthetic data} & linear &  53.88 & 62.52 &    71.63    & 62.68\\
         & geodesic &  $\mathbf{71.41}$       & $\mathbf{81.39}$    &     $\mathbf{79.49}$    & $\mathbf{77.43}$\\
         \hline\hline
         \multirow{2}{*}{MNIST 1} & linear &  89.73 & 93.11 &    91.80    & 91.55\\
         & geodesic &  $\mathbf{91.68}$       & $\mathbf{94.51}$    &     $\mathbf{95.63}$    & $\mathbf{93.94}$\\
        \hline
        \multirow{2}{*}{MNIST 2} & linear &   68.24   &  69.22   &  79.05 & 71.17    \\
        & geodesic & $\mathbf{70.35}$     & $\mathbf{71.34}$    &  $\mathbf{79.64}$   &     $\mathbf{73.78}$\\
        \hline
        \multirow{2}{*}{MNIST 3} & linear   &  75.55    & 75.76     & 81.70    &  77.67\\
        & geodesic &  $\mathbf{76.08}$    & $\mathbf{77.94}$     &   $\mathbf{81.96}$  & $\mathbf{78.66}$\\
        \hline \hline
        \multirow{2}{*}{FashionMNIST 1} & linear   &  90.47    & 91.63     & 86.78    &  89.63\\
        & geodesic &  $\mathbf{91.44}$    & $\mathbf{92.55}$     &   $\mathbf{87.46}$  & $\mathbf{90.48}$\\
        \hline
        \multirow{2}{*}{FashionMNIST 2} & linear   &  92.20    & 91.26     & 93.30    & 92.25  \\
        & geodesic &  $\mathbf{93.56}$    & $\mathbf{91.80}$     &   $\mathbf{94.12}$  & $\mathbf{93.16}$\\
        \hline
        \multirow{2}{*}{FashionMNIST 3} & linear   &  72.46    & 79.58     & 83.16    &  78.40\\
        & geodesic &  $\mathbf{74.89}$    & $\mathbf{81.88}$     &   $\mathbf{84.83}$  & $\mathbf{80.53}$\\
    \bottomrule
    \end{tabular}
    \caption{F1-Scores. Clustering accuracy of a RHVAE model using either euclidean distances or geodesic distances under the learned metric. The model is trained with $n_{\mathnormal{lf}} =10$, $\varepsilon_{\mathnormal{lf}} = 10^{-2}$, $\lambda = 10^{-3}$ and a metric temperature $T = 0.8$.}
    \label{tab: clustering scores}
\end{table}

\section{Conclusion}

In this paper, we proposed to consider that the latent space learned by a variational auto-encoder is a Riemannian manifold endowed with a Riemannian metric. Using this interesting property led us to introduce the Riemannian Hamiltonian variational auto-encoder extending the concept of normalizing flows to Riemannian manifolds. Since this latent space modelling requires a Riemannian metric to be defined, we also proposed to learn a parametrized metric directly for the data, the learning of which can easily be intergrated within the VAE learning process. This model revealed to outperform both vanilla VAE and non {\it geometry-aware} Hamiltonian VAE in terms of reconstruction error and Log-Likelihood estimate. Finally, this metric proved to provide very useful information on the underlying latent space structure allowing for far more meaningful geodesic interpolations, better data clustering along with a more diverse and realistic data generation. Future work would consist in amending the proposed metric to perhaps enhance clustering and testing this metric on other {\it real-life} data sets such as medical images.



\acks{This work was supported in part by the French government under management of Agence Nationale de la Recherche as part of the “Investissements d’avenir” program, reference ANR19-P3IA-0001 (PRAIRIE
3IA Institute).\\
Data were provided in part by OASIS: Cross-Sectional: Principal Investigators: D. Marcus, R, Buckner, J, Csernansky J. Morris; P50 AG05681, P01 AG03991, P01 AG026276, R01 AG021910, P20 MH071616, U24 RR021382}


\clearpage

\appendix
\section{}
\label{app: Training curves}

In this appendix we provide the training curves corresponding to Table~\ref{Table: Sensitivities metrics} which allow for an easier reading than the table. Each plot presents either the value of the $ELBO$ or the Log-likelihood estimates computed on the test set and using importance sampling with 200 samples and cross-validated 5 times. To improve readability, we decide to display the moving average on 10 values. We recall that an early-stopping strategy is adopted (i.e. training is stopped if the test loss does not improve for 100 epochs).

\begin{figure}[ht]
    \centering
    \begin{minipage}[c]{0.48\linewidth}
        \centering
         \includegraphics[scale=0.4]{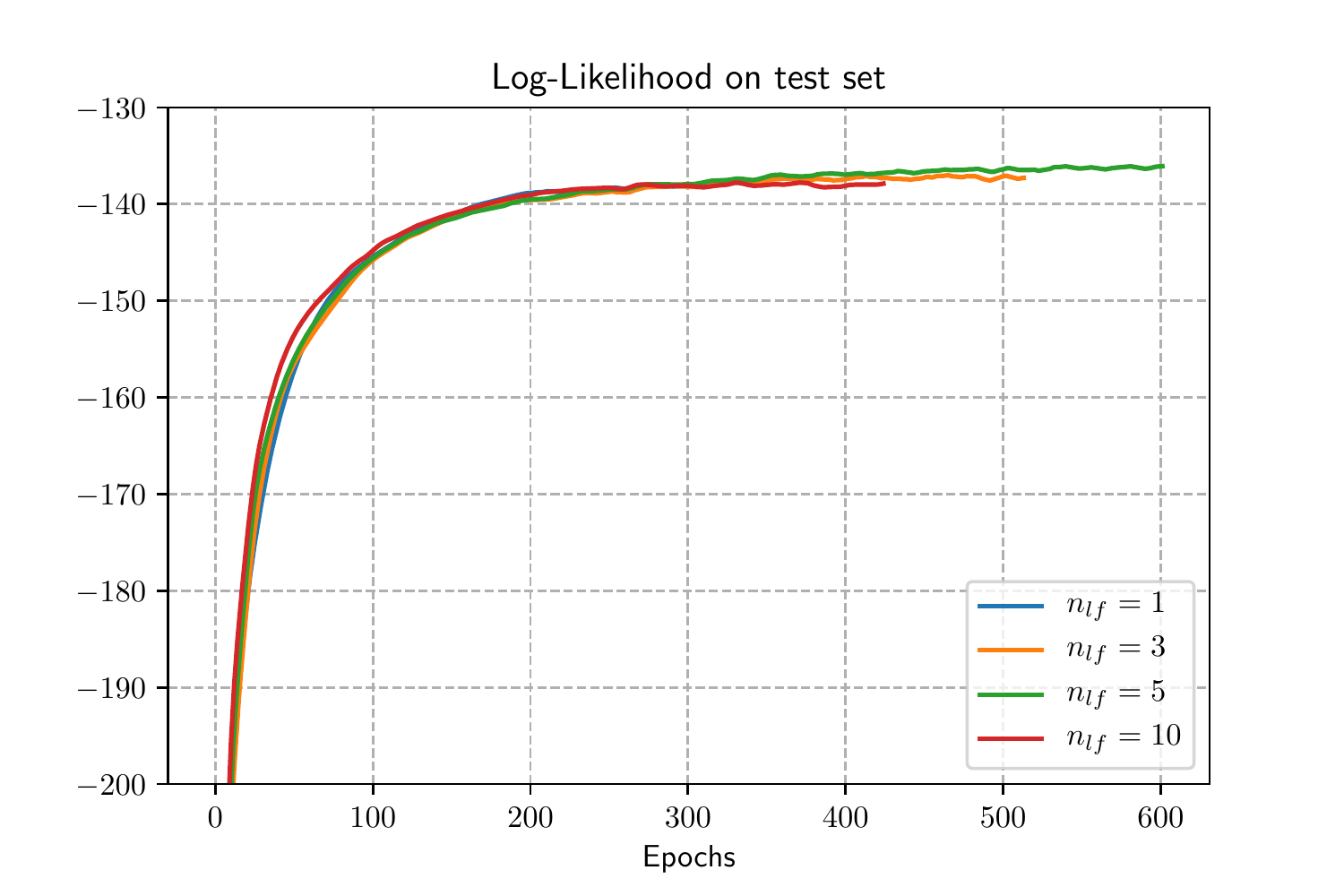}
    \end{minipage}
    \begin{minipage}[c]{0.48\linewidth}
        \centering
         \includegraphics[scale=0.4]{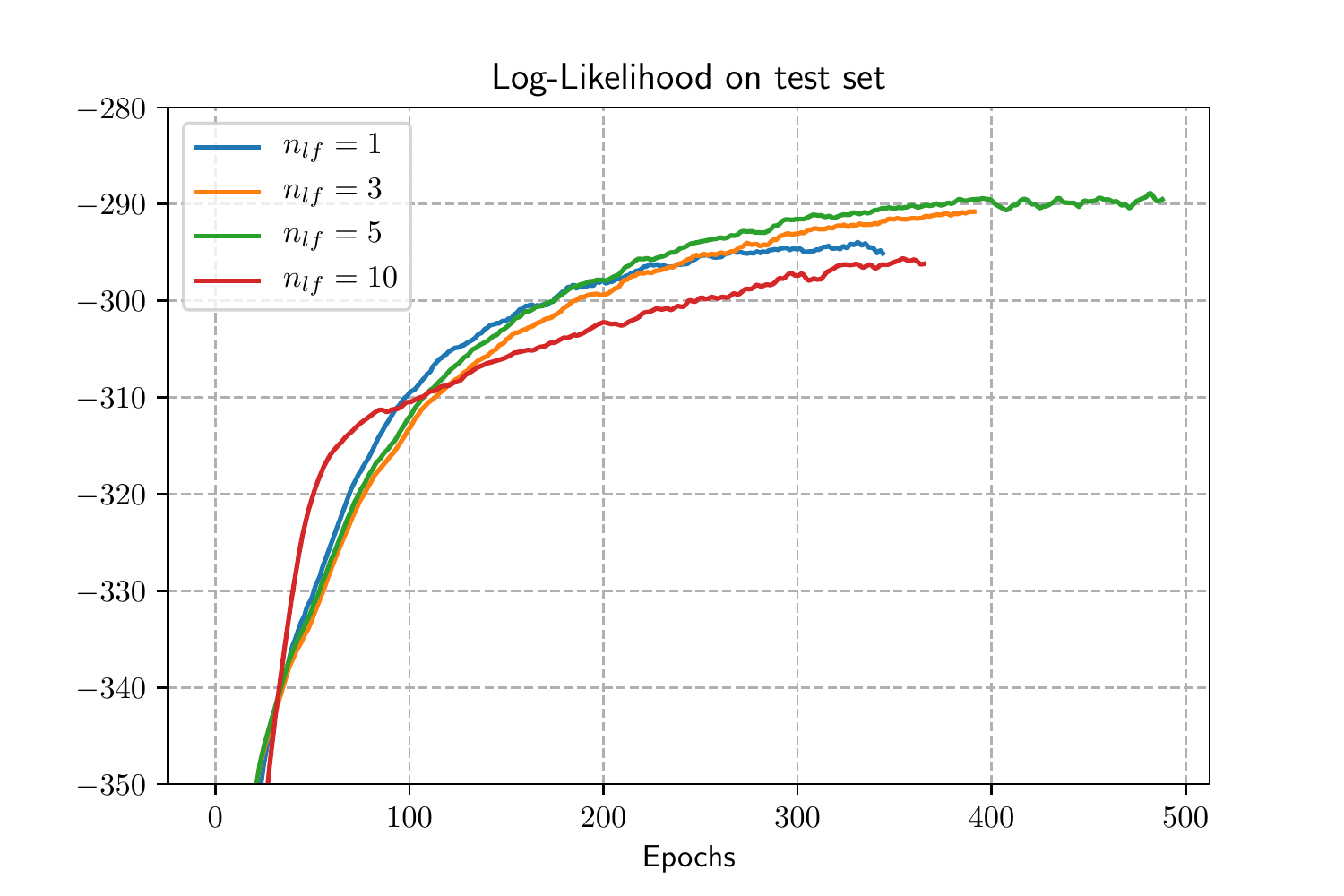}
    \end{minipage}
    \quad
    \begin{minipage}[c]{0.48\linewidth}
        \centering
         \includegraphics[scale=0.4]{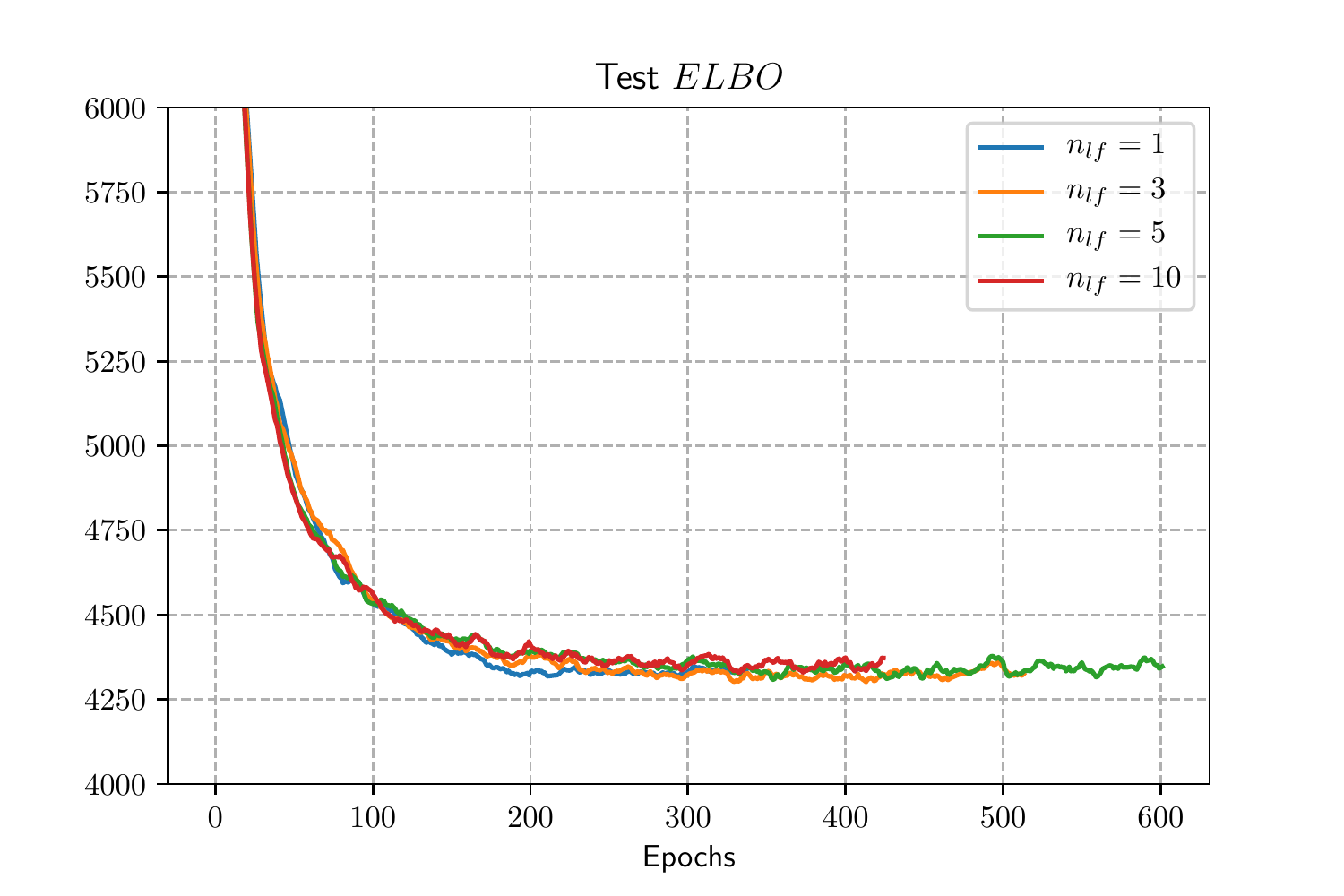}
        \subcaption*{MNIST}
        \end{minipage}
    \begin{minipage}[c]{0.48\linewidth}
        \centering
         \includegraphics[scale=0.4]{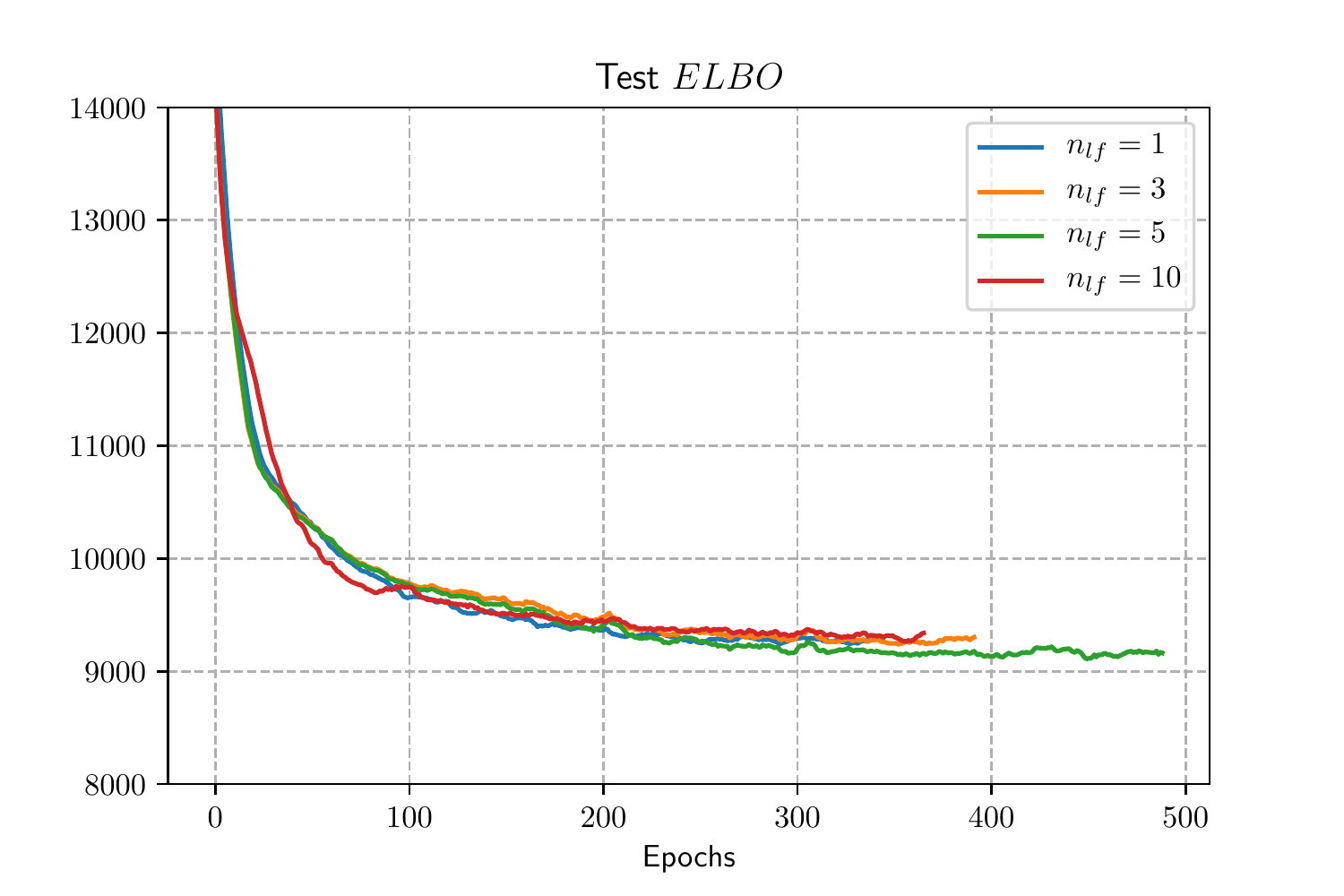}
         \subcaption*{FashionMNIST}
        \end{minipage}
    \caption{Averaged log-Likelihood values computed on the test set and test ELBO values throughout training for different values of $n_{\mathnormal{leapfrog}}$.}
    \label{fig: Sensitivities metrics leapfrog}
\end{figure}

\begin{figure}[p]
    \centering
    \begin{minipage}[c]{0.48\linewidth}
        \centering
         \includegraphics[scale=0.4]{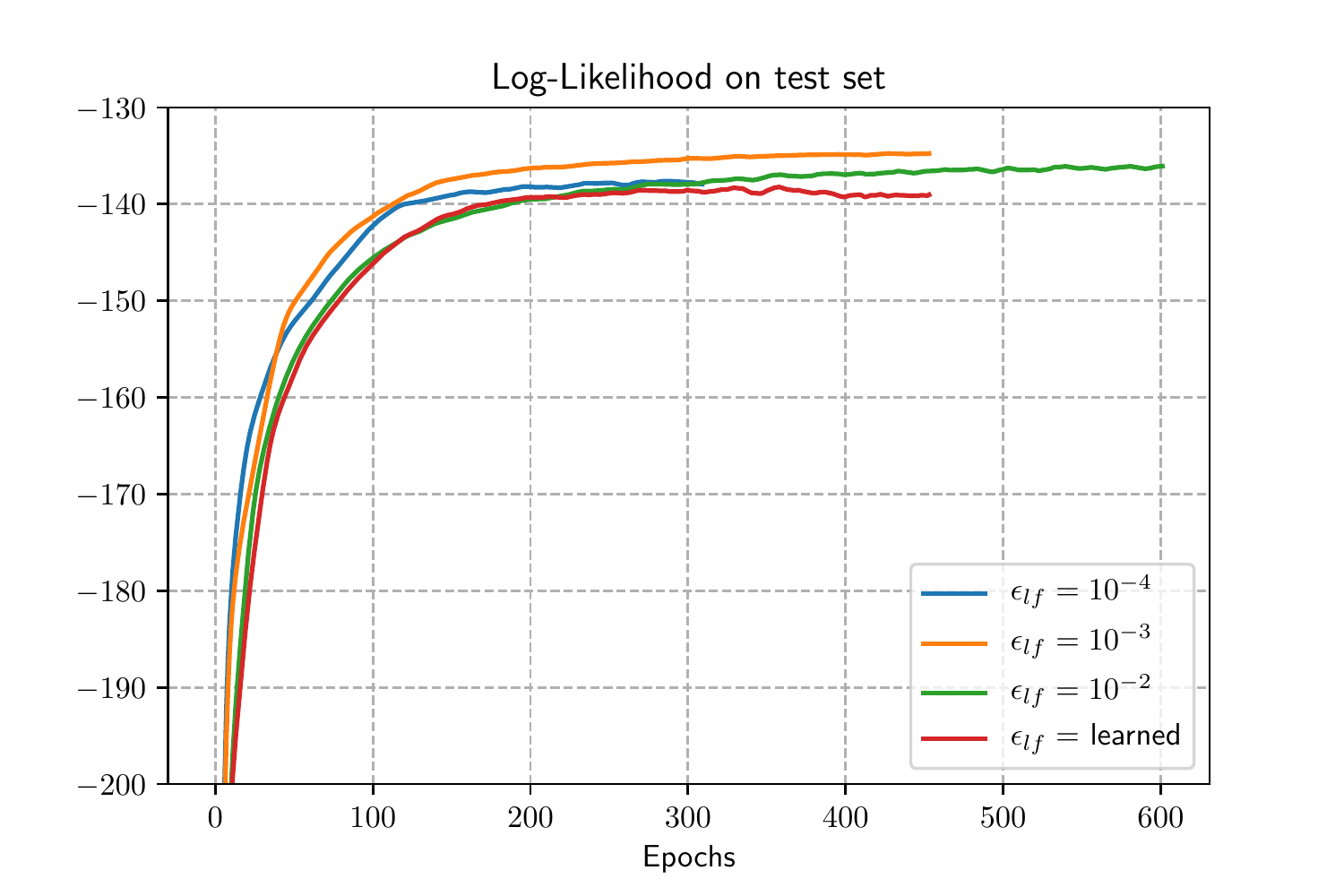}
    \end{minipage}
    \begin{minipage}[c]{0.48\linewidth}
        \centering
         \includegraphics[scale=0.4]{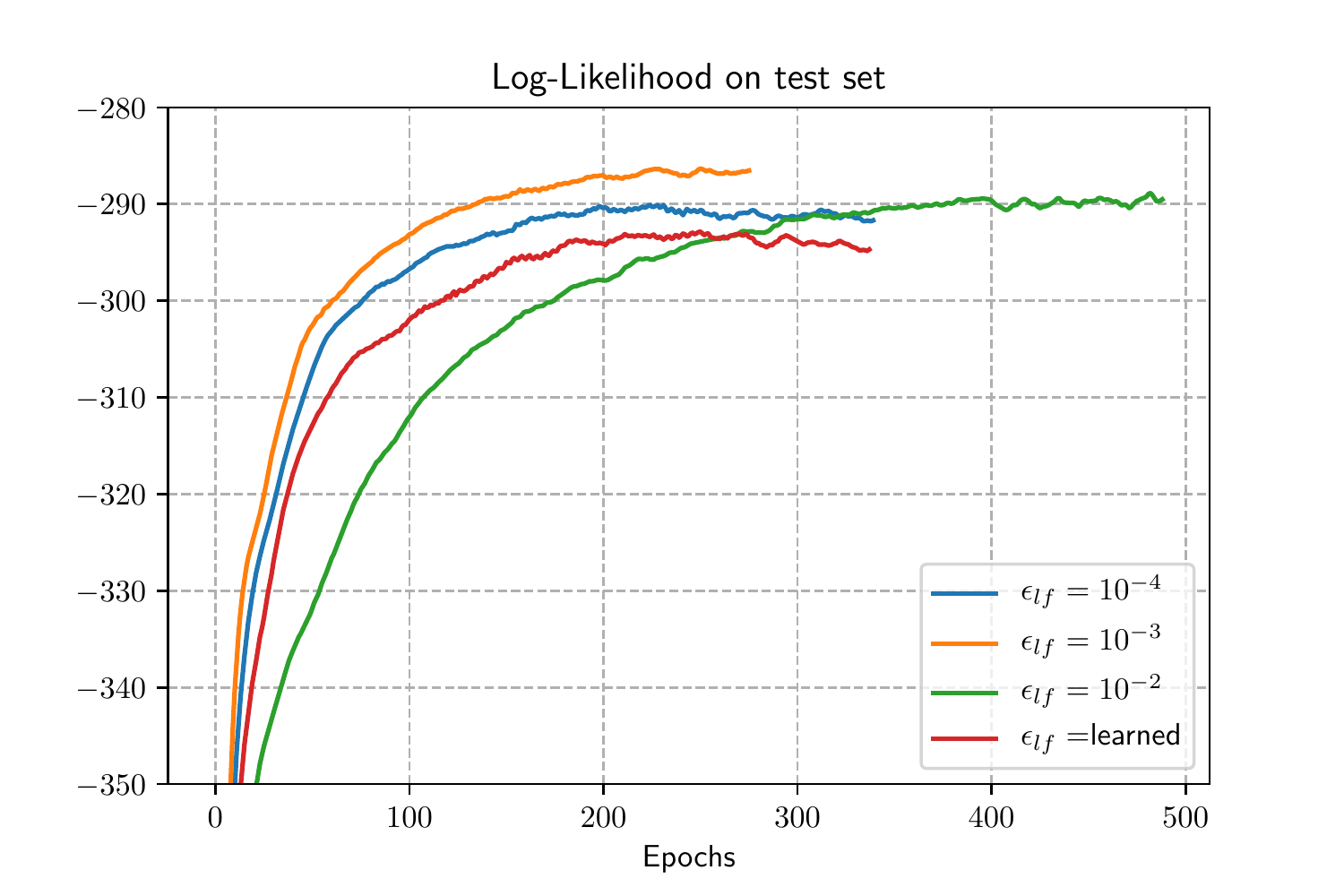}
    \end{minipage}
    \quad
    \begin{minipage}[c]{0.48\linewidth}
        \centering
         \includegraphics[scale=0.4]{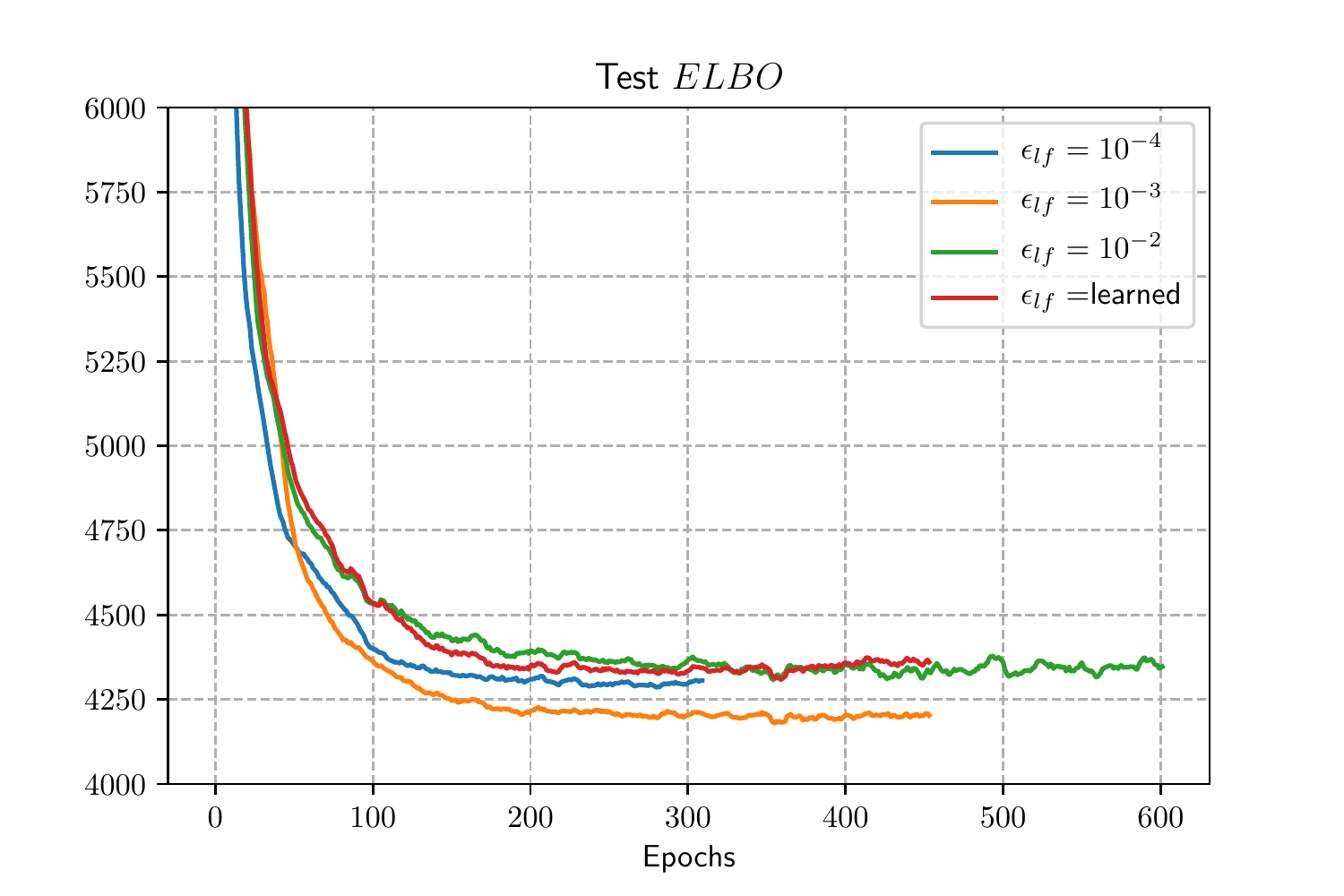}
        \subcaption*{MNIST}
        \end{minipage}
    \begin{minipage}[c]{0.48\linewidth}
        \centering
         \includegraphics[scale=0.4]{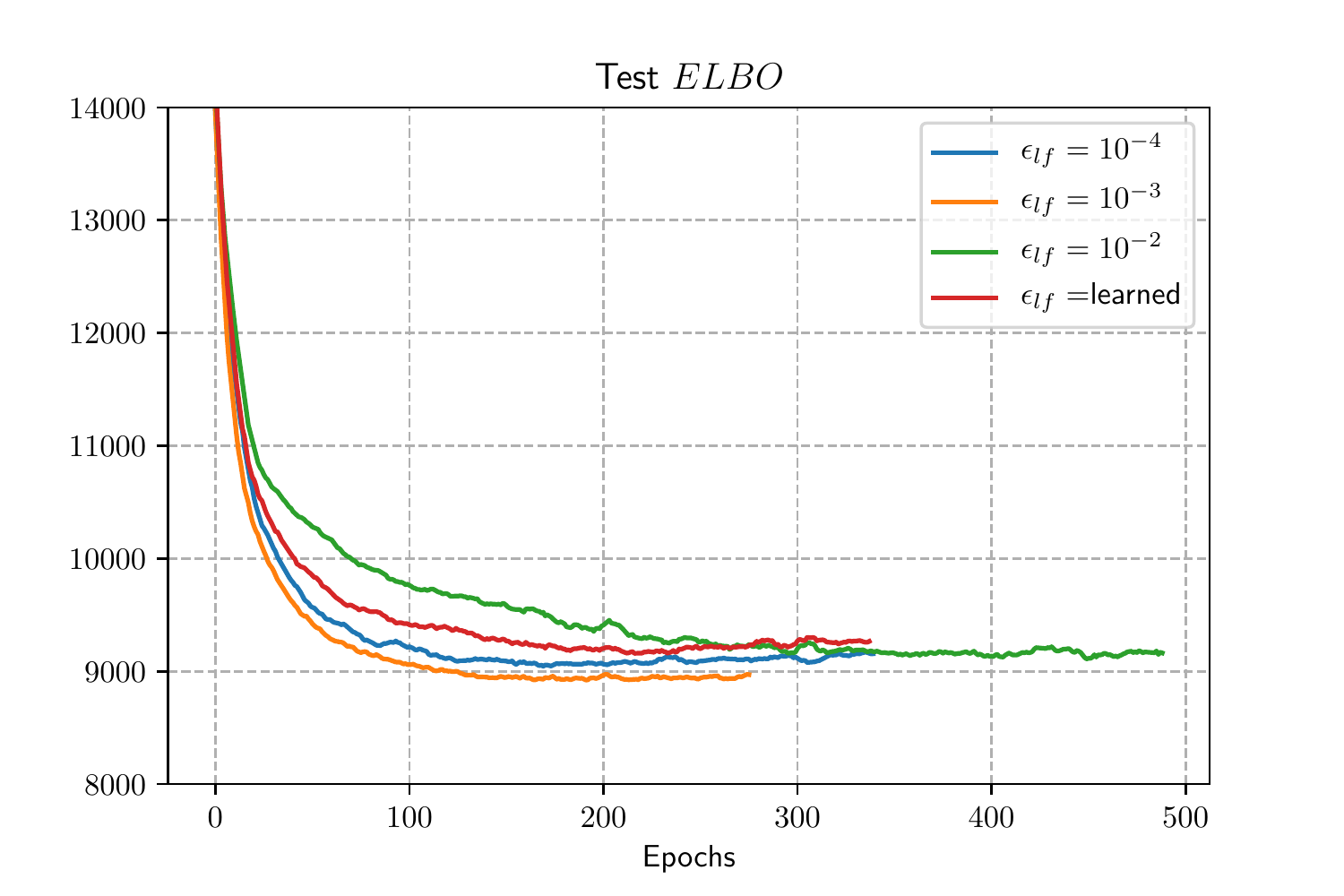}
         \subcaption*{FashionMNIST}
        \end{minipage}
    \caption{Averaged log-Likelihood values computed on the test set and test ELBO values throughout training for different leapfrog step sizes $\varepsilon_{\mathnormal{leapfrog}}$).}
    \label{fig: Sensitivities metrics leapfrog stepsize}
\end{figure}

\begin{figure}[p]
    \centering
    \begin{minipage}[c]{0.48\linewidth}
        \centering
         \includegraphics[scale=0.4]{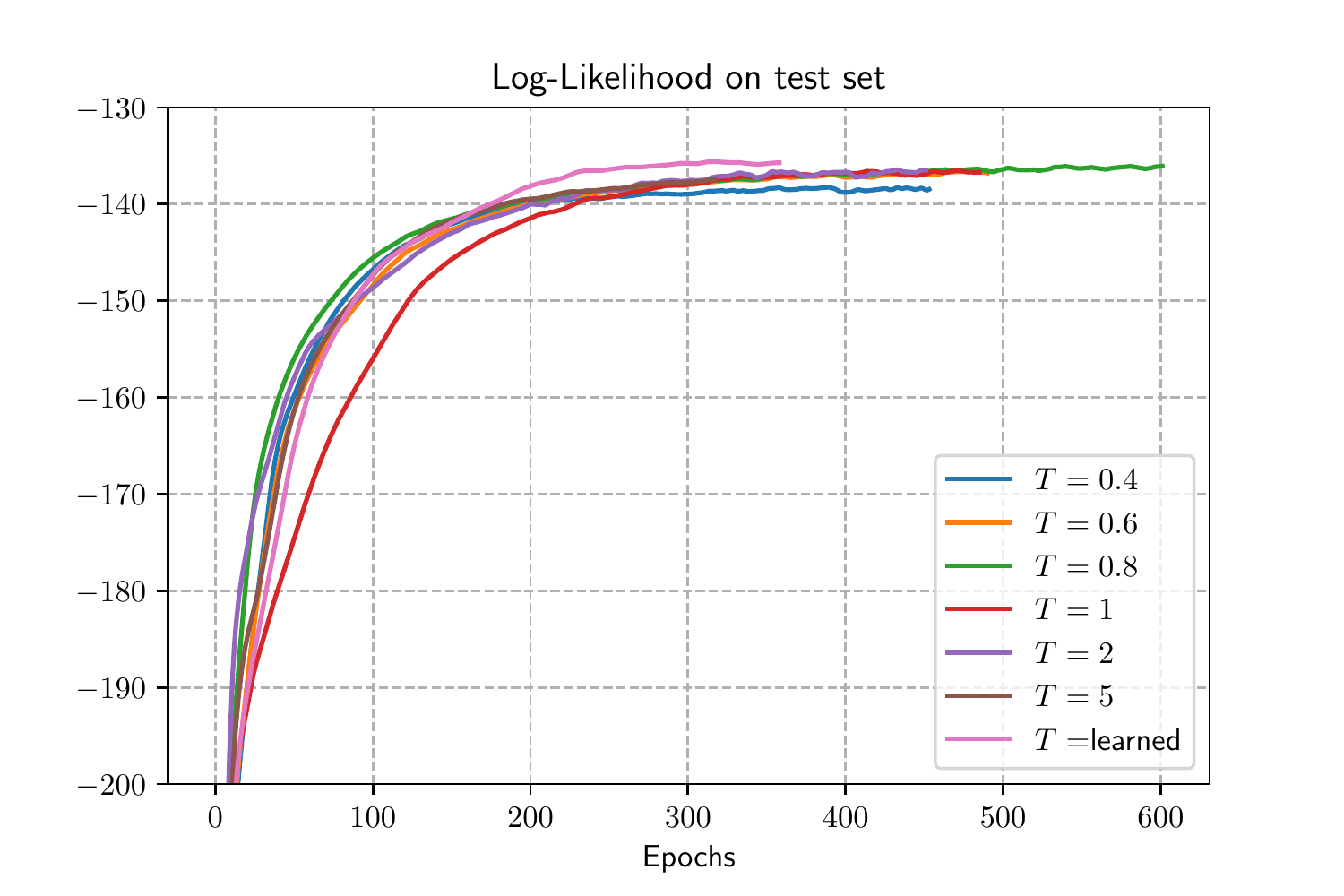}
    \end{minipage}
    \begin{minipage}[c]{0.48\linewidth}
        \centering
         \includegraphics[scale=0.4]{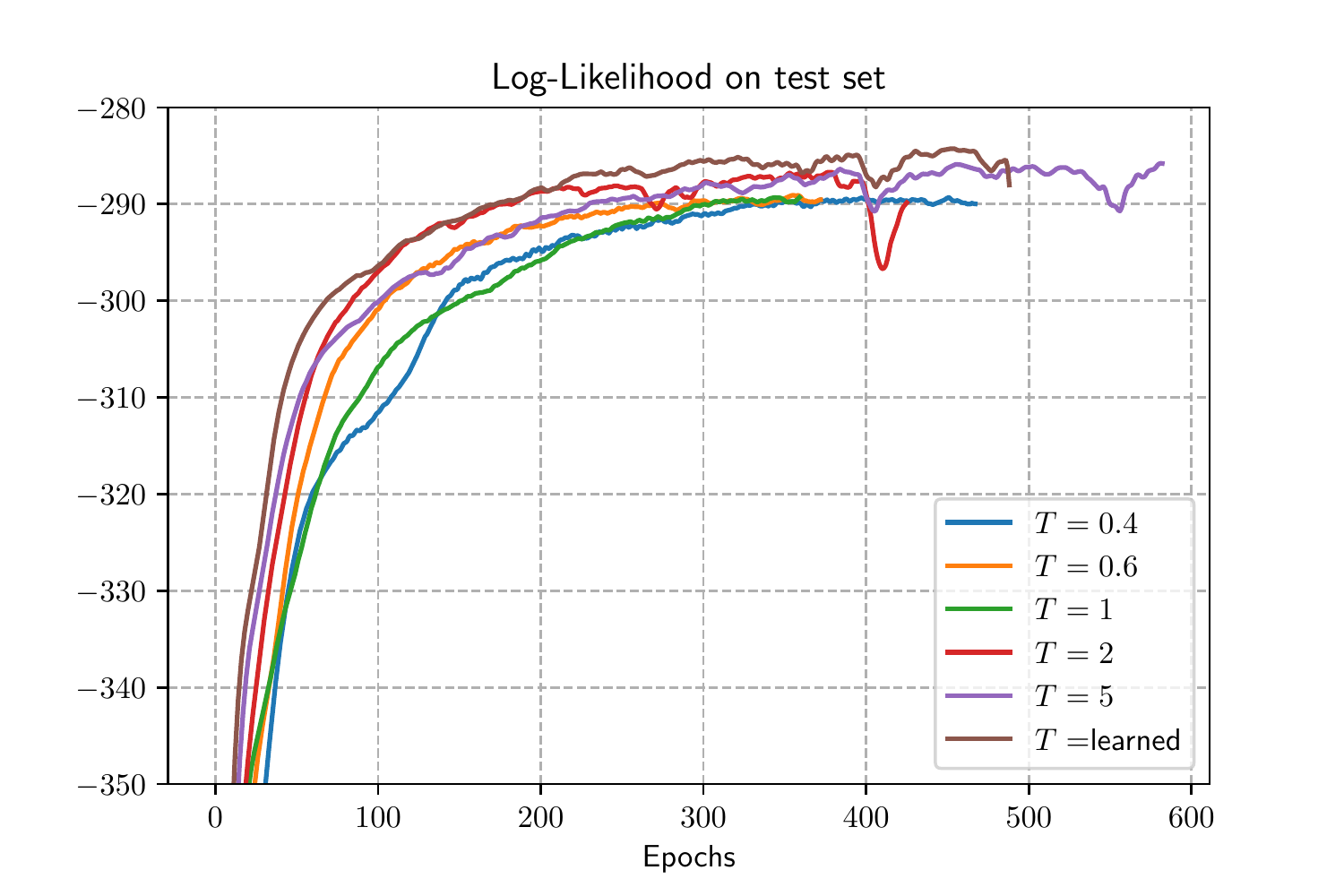}
    \end{minipage}
    \quad
    \begin{minipage}[c]{0.48\linewidth}
        \centering
         \includegraphics[scale=0.4]{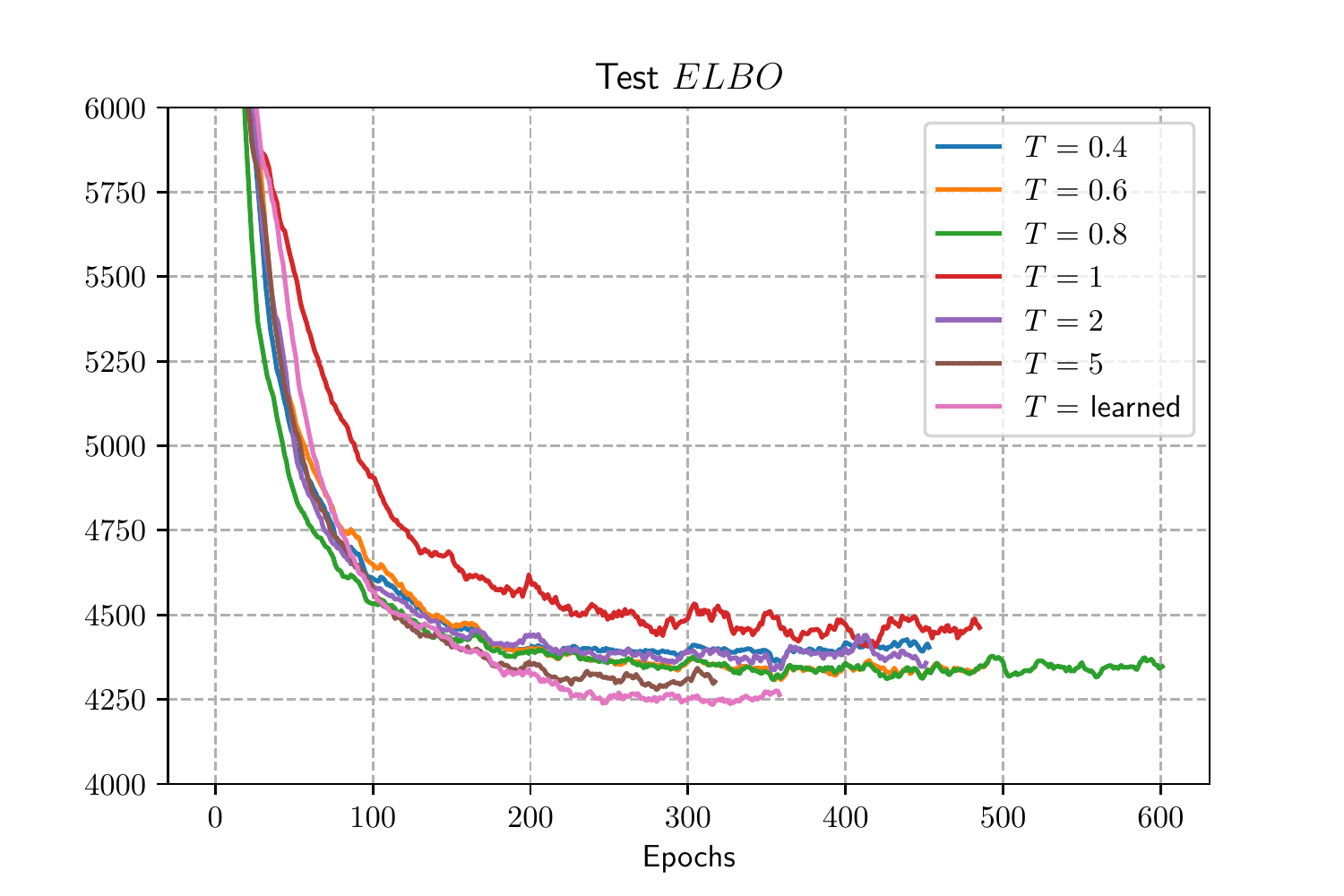}
        \subcaption*{MNIST}
        \end{minipage}
    \begin{minipage}[c]{0.48\linewidth}
        \centering
         \includegraphics[scale=0.4]{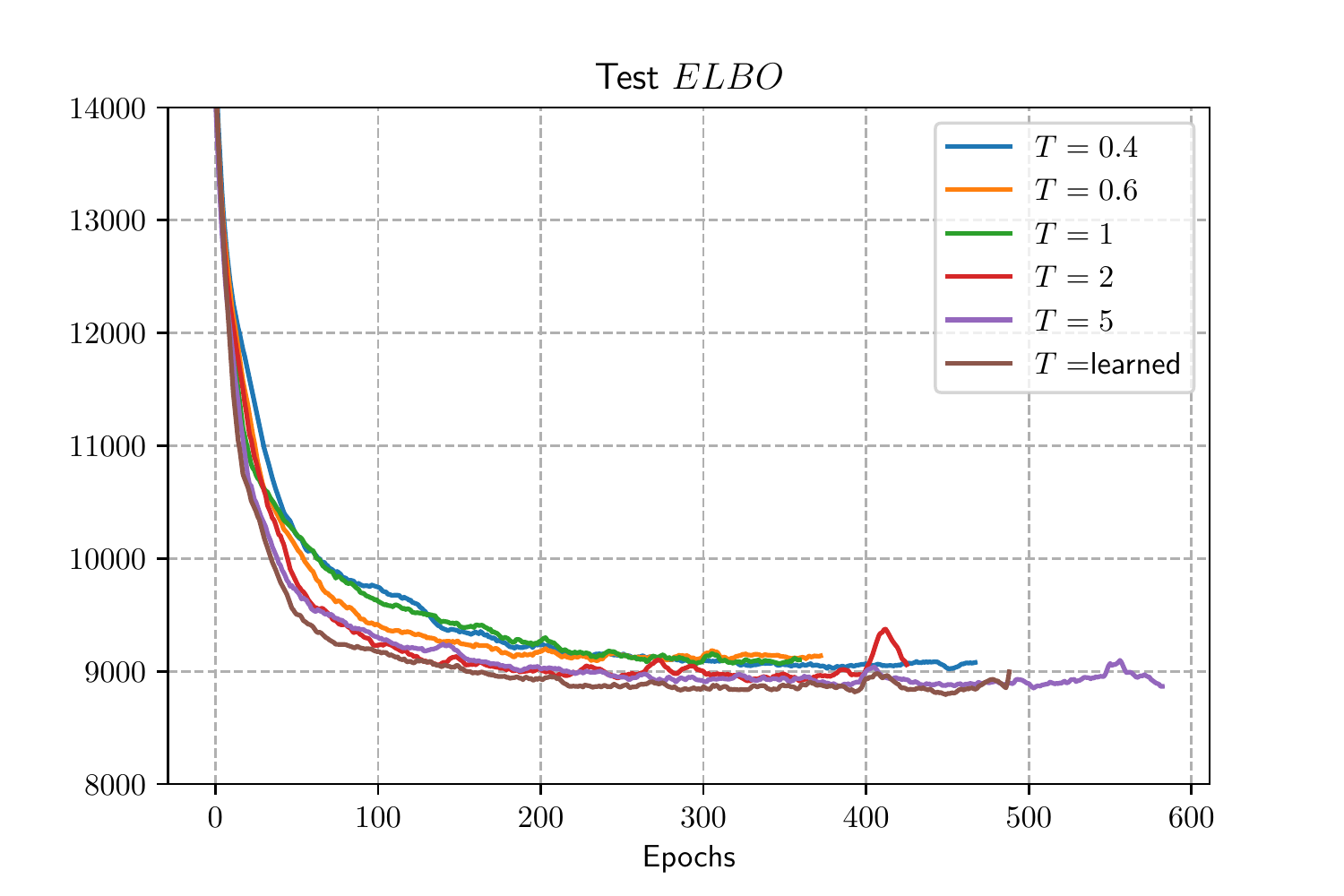}
         \subcaption*{FashionMNIST}
        \end{minipage}
    \caption{Averaged log-Likelihood values computed on the test set and test ELBO values throughout training for different metric temperatures $T$.}
    \label{fig: Sensitivities metrics temperature}
\end{figure}

\begin{figure}[p]
    \centering
    \begin{minipage}[c]{0.48\linewidth}
        \centering
         \includegraphics[scale=0.4]{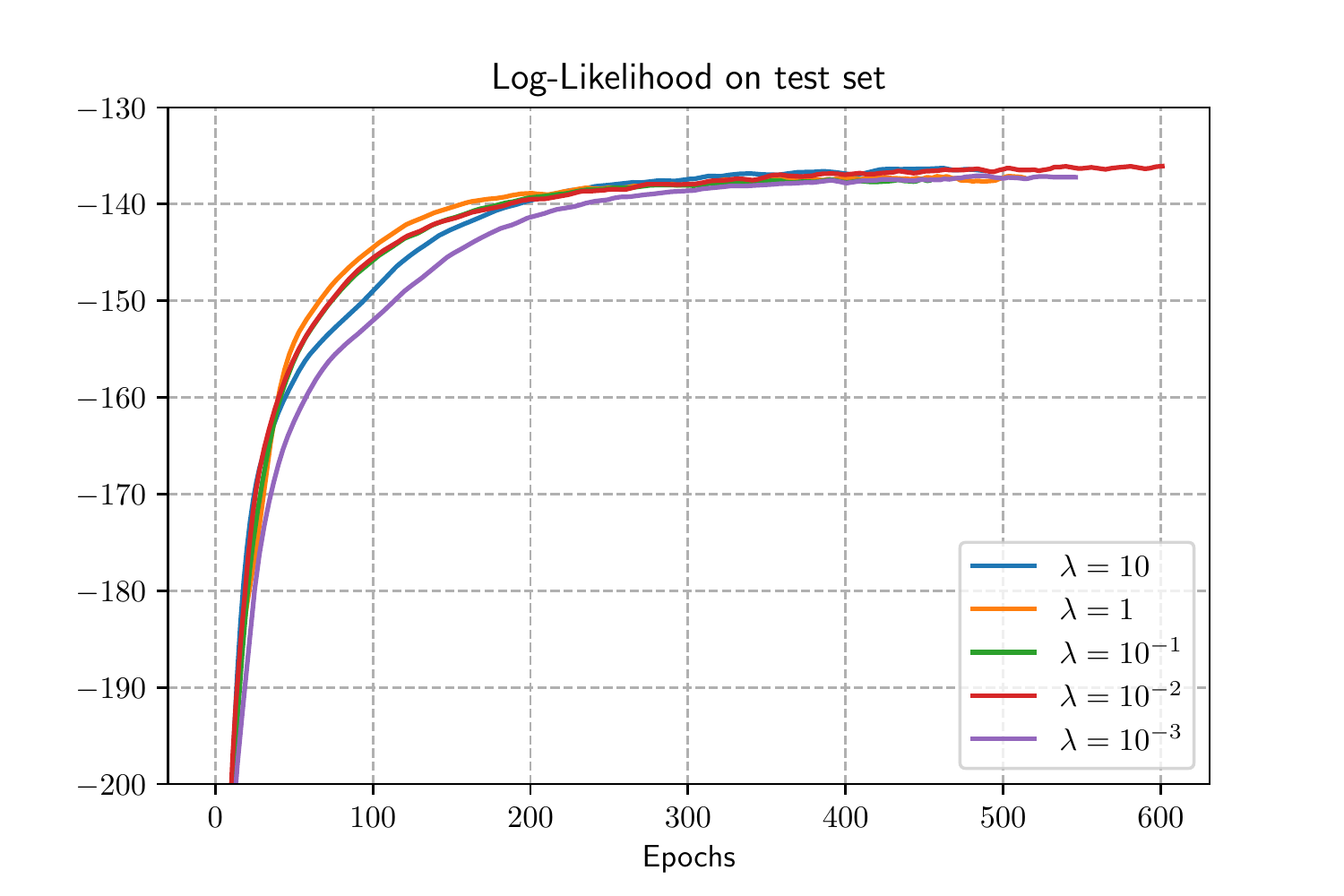}
    \end{minipage}
    \begin{minipage}[c]{0.48\linewidth}
        \centering
         \includegraphics[scale=0.4]{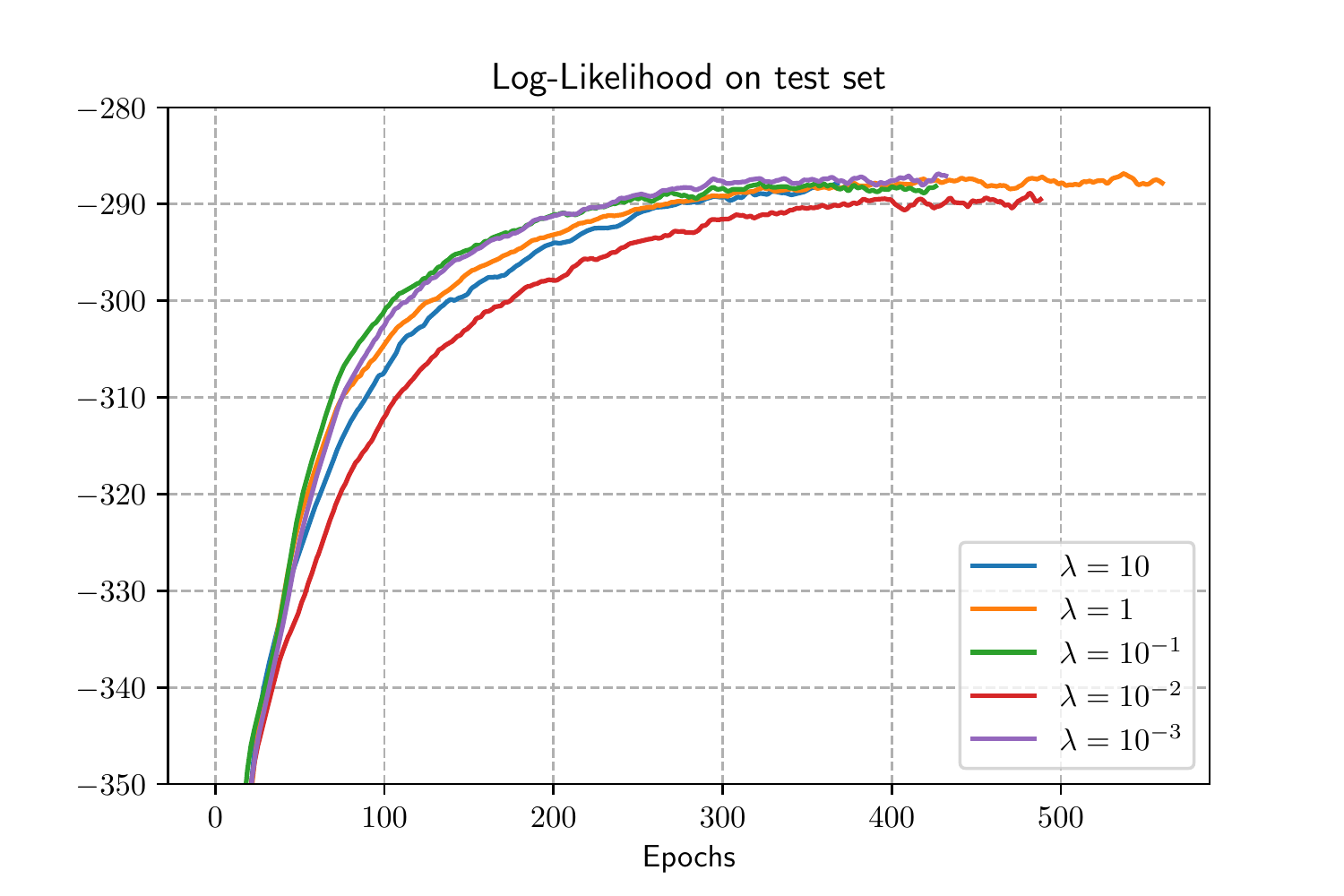}
    \end{minipage}
    \quad
    \begin{minipage}[c]{0.48\linewidth}
        \centering
         \includegraphics[scale=0.4]{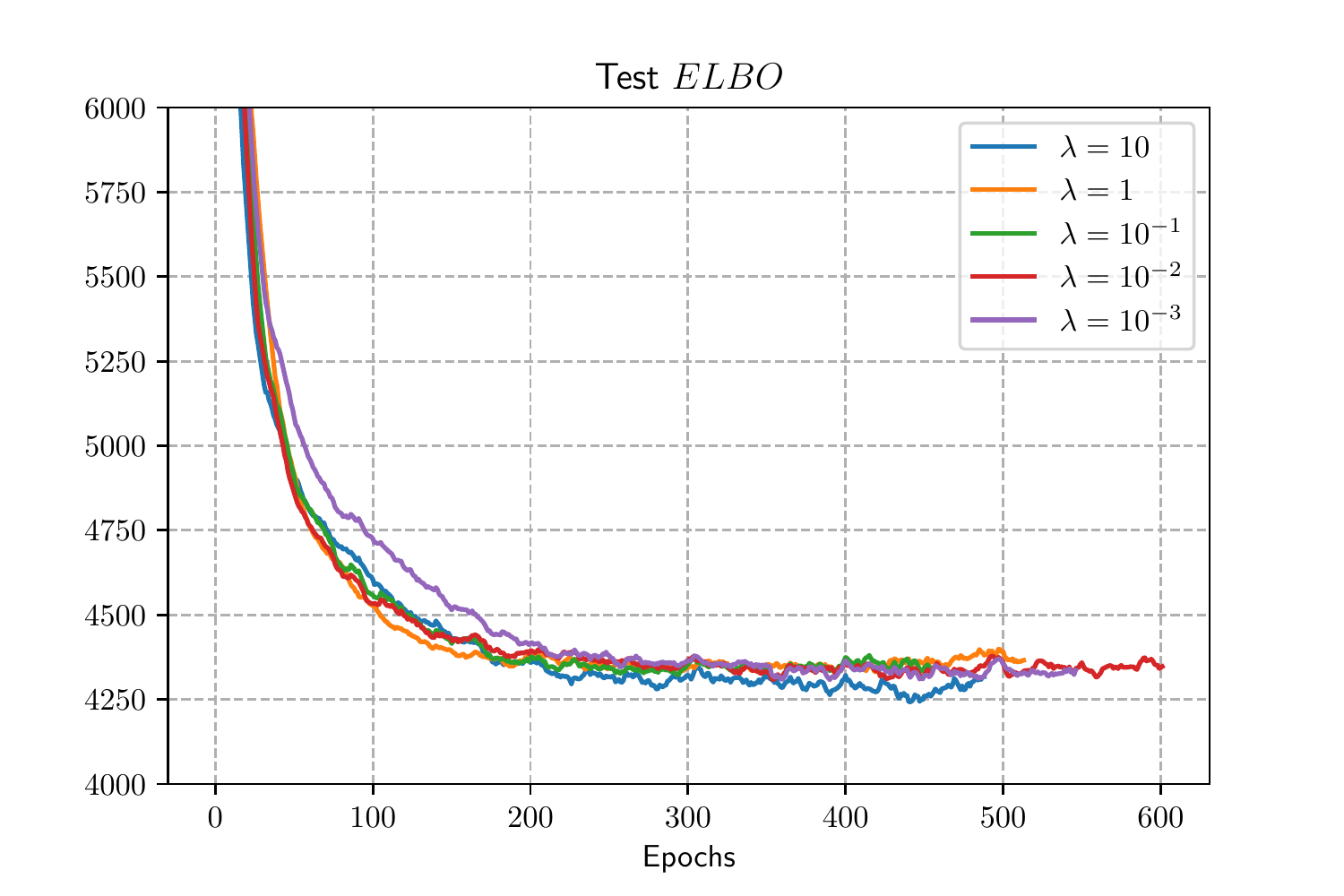}
        \subcaption*{MNIST}
        \end{minipage}
    \begin{minipage}[c]{0.48\linewidth}
        \centering
         \includegraphics[scale=0.4]{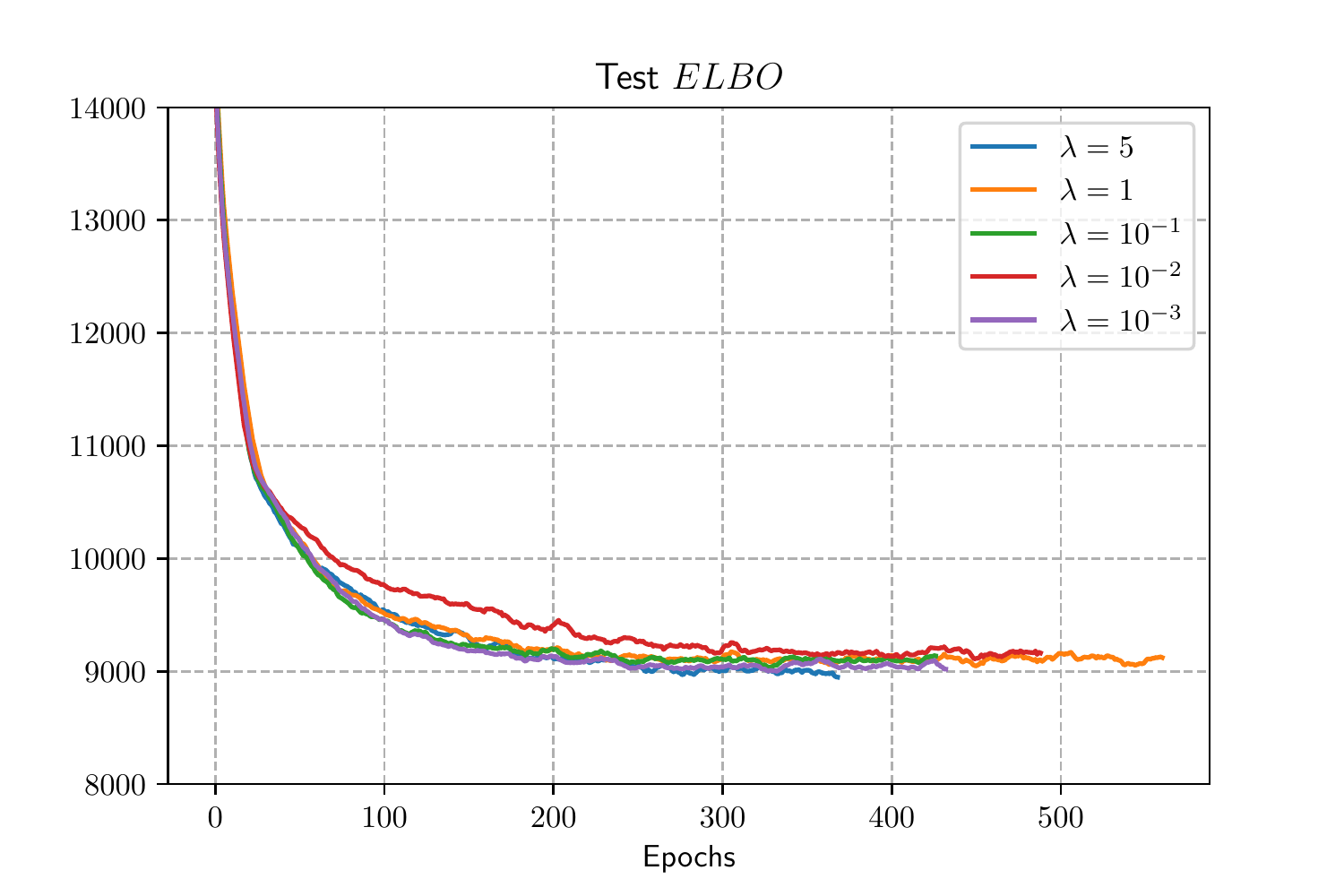}
         \subcaption*{FashionMNIST}
        \end{minipage}
    \caption{Averaged log-Likelihood values computed on the test set and test ELBO values throughout training for different metric regularization $\lambda$.}
    \label{fig: Sensitivities metrics regularization}
\end{figure}



\clearpage

\section{}
\label{app: Geodesic training samples}

\begin{figure}[ht]
    \centering
    \begin{minipage}[c]{0.32\linewidth}
        \centering
         \includegraphics[scale=0.3]{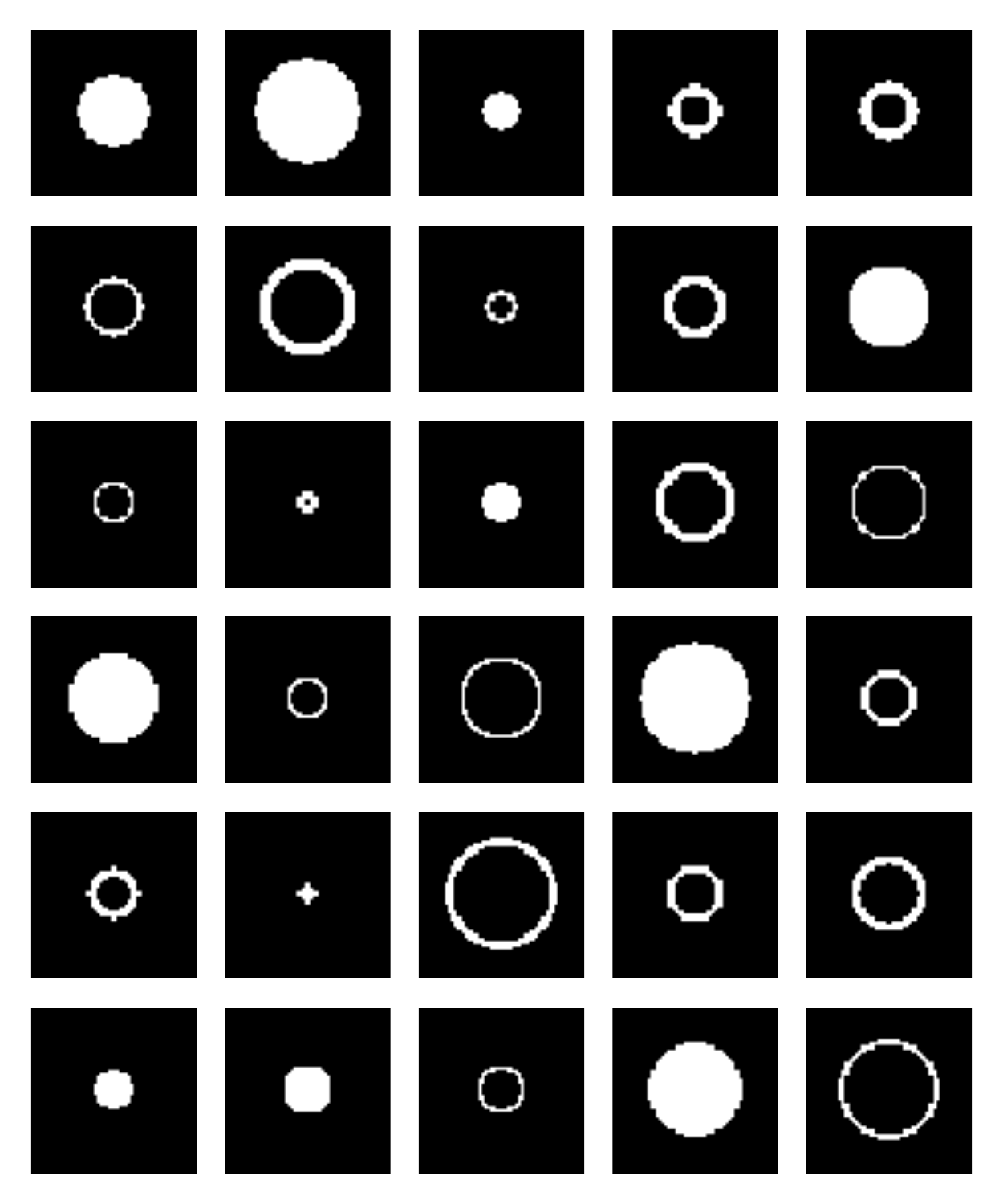}
         \vskip -0.5em
         \subcaption*{Synthetic data}
    \end{minipage}
    \begin{minipage}[c]{0.32\linewidth}
        \centering
         \includegraphics[scale=0.3]{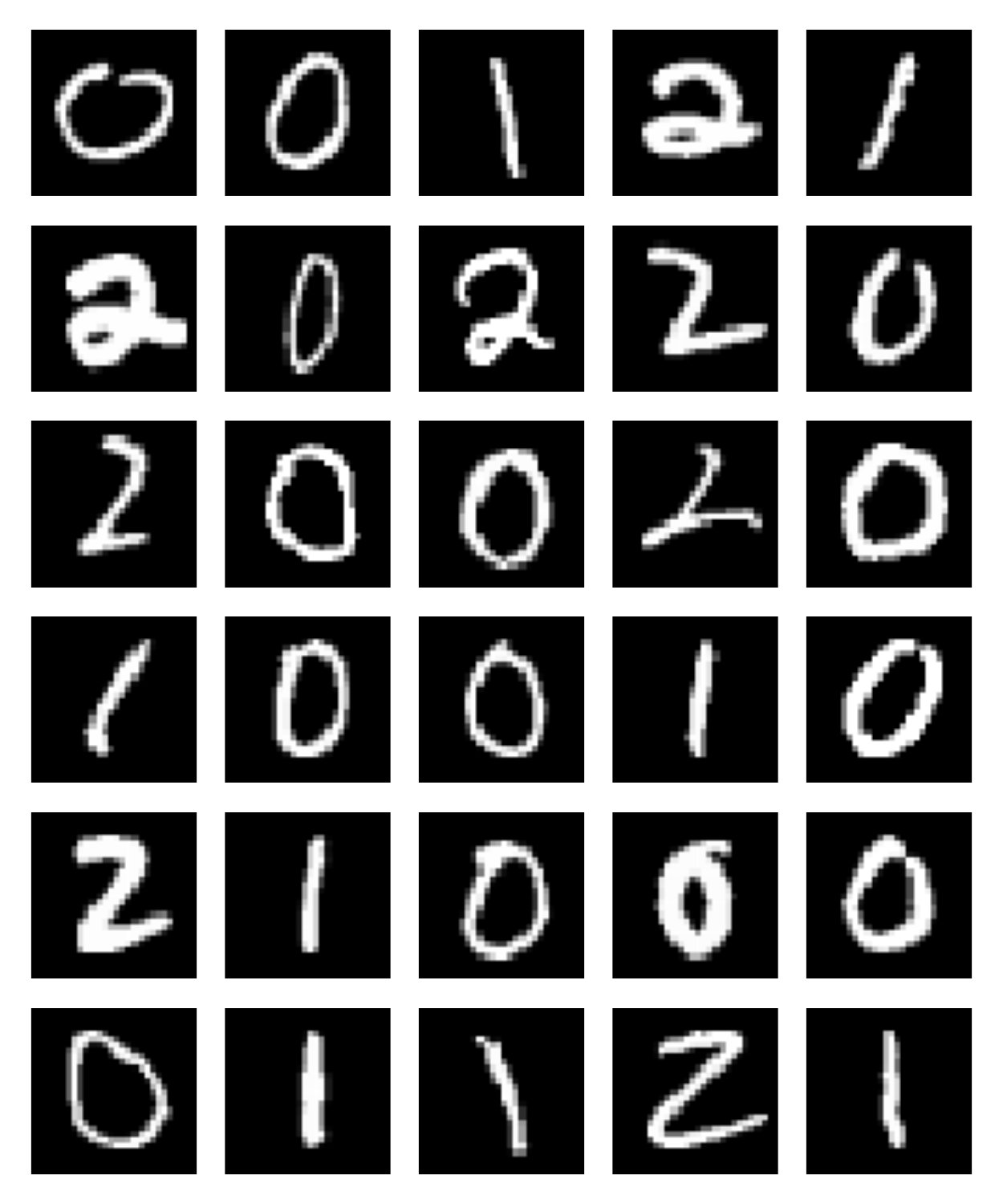}
         \vskip -0.5em
         \subcaption*{MNIST}
    \end{minipage}
    \begin{minipage}[c]{0.32\linewidth}
        \centering
         \includegraphics[scale=0.3]{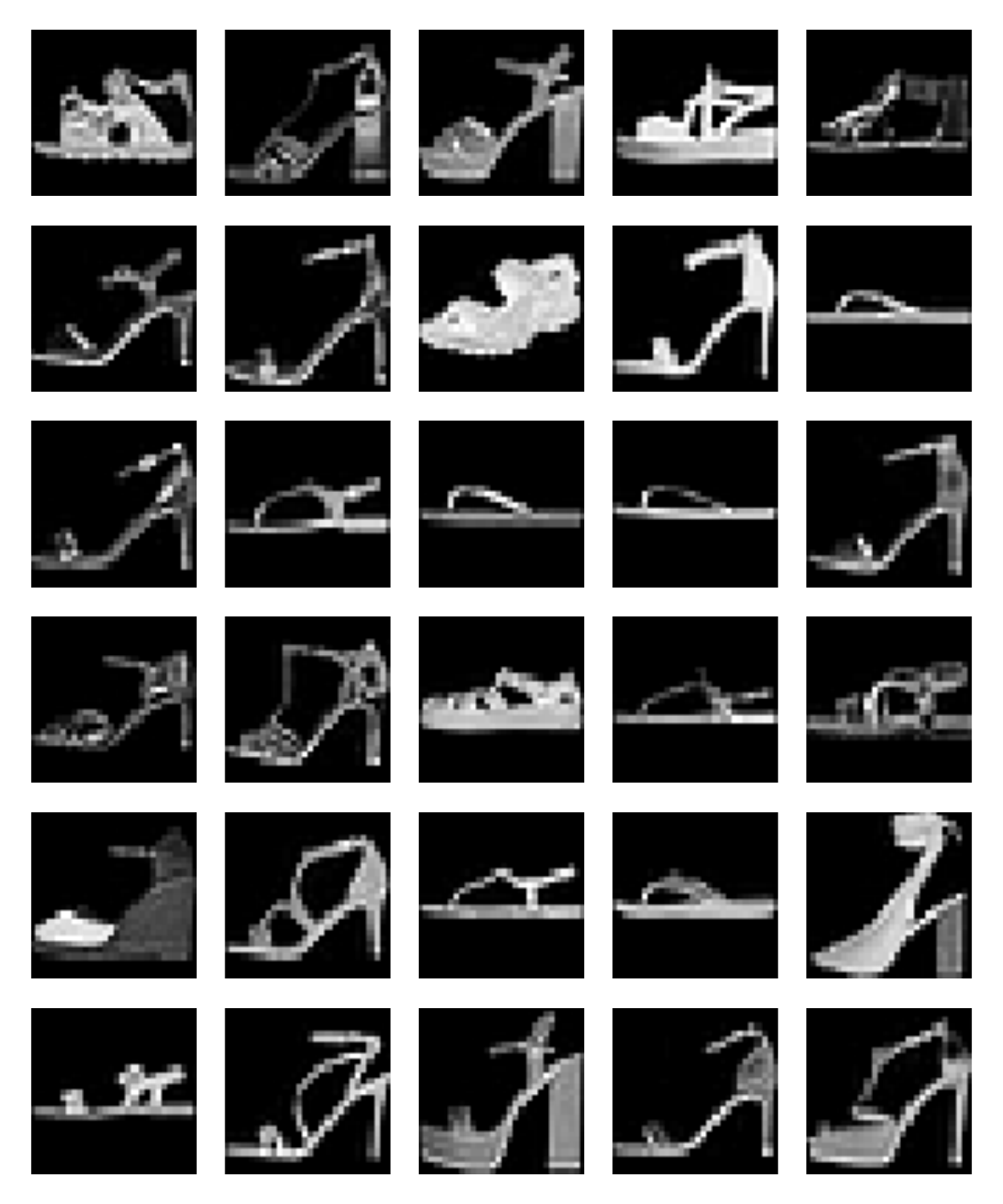}
         \vskip -0.5em
         \subcaption*{FashionMNIST}
        \end{minipage}
        \quad
    \begin{minipage}[c]{0.32\linewidth}
        \centering
         \includegraphics[scale=0.3]{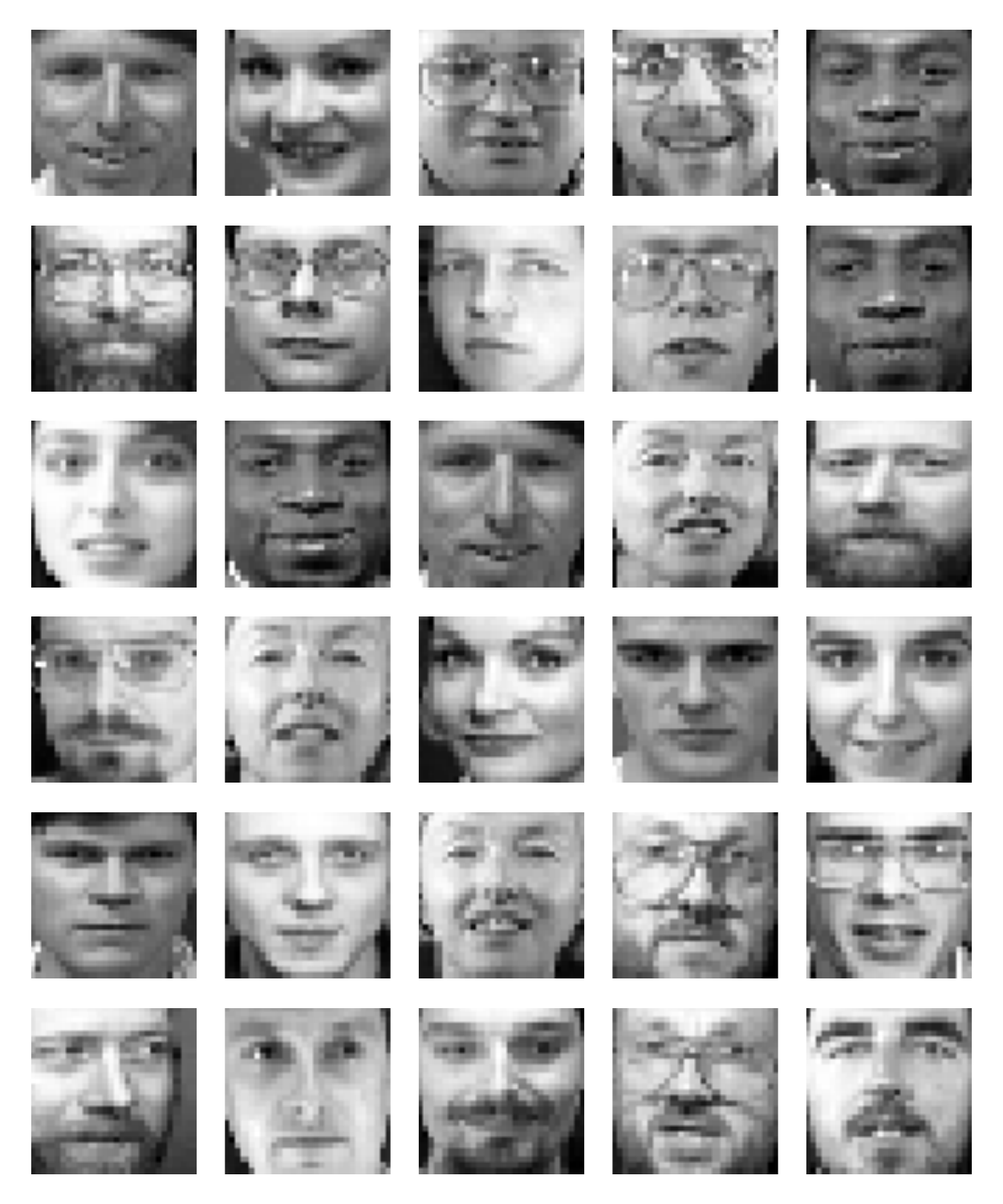}
         \vskip -0.5em
         \subcaption*{Olivetti}
        \end{minipage}
    \begin{minipage}[c]{0.32\linewidth}
        \centering
         \includegraphics[scale=0.3]{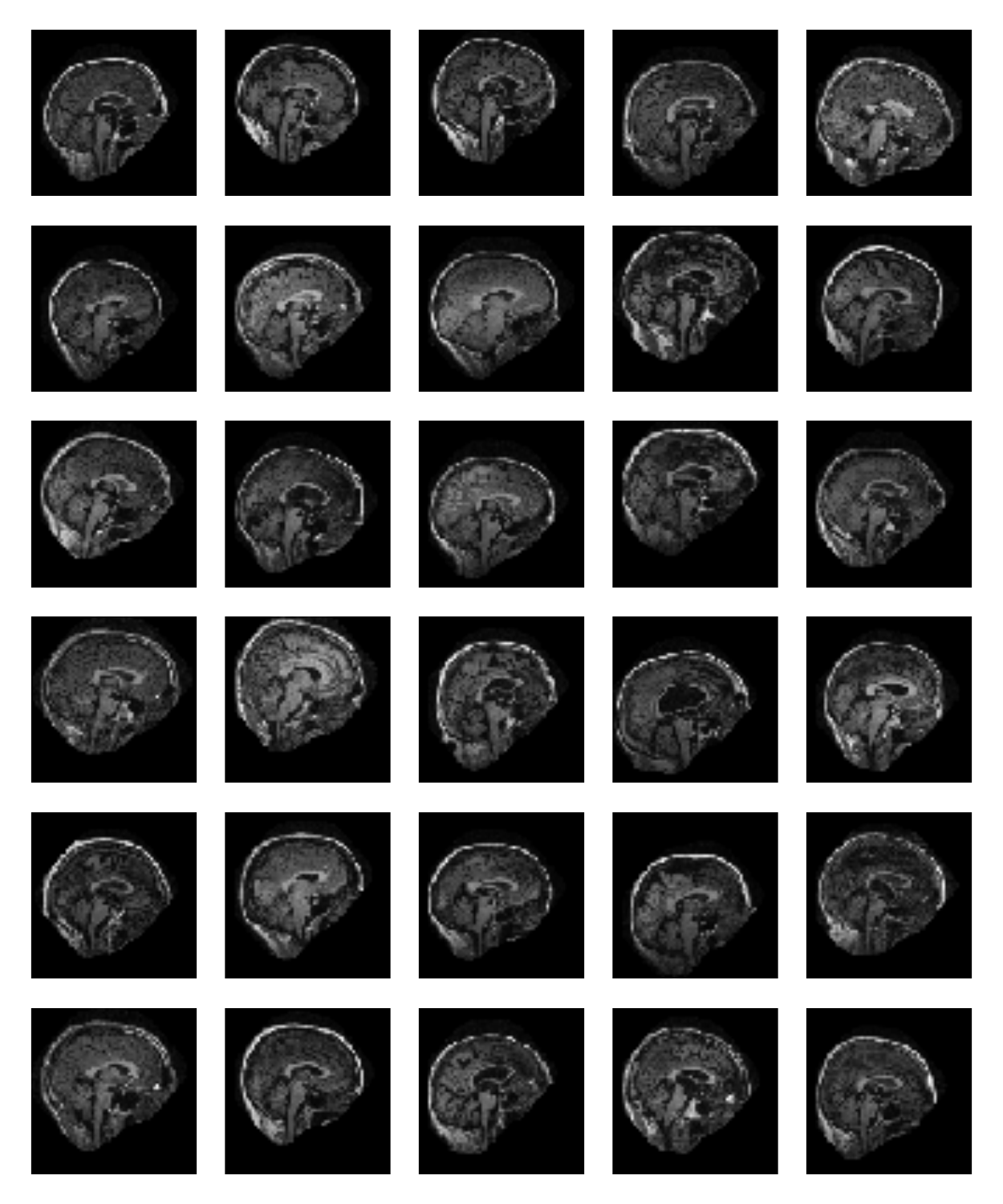}
         \vskip -0.5em
         \subcaption*{OASIS}
        \end{minipage}
    \caption{Training samples used for geodesic computation of Section~\ref{Sec: Geodesics synthetic} and Section~\ref{Sec: Geodesics real}}
    \label{fig: app Geodesic Training samples}
\end{figure}

\clearpage

\section{}
\label{app: Geodesic curves}
\subsection{VAE models architectures}\label{app: VAE models architectures}
\begin{table}[ht]
    \centering
    \begin{tabular}{c|l|l}
        \toprule
            Networks   & \multicolumn{2}{c}{Configurations}\\
            \hline                               
        $\mu_{\varphi}$ &  MLP - $(D, 400, ReLu)^{*}$\tnote{*}    & MLP - $(400, d, Linear)$ \\
        $\Sigma_{\varphi} $ & MLP - $(D, 400, ReLu)^{*}$ & MLP - $(400, d, Linear)$ \\                                                 
        $\mu_{\theta}$      & MLP - $(d, 400, Softplus)^{**}$  & MLP - $(400, D, Sigmoid)$   \\
        $\Sigma_{\theta}$ & MLP - $(d, 400, Softplus)^{**}$ & MLP - $(400, D, \tanh)$ \\
        \bottomrule
    \end{tabular}
    \begin{tablenotes}[*]\footnotesize
        \item[*] * Same layers, ** Same layers
        \end{tablenotes}
        \vspace{10pt}
        \begin{tabular}{c|l|l}
            \toprule
                Networks   & \multicolumn{2}{c}{Configurations}\\
                \hline                               
            $\mu_{\varphi}$ &  MLP - $(D, 400, ReLu)^{*}$\tnote{*}    & MLP - $(400, d, Linear)$ \\
            $\Sigma_{\varphi} $ & MLP - $(D, 400, ReLu)^{*}$ & MLP - $(400, d, Linear)$ \\                                                 
            $\pi_{\theta}$      & MLP - $(d, 400, Softplus)$  & MLP - $(400, D, Sigmoid)$   \\
            \bottomrule
        \end{tabular}
        \begin{tablenotes}[*]\footnotesize
            \item[*] * Same layers
            \end{tablenotes}

            \caption{Inference and generator neural networks architectures. \textit{Top :} Architecture used to reproduce \citet{arvanitidis2017latent}'s model with a $\mathcal{N}$-VAE. \textit{Bottom :} Architecture used to reproduce \citet{chen2018metrics}'s model with a $\mathcal{B}$-VAE. Noteworthy is the fact that we only amend the activation function and not the overall model structure.}
            \label{Table: Model arhitecture geodesics}
\end{table}

\subsection{RHVAE parameters}

Parameters used for the RHVAE used in the experiments in Section~\ref{Sec: Geodesics synthetic} Section~\ref{Sec: Geodesics real}.

\begin{table}[h!t]
    \centering
    \begin{tabular}{c|ccccc}
        \toprule
         Data set & \multicolumn{5}{c}{Parameters} \\
         & $n_{\mathnormal{lf}}$ & $\varepsilon_{\mathnormal{lf}}$ & $T$ & $\lambda$ & $\sqrt{\beta_0}$ \\
         \hline
         Synthetic data & $5$ & $10^{-2}$ & $1$ & $10^{-3}$ &0.3\\
         \hline
         MNIST & 10 & $10^{-2}$ & 1 & $10^{-2}$   & 0.3 \\
         \hline
         FashionMNIST & 3 & $10^{-2}$ & 0.5 & $10^{-1}$ & 0.3 \\
         \hline
         Olivetti & 10 & $10^{-2}$ & 0.8 & $10^{-3}$ & 0.3 \\
         \hline
         OASIS & $5$ & $10^{-3}$ & 0.8 & $10^{-3}$ &$ 0.3$\\
         \bottomrule
    \end{tabular}
    \caption{Parameters used to train the proposed RHVAE models and perform geodesic interpolations on 5 data sets.}
\end{table}

\subsection{FashionMNIST}\label{ap: Geodesic interpolation Fashion}
Geodesic interpolations on FashionMNIST.

\begin{figure}[p]
    \centering
    \begin{minipage}[c]{\linewidth}
        \centering
        \subcaption*{VAE}
        \vskip -0.5em
         \includegraphics[scale=0.5]{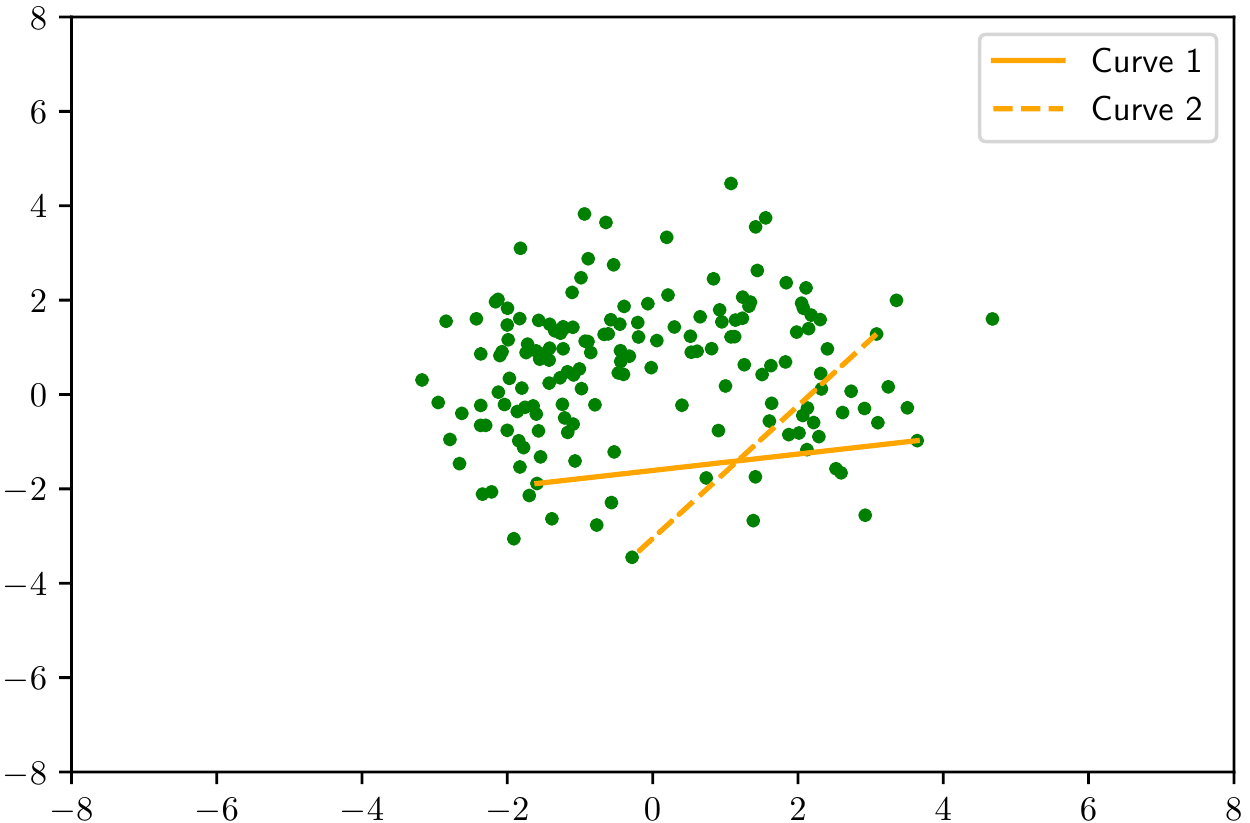}
    \end{minipage}
    \centering
    \vskip 0.5em
    \begin{minipage}[c]{\linewidth}
        \centering
         \includegraphics[scale=0.28]{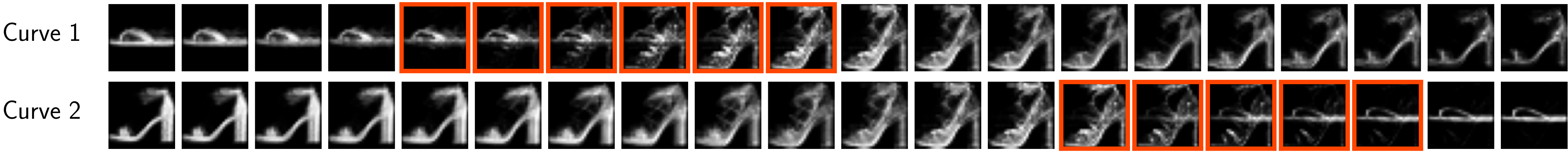}
    \end{minipage}
    \caption{Affine interpolations with a classic VAE trained with 160 samples of a single class extracted from the FashionMNIST data set and with 1000 epochs. \textit{Top:} The latent space along with the means of encoded data points and interpolation curves. \textit{Bottom}: The decoded samples along the curves (granularity of 5 time steps).}
    \label{fig: VAE Interpolation FashionMNIST}
\end{figure}

\begin{figure}[p]
    \centering
    \begin{minipage}[c]{0.32\linewidth}
        \centering
        \subcaption*{}
                \vskip -0.5em
         \includegraphics[scale=0.38]{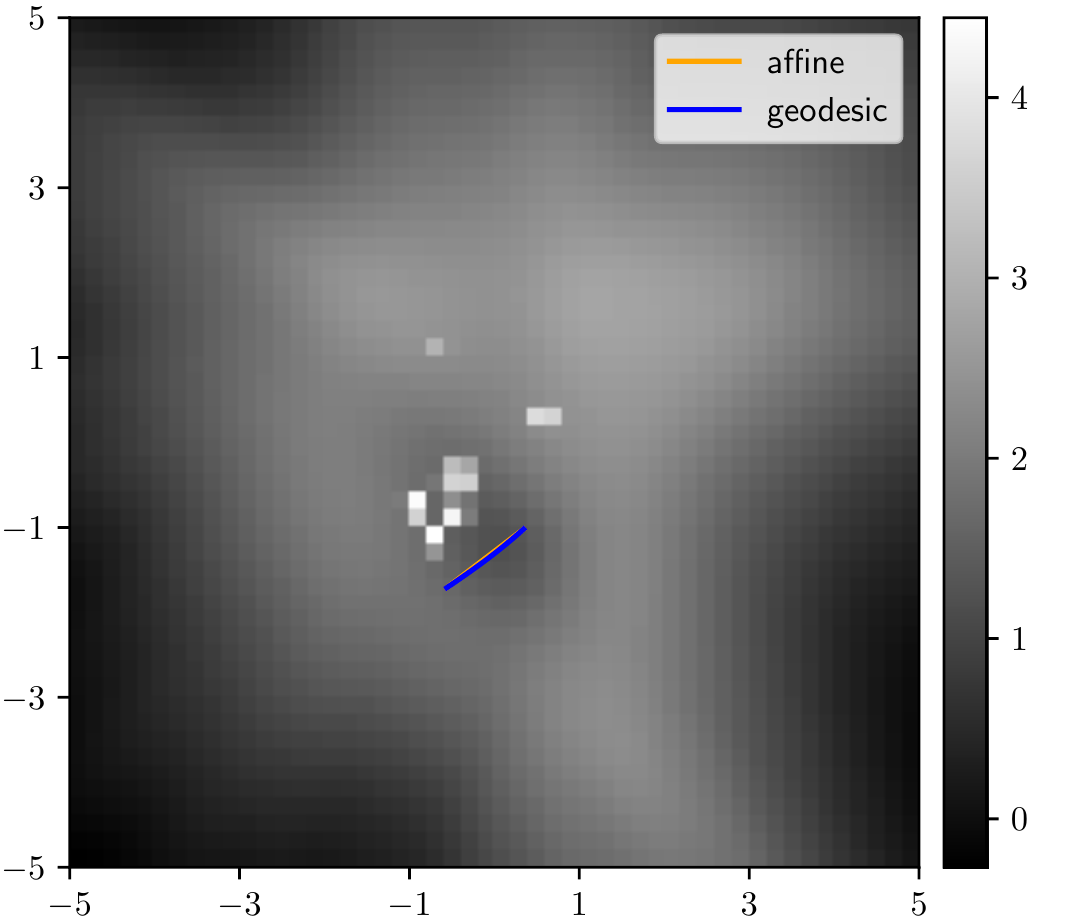}
    \end{minipage}
    \begin{minipage}[c]{0.32\linewidth}
        \centering
        \subcaption*{VAE \citep{arvanitidis2017latent}}
         \includegraphics[scale=0.38]{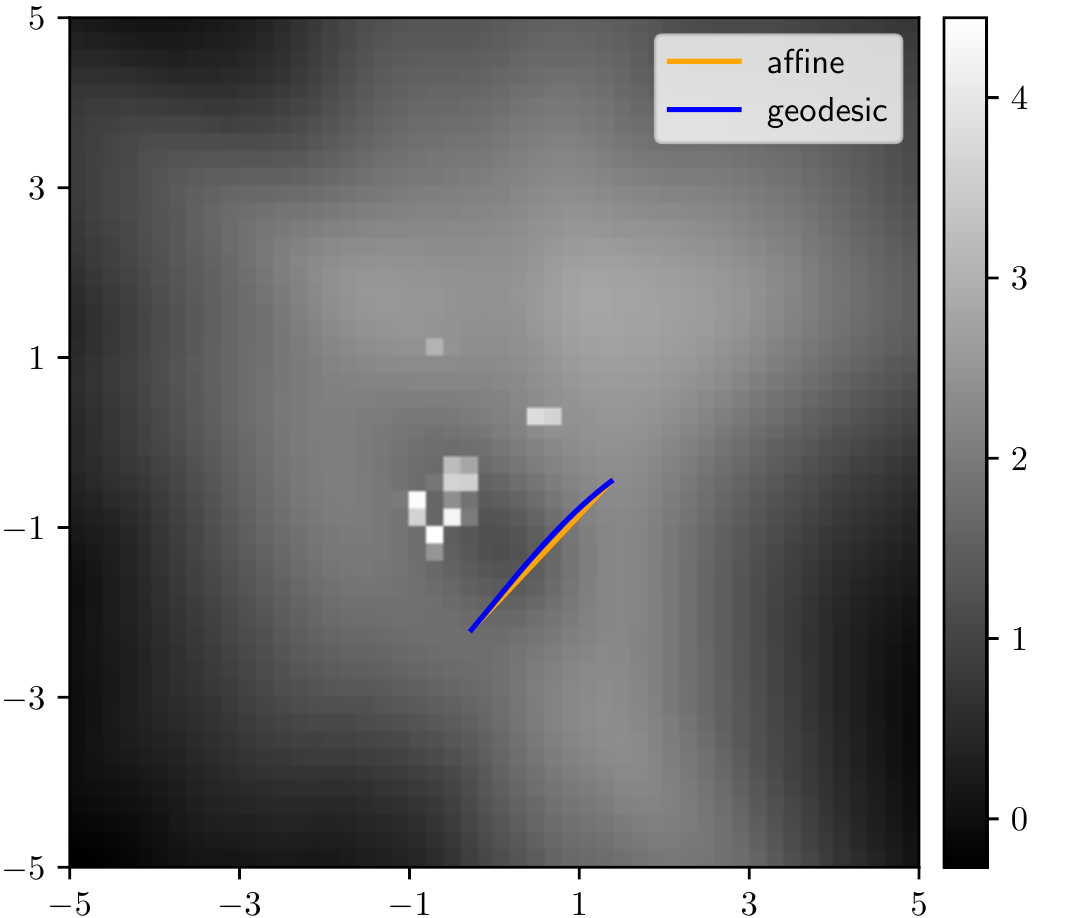}
    \end{minipage}
    \begin{minipage}[c]{0.32\linewidth}
        \centering
        \subcaption*{}
         \includegraphics[scale=0.38]{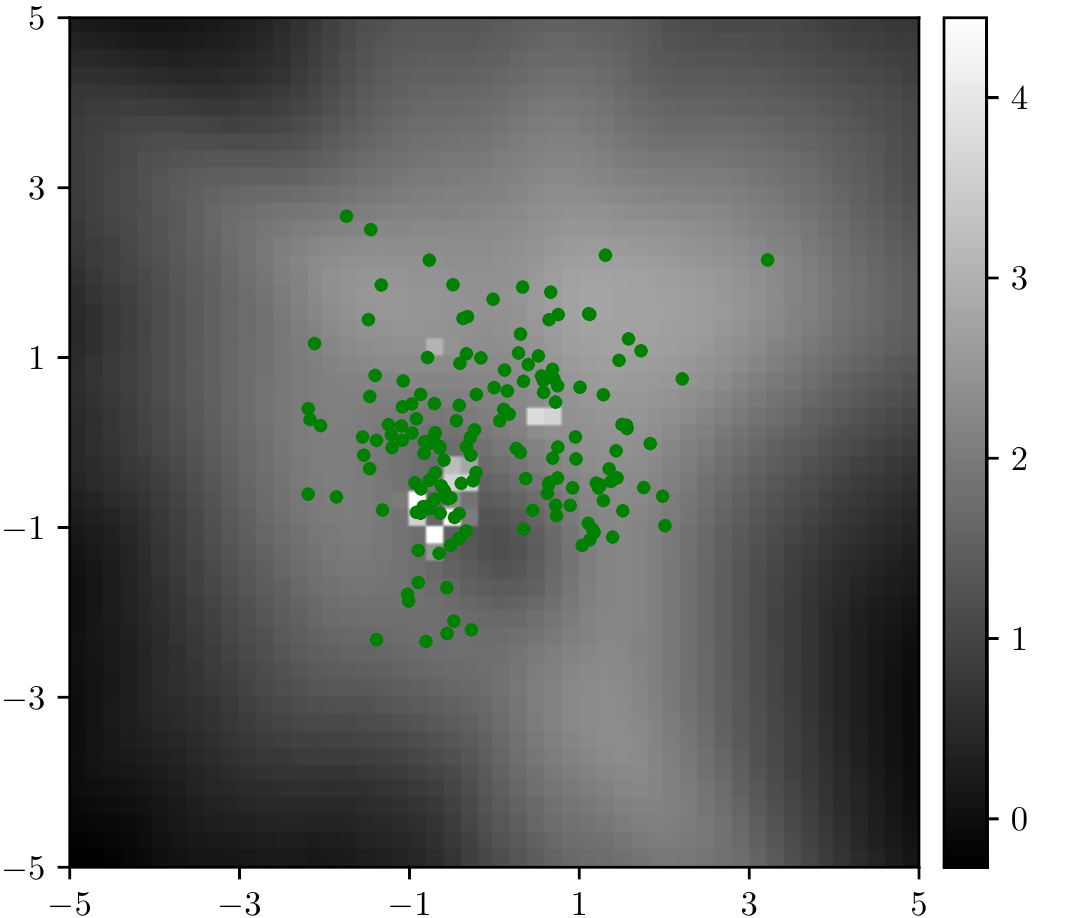}
    \end{minipage}
        \vskip 0.5em
    \centering
    \begin{minipage}[c]{\linewidth}
        \centering
         \includegraphics[scale=0.28]{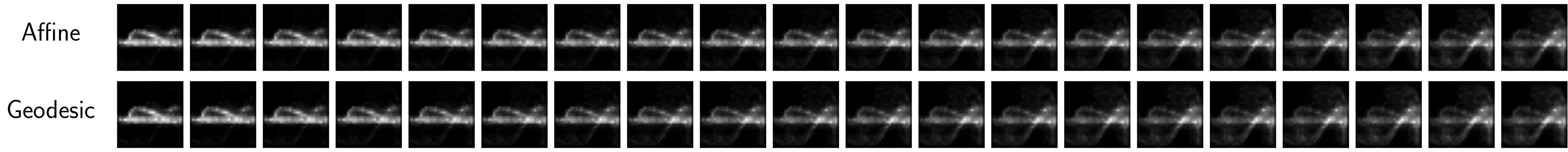}
    \end{minipage}
    \vskip 0.5em
    \centering
    \begin{minipage}[c]{\linewidth}
        \centering
         \includegraphics[scale=0.28]{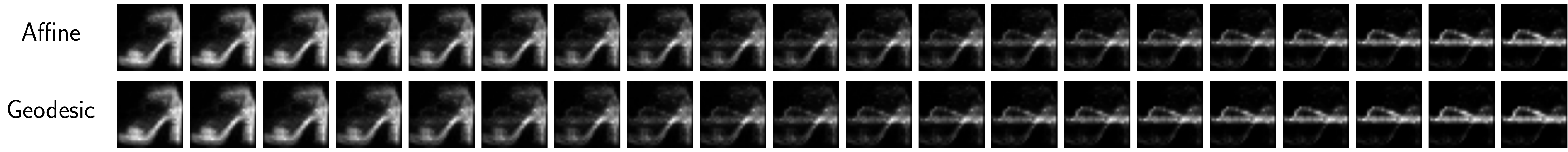}
    \end{minipage}
    \caption{Affine and geodesic interpolations with the proposed VAE trained as specified in \citep{arvanitidis2017latent} with 160 samples of a single class extracted from the FashionMNIST data set and with 1000 epochs. \textit{Top:} The latent space along with the logarithm of the volume element and interpolation curves. \textit{Bottom:} The decoded samples along the curves (granularity of 5 time steps).}
    \label{fig: Geodesic Interpolation Arvanitidis FashionMNIST}
\end{figure}

\begin{figure}[ht]
    \centering
    \begin{minipage}[c]{0.32\linewidth}
        \centering
        \subcaption*{}
        \vskip -0.5em
         \includegraphics[scale=0.38]{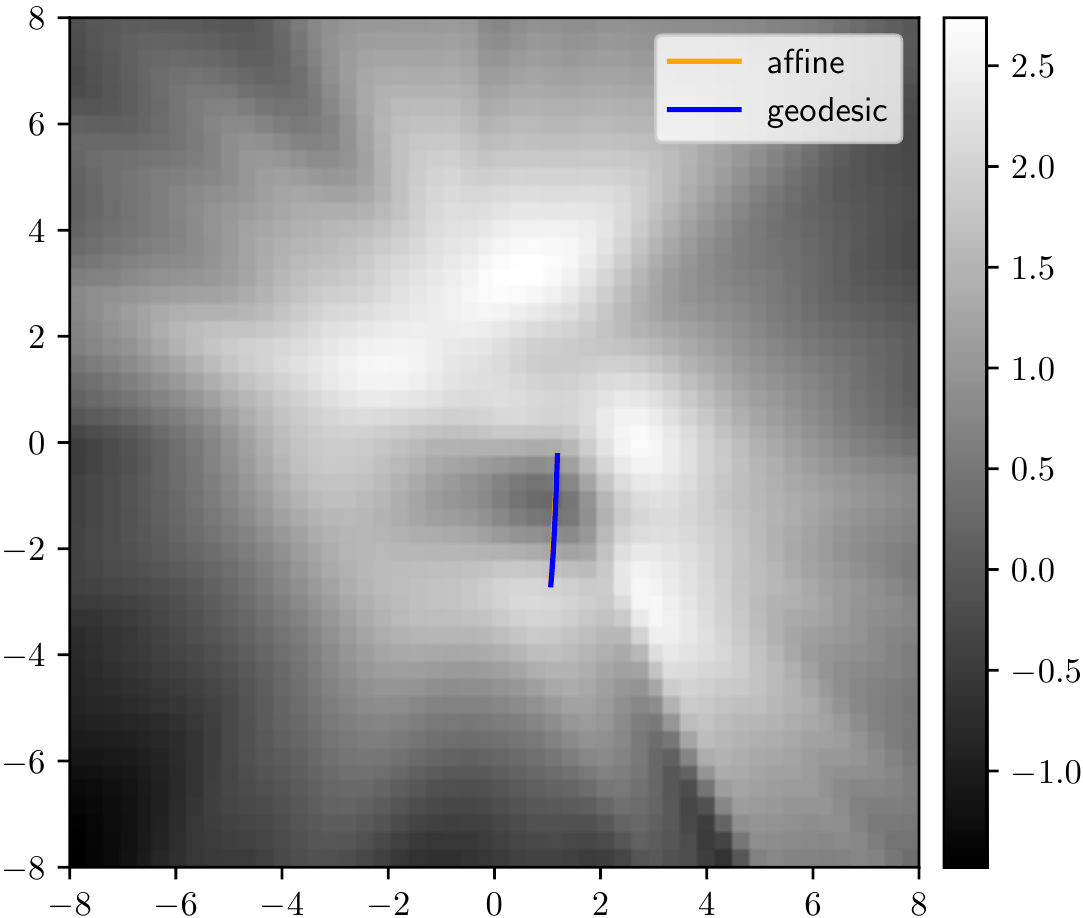}
    \end{minipage}
    \begin{minipage}[c]{0.32\linewidth}
        \centering
        \subcaption*{VAE \citep{chen2018metrics}}
         \includegraphics[scale=0.38]{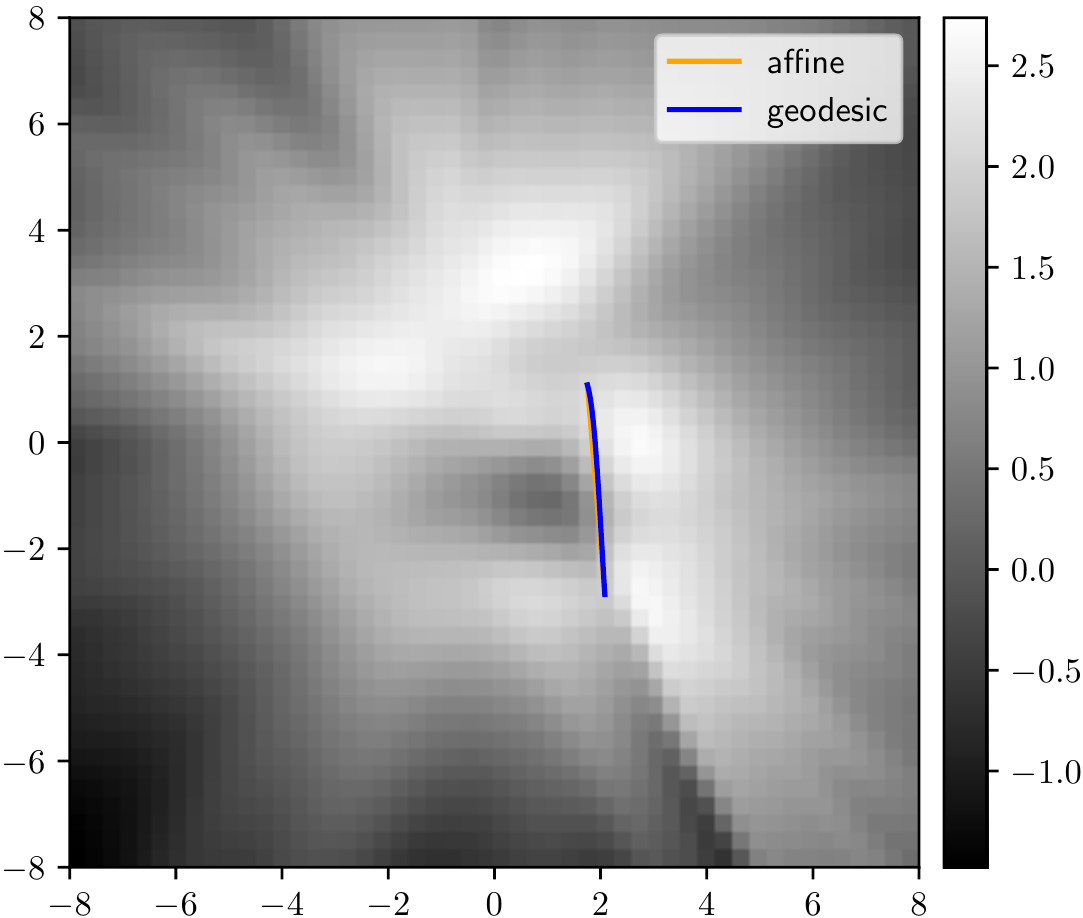}
    \end{minipage}
    \begin{minipage}[c]{0.32\linewidth}
        \centering
        \subcaption*{}
         \includegraphics[scale=0.38]{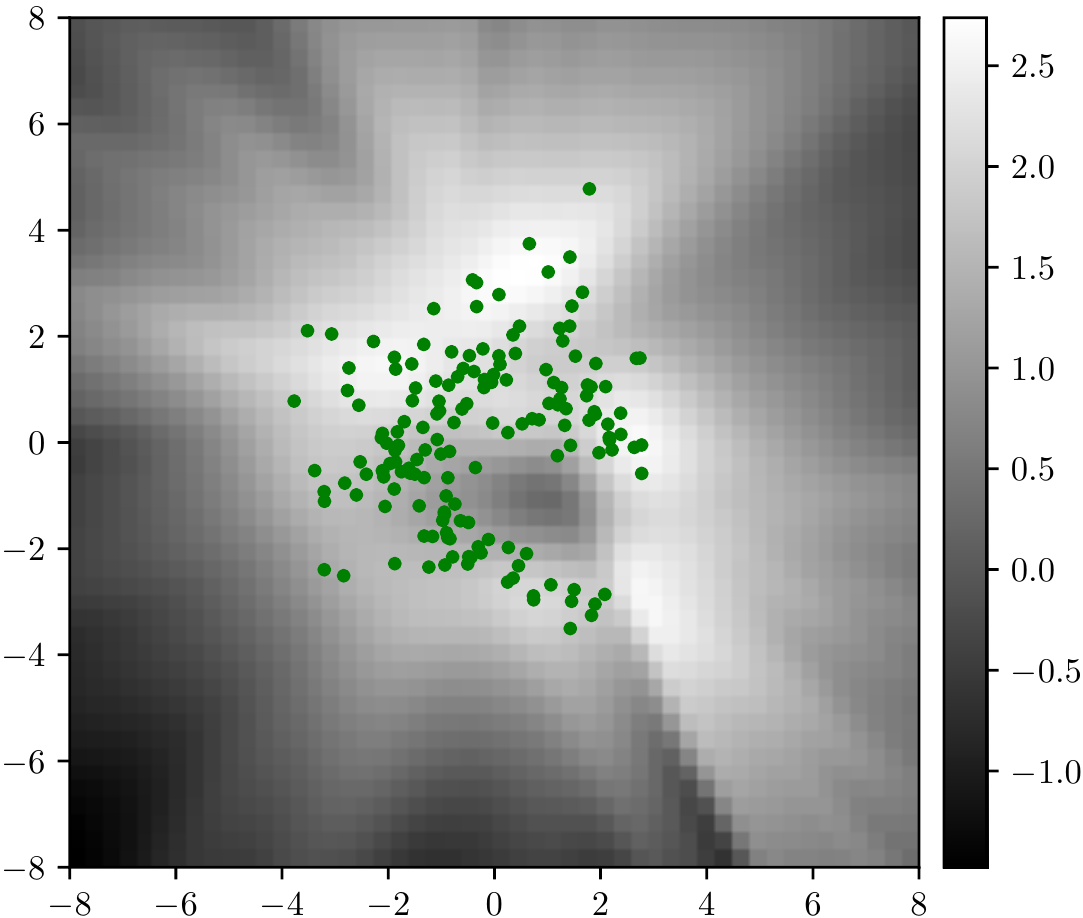}
    \end{minipage}
    \vskip 0.5em
    \centering
    \begin{minipage}[c]{\linewidth}
        \centering
         \includegraphics[scale=0.28]{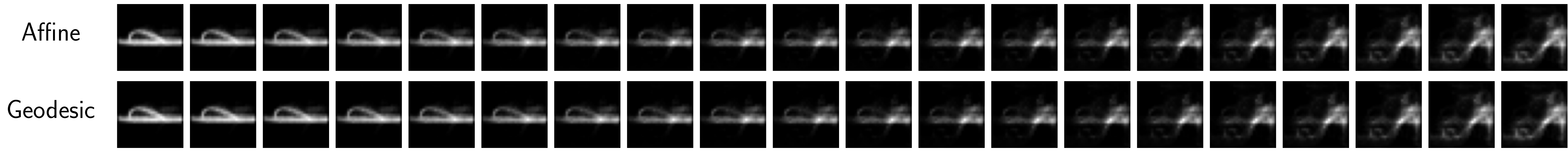}
    \end{minipage}
    \vskip 0.5em
    \centering
    \begin{minipage}[c]{\linewidth}
        \centering
         \includegraphics[scale=0.28]{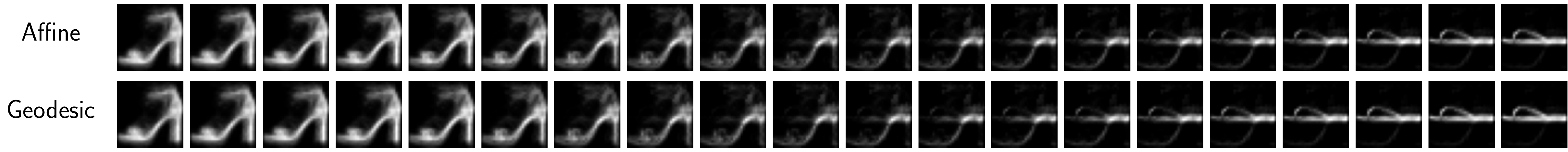}
    \end{minipage}
    \caption{Affine and geodesic interpolations with a VAE trained as specified
    in \citep{chen2018metrics} with 160 samples of a single class extracted from the FashionMNIST data set and with 1000 epochs. \textit{Top:} The latent space along with the logarithm of the volume element and interpolation curves. \textit{Bottom:} The decoded samples along the curves (granularity of 5 time steps).}
    \label{fig: Geodesic Interpolation Chen FashionMNIST}
\end{figure}

\subsection{Olivetti faces}\label{app: Geodesic interpolations Olivetti}

\begin{figure}[ht]
    \centering
    \begin{minipage}[c]{0.32\linewidth}
        \centering
        \subcaption*{}
        \vskip -0.5em
         \includegraphics[scale=0.38]{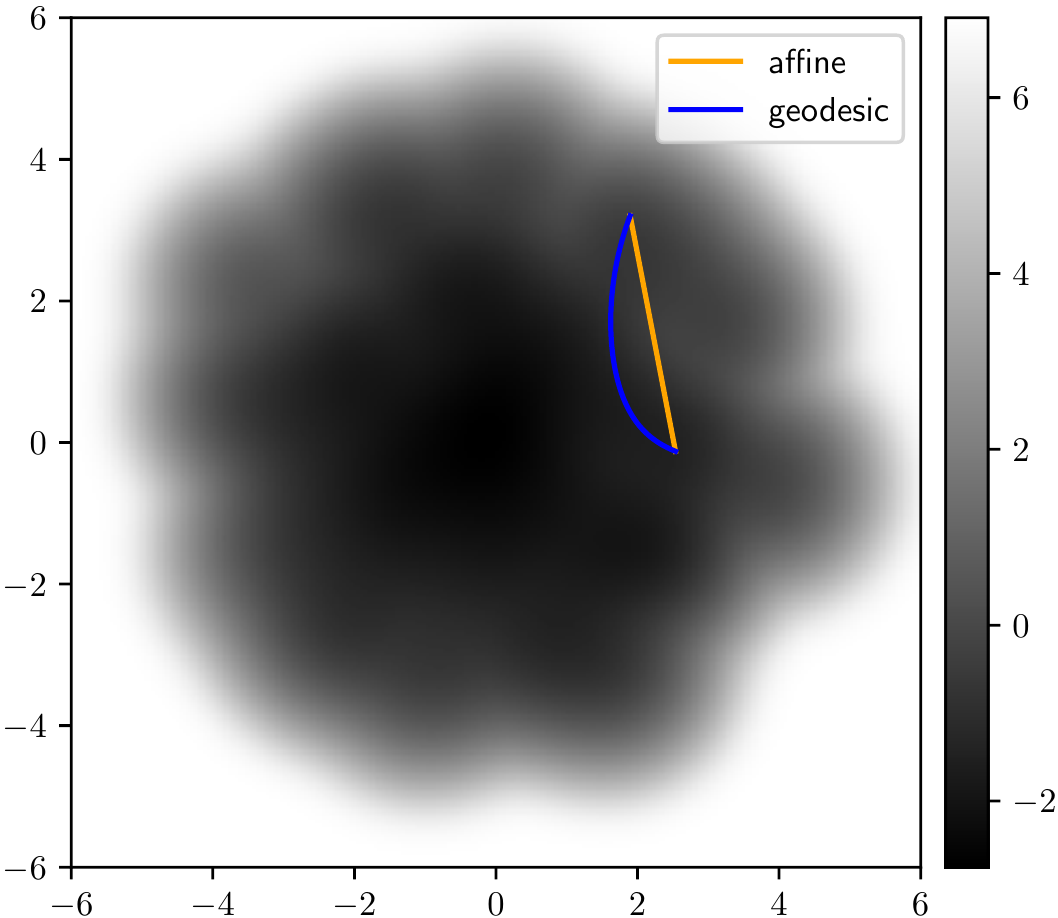}
    \end{minipage}
    \begin{minipage}[c]{0.32\linewidth}
        \centering
        \subcaption*{RHVAE (Ours)}
        \vskip -0.5em
         \includegraphics[scale=0.38]{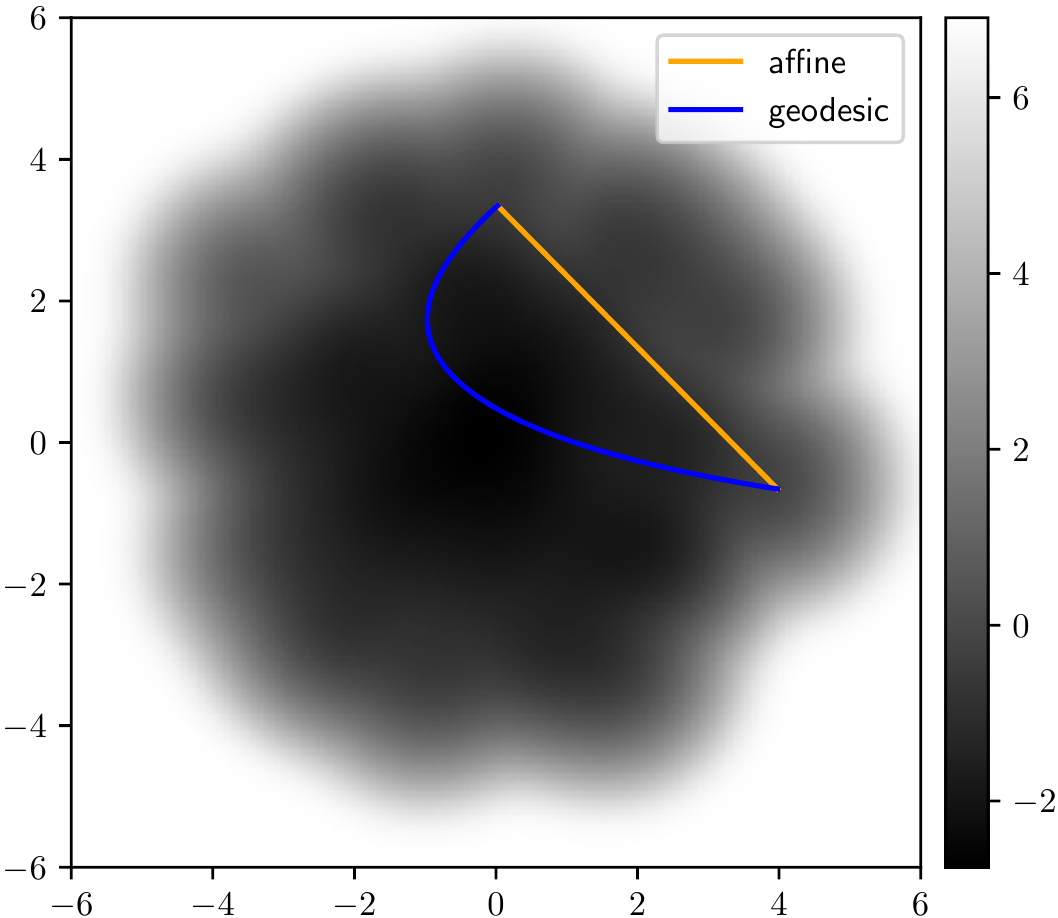}
    \end{minipage}
    \begin{minipage}[c]{0.32\linewidth}
        \centering
        \subcaption*{}
        \vskip -0.5em
         \includegraphics[scale=0.38]{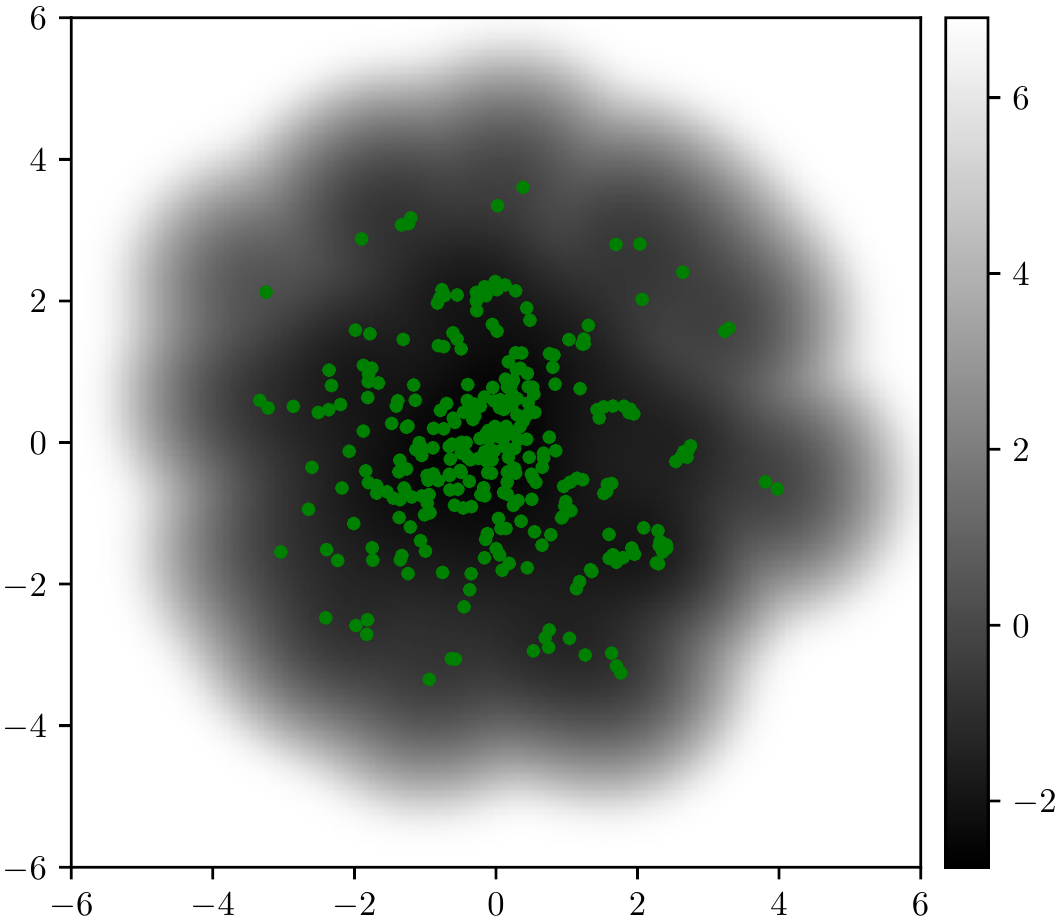}
    \end{minipage}
    \vskip 0.3em
    \centering
    \begin{minipage}[c]{\linewidth}
        \centering
         \includegraphics[scale=0.28]{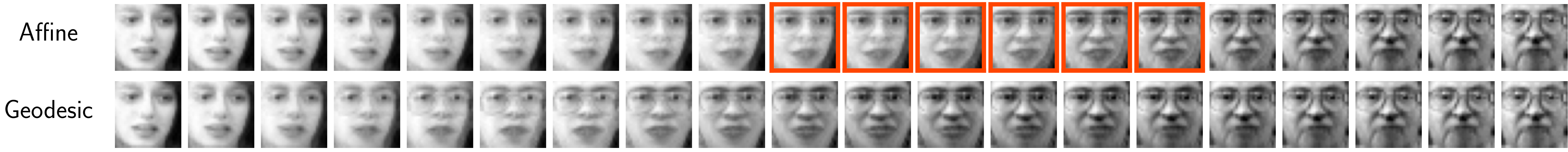}
    \end{minipage}
    \vskip 0.5em
    \centering
    \begin{minipage}[c]{\linewidth}
        \centering
         \includegraphics[scale=0.28]{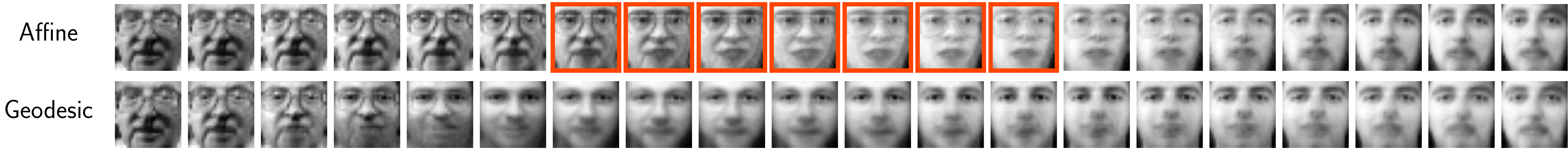}
    \end{minipage}
    \caption{Affine and geodesic interpolations with the proposed RHVAE trained on the Olivetti faces data set. The model is trained with 3000 epochs on 80 \% of the data set randomly chosen. \textit{Top:} The latent space along with the logarithm of the volume element and interpolation curves. \textit{Bottom:} The decoded samples along the curves (granularity of 5 time steps).}
    \label{fig: Geodesic Interpolation Olivetti}
\end{figure}

We also test the model on the Olivetti faces data set
\citep{olivettifaces} and try to see if geodesic interpolations between faces remain meaningful. On this very data set, the improvement coming from geodesic interpolation is more difficult to perceive since the learned metric is more round. Nonetheless, it can be noted that again the geodesic interpolation is smoother. On the top rows of Figure~\ref{fig: Geodesic Interpolation Olivetti}, the face contour are blurrier for the affine interpolation since it only superposes two different faces whereas this aspect is mitigated for the geodesic interpolation. In the bottom rows, some frames of the interpolation does not make much sense. Indeed, we try to interpolate two faces of people looking right in front
of them but some points of the latent space correspond to people looking on the
right side. We would expect to see the face keeping the same orientation all
along the interpolation which seems to be the case for the geodesic
interpolation.

\clearpage

\section{}\label{app: Generation parameters}

Parameters used for the RHVAE used in the experiments in Section~\ref{Sec: Generation}.

\begin{table}[h!t]
    \centering
    \begin{tabular}{c|cccccc}
        \toprule
         Data set & \multicolumn{5}{c}{Parameters} \\
         &$d^{*}$ & $n_{\mathnormal{lf}}$ & $\varepsilon_{\mathnormal{lf}}$ & $T$ & $\lambda$ & $\beta_0$ \\
         \hline
         MNIST &$ 2$ &3 & $10^{-3}$ & 0.8 & $10^{-2}$   & 0.3 \\
         \hline
         FashionMNIST & $10$ &3 & $10^{-3}$ & 0.8 & $10^{-2}$ & 0.3 \\
         \hline
         Olivetti & $15$ & 3 & $10^{-3}$ & 0.8 & $10^{-3}$ & 0.3 \\
         \bottomrule
    \end{tabular}
    \begin{tablenotes}[*]\footnotesize
        \item[*] * The latent space dimension is the same for VAE models
        \end{tablenotes}
    \caption{Parameters used to train the proposed RHVAE models to perform samples generation on 3 data sets.}
    
\end{table}

\clearpage
\section{}\label{app: Clustering paramters}

Parameters used for the experiments in Section~\ref{Sec: Clustering}.

\begin{table}[h!t]
    \centering
    \begin{tabular}{c|ccccc}
        \toprule
         data set & \multicolumn{5}{c}{Parameters} \\
         & $n_{\mathnormal{lf}}$ & $\varepsilon_{\mathnormal{lf}}$ & $T$ & $\lambda$ & $\sqrt{\beta_0}$ \\
         \hline
         Synthetic data & 3 & $10^{-2}$ & 0.8 & $10^{-3}$   & 0.3\\
         \hline \hline
         MNIST 1 & 10 & $10^{-2}$ & 0.8 & $10^{-3}$   & 0.3 \\
         MNIST 2 & 10 & $10^{-2}$ & 0.8 & $10^{-3}$  & 0.3 \\
         MNIST 3 & 10 & $10^{-2}$ & 0.8 & $10^{-3}$ & 0.3\\
         \hline \hline
         FashionMNIST 1 & 10 & $10^{-2}$ & 0.8 & $10^{-3}$ & 0.3 \\
         FashionMNIST 2 & 10 & $10^{-2}$ & 0.8 & $10^{-3}$ & 0.3 \\
         FashionMNIST 3 & 10 & $10^{-3}$ & 0.8 & $10^{-3}$ & 0.3 \\
         \bottomrule
    \end{tabular}
    \caption{Parameters used to train the proposed RHVAE models to compare affine and geodesic clustering under both metrics.}
\end{table}

\begin{table}[ht]
\centering
\begin{tabular}{c|l|l}
        \toprule
            Networks   & \multicolumn{2}{c}{Configurations}\\
            \hline                               
        $\mu_{\varphi}$ &  MLP - $(D, 400, ReLu)^{*}$\tnote{*}    & MLP - $(400, d, Linear)$ \\
        $\Sigma_{\varphi} $ & MLP - $(D, 400, ReLu)^{*}$ & MLP - $(400, d, Linear)$ \\                                                 
        $\pi_{\theta}$      & MLP - $(d, 400, ReLu)$  & MLP - $(400, D, Sigmoid)$   \\
        $L_{\psi} $ (diag)  & MLP - $(D, 400, ReLu)^{**}$& MLP - $(400, d, Linear)$ \\
        $L_{\psi} $ (lower) & MLP - $(D, 400, ReLu)^{**}$& MLP - $(400, \frac{d(d-1)}{2}, Linear)$\\
        \bottomrule
    \end{tabular}
    \begin{tablenotes}[*]\footnotesize
        \item[*] * Same layers, ** Same layers
        \end{tablenotes}
        \caption{Inference and generator neural networks used for the RHVAE along with the neural network shapes used for metric learning in Section~\ref{Sec: Clustering}.}
        \label{Table: Model architectures - Clustering}
\end{table}

\clearpage
\bibliography{references}

\end{document}